\documentclass{article}

\PassOptionsToPackage{square, numbers}{natbib} 

\usepackage{graphicx}
\usepackage{amsmath}
\usepackage{caption}

\usepackage{algorithm}
\usepackage{algpseudocode}
\usepackage[noend]{algcompatible}
\newcommand{\Commentx}[1]{\hfill\(\triangleright\) #1}
\usepackage{subcaption}

\usepackage{booktabs}
\usepackage{multirow}
\usepackage{graphicx}
\usepackage{amsmath}
\usepackage{caption}
\usepackage{hyperref}
\usepackage{siunitx}
\usepackage{subcaption}

\usepackage{subcaption}
\usepackage{float}  
\usepackage[verbose=true,letterpaper]{geometry}
\usepackage[preprint]{neurips_2025}

\usepackage[utf8]{inputenc} 
\usepackage[T1]{fontenc}    
\usepackage{hyperref}       
\usepackage{url}            
\usepackage{booktabs}       
\usepackage{amsfonts}       
\usepackage{nicefrac}       
\usepackage{microtype}      
\usepackage{xcolor}         

\title{Extending Dataset Pruning to Object Detection: \\A Variance-based Approach}

\author{%
  Ryota Yagi \\
  Department of Computer Science and Engineering\\
  University of Nevada, Reno\\
  1664 N Virginia St, Reno, NV 89557 \\
  \texttt{ryagi@unr.edu} \\
}

\begin{document}

\maketitle

\begin{abstract}
Dataset pruning---selecting a small yet informative subset of training data---has emerged as a promising strategy for efficient machine learning, offering significant reductions in computational cost and storage compared to alternatives like dataset distillation. 
While pruning methods have shown strong performance in image classification, their extension to more complex computer vision tasks, particularly object detection, remains relatively underexplored. 
In this paper, we present the first principled extension of classification pruning techniques to the object detection domain, to the best of our knowledge.
We identify and address three key challenges that hinder this transition: the Object-Level Attribution Problem, the Scoring Strategy Problem, and the Image-Level Aggregation Problem. 
To overcome these, we propose tailored solutions, including a novel scoring method called Variance-based Prediction Score (VPS). 
VPS leverages both Intersection over Union (IoU) and confidence scores to effectively identify informative training samples specific to detection tasks.
Extensive experiments on PASCAL VOC and MS COCO demonstrate that our approach consistently outperforms prior dataset pruning methods in terms of mean Average Precision (mAP). 
We also show that annotation count and class distribution shift can influence detection performance, but selecting informative examples is a more critical factor than dataset size or balance. 
Our work bridges dataset pruning and object detection, paving the way for dataset pruning in complex vision tasks.
\end{abstract}

\section{Introduction}
As modern machine learning continues to scale, the pursuit of Efficient AI—learning with less computational, data, and energy resources—has become one of central themes across domains. 
Efforts to reduce the cost of training and deployment span various dimensions, including model compression~\cite{han2016deep}, efficient architectures~\cite{tan2019efficientnet}, low-precision computation~\cite{jacob2018quantization}, and hardware-aware optimization~\cite{howard2017mobilenets}.

Beyond architectural and algorithmic improvements, a parallel line of research has focused on improving data efficiency—reducing the amount of data required to train high-performing models. 
In this context, two prominent approaches have emerged: dataset pruning and the more recent dataset distillation. 
Both methods aim to construct a smaller, yet highly informative subset of the original dataset that can match or even surpass the performance of training on the original set.

The two approaches differ fundamentally in how these subsets are obtained. Dataset pruning involves selecting a subset of real examples from the original dataset, typically based on some informativeness criterion. 
In contrast, dataset distillation seeks to generate synthetic data designed to approximate its distribution of original datasets~\cite{wang2018dataset, zhao2021dataset}. 
While dataset distillation is conceptually appealing, it faces several practical limitations. 
It often entails significant computational and memory overheads that hinder scalability~\cite{cazenavette2022dataset, du2023minimizing}, relies heavily on soft-label supervision~\cite{su2024d, yin2023squeeze} that increases storage costs~\cite{xiao2024are}, and struggles to generalize unseen model architectures~\cite{cazenavette2023generalizing, zhao2023boosting}. 
Moreover, distilled datasets frequently fall short in performance when compared to models trained on the original data~\cite{chen2025curriculum, qi2024fetch}.

In contrast, dataset pruning offers a more pragmatic and scalable solution. 
It operates linearly with dataset size, retains the original data distribution, and requires no synthetic generation or additional storage beyond the selected subset.
Pruned datasets have also demonstrated robust generalization across different model architectures and, in some cases, even outperform training on the full dataset by removing noisy or redundant samples~\cite{paul2021deep}.

Despite the demonstrated effectiveness of dataset pruning in image classification tasks, its application to more complex visual recognition problems—particularly object detection—remains underexplored. 
Object detection introduces additional challenges due to its dual objectives of localization and classification, the presence of multiple objects per image, and the high variability in object scale, aspect ratio, and spatial arrangement. 
These factors complicate the direct transfer of pruning strategies originally designed for classification.

In this work, we address this gap by extending dataset pruning techniques to object detection. Our main contributions are as follows:

\begin{itemize}
\item We introduce a principled extension of dataset pruning for object detection and address three key challenges: the Object-Level Attribution Problem, the Scoring Strategy Problem, and the Image-Level Aggregation Problem.
\item We propose a scoring method called Variance-based Prediction Score (VPS), which utilizes Intersection over Union (IoU) and the variance of confidence scores to identify informative training examples.
\item We demonstrate through extensive experiments on PASCAL VOC and MS COCO that our method consistently improves mean Average Precision (mAP), highlighting the importance of sample informativeness over annotation count or class distribution balance.
\end{itemize}

\section{Related Works}
\subsection{Dataset Pruning on Classification Task}
Dataset pruning, often referred to as coreset selection, has been a long-standing area of research focused on reducing the size of the data set while maintaining performance~\cite{AgarwalHV04, badoiu2003smaller, STOC2004HarPeledM, har2004coresets, skalak1997cmpsci}. 
More recent advancements have categorized pruning techniques into three primary approaches: \textbf{geometry-based}, \textbf{score-based} and \textbf{post-hoc sampling} strategies, each addressing the challenge of subset selection from distinct perspectives.

Geometry-based methods exploit the spatial structure of data in feature space to select representative subsets. Herding~\cite{Welling2009} minimizes the distance between the full and subset centroids, while K-Center~\cite{Farahani2009} selects points to minimize the maximum distance to the nearest center. SSP~\cite{sorscher2022beyond} ensures diversity by choosing samples farthest from k-means centers.

Score-based methods evaluate data points to select representative subsets by assigning importance scores. 
For example, EL2N~\cite{paul2021deep} uses the L2 norm of prediction errors, while Entropy~\cite{coleman2019selection} leverages the prediction probability entropy at the end of training. 
Forgetting~\cite{toneva2019empirical} tracks "forgetting events"—when a model's prediction shifts from correct to incorrect across training—to identify valuable samples. 
Nevertheless, the discrete nature of the score often necessitates random selection among equally scored data points, thereby hindering strict evaluation based solely on the score.
Advanced methods like Dyn-UNC~\cite{he2024large} and TDDS~\cite{zhang2024spanning} refine selection by incorporating training dynamics. 
Similarly, AUM~\cite{pleiss2020identifying} detects mislabeled data by accumulating the margin between the top and second-highest predicted probabilities. 
A key limitation of these methods is that they typically require hundreds of training epochs to exploit such training dynamics, and their effectiveness does not necessarily extend to object detection tasks.

At a higher level, post-hoc sampling strategies mainly focus on how to sample once a scoring metric has been defined.
CCS~\cite{zheng2022coverage} uses stratified sampling but is limited by random selection within bins. Moderate~\cite{xia2022moderate} targets mid-score samples for balanced representation. More sophisticated methods like BOSS~\cite{acharya2024balancing} and DUAL~\cite{cho2025lightweightdatasetpruningtraining} apply beta sampling to capture both difficulty and diversity, while D2~\cite{maharana2023d2} introduces graph-based message passing to refine this balance. 
Rather than focusing on how to assign scores to data points, these methods primarily explore how to sample from already-scored instances.
In contrast, we leverage the unique characteristics of object detection to introduce a strictly score-based sampling strategy that avoids reliance on any sampling methods.

\subsection{Dataset Pruning Beyond Image Classification}

Dataset pruning has been actively studied not only in vision tasks but also in various other fields, including natural language processing~\cite{du2022glam, wenzek2020ccnet}, speech recognition~\cite{lai2021parp, xiao2024dynamic}, time series analysis~\cite{jin2024llm}, and recommendation systems~\cite{Arabzadeh2024GreenRecSys, Sachdeva2021SVPCF}.
In the vision domain, recent efforts have applied pruning techniques to instance segmentation~\cite{dai2025trainingfree}, object re-identification~\cite{yang2024data}, and object detection. 
For example, prior work includes anchor pruning for one-stage detectors~\cite{bonnaerens2022anchor}, as well as active learning~\cite{wang2025adp} and geometry-based methods~\cite{lee2024coreset} for informative sample selection.
However, these studies overlook connections to well-established techniques in image classification and are limited in scope and evaluation. 
In contrast, our work is the first to naturally extend traditional dataset pruning techniques to object detection, aiming to unify and generalize principles across tasks.

\section{Methodology}

\begin{figure}[t]
  \centering
  \includegraphics[width=\linewidth]{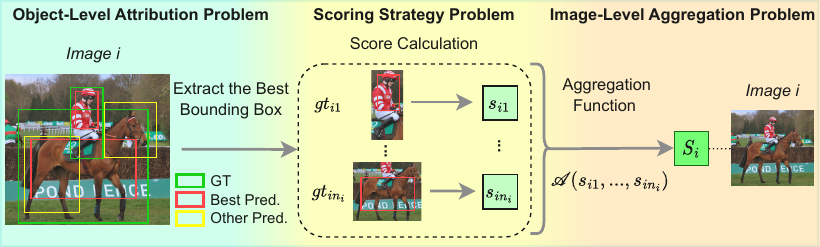}
  \caption{
  Overview of the score assignment process for object detection. For each ground-truth box, the most plausible prediction is selected (teal; Section~\ref{sec:defprediction}). 
  Scores are then computed from the associated model outputs (yellow; Sections~\ref{sec:traditional},~\ref{sec:VPS}), aggregated (orange; Section~\ref{sec:aggscore}), and assigned to the image.
  }
  \label{fig:odpruning}
\end{figure}

Several recent studies have investigated dataset pruning in the context of object detection~\cite{bonnaerens2022anchor, lee2024coreset, wang2025adp}.  
However, these approaches largely overlook the rich insights and methodologies developed in the classification task literature.  
We argue that this disconnect arises from fundamental differences between classification and object detection,  
which make direct adaptation of pruning strategies non-trivial.
In particular, dataset pruning on object detection introduces the following key questions:

\begin{enumerate}
    \item \textbf{How can we assign the best prediction to individual objects?}
    \item \textbf{What is the most effective scoring method for object detection task?}
    \item \textbf{How can we aggregate scores for multiple objects?}
\end{enumerate}

We refer to the first challenge as the \textit{Object-Level Attribution Problem}, the second as the \textit{Scoring Strategy Problem}, 
and the last as the \textit{Image-Level Aggregation Problem}.  
These issues are fundamental to dataset pruning for object detection and must be explicitly addressed.  
To illustrate these challenges and their interconnections, we provide a conceptual overview in Figure~\ref{fig:odpruning}, which highlights the distinct stages of pruning in object detection and the associated complexities.

In the remainder of this section, we first formalize the problem setting in Section~\ref{sec:defprob}.  
We then address the \emph{Object-Level Attribution Problem} in Section~\ref{sec:defprediction}.  
Next, we discuss the \emph{Scoring Strategy Problem}, focusing on the application of traditional scoring methods to object detection in Section~\ref{sec:traditional}, 
and propose a novel scoring strategy specifically tailored for this task in Section~\ref{sec:VPS}, 
with a detailed analysis provided in Section~\ref{sec:VPSAnalyze}.
Finally, we introduce a solution to the \emph{Image-Level Aggregation Problem} in Section~\ref{sec:aggscore}.

\subsection{Formulation of Dataset Pruning in Object Detection}
\label{sec:defprob}
Let the original dataset be \( \mathcal{D} = \{(x_i, b_i, y_i)\}_{i=1}^{N} \), where \( x_i \in \mathcal{X} \) is the \( i \)-th input image, \( b_i = \{b_{ij}\}_{j=1}^{n_i} \subset \mathbb{R}^{4 \times n_i} \) is the set of bounding boxes for the \( n_i \) objects in \( x_i \), and \( y_i = \{y_{ij}\}_{j=1}^{n_i} \subset \mathcal{Y} \) is the corresponding set of class labels. Dataset pruning aims to find the smallest subset \( \mathcal{S} \subseteq \mathcal{D} \) such that the model performs well on the test set. There exists an inherent trade-off between data size and model performance: reducing dataset size too much may degrade accuracy, while retaining more data improves performance but increases computational cost. This is formally expressed as:
\begin{equation}
\arg\min_{\mathcal{S} \subseteq \mathcal{D}} \left\{ |\mathcal{S}| - \lambda \cdot \mathbb{E}_{(x, b, y) \sim \mathcal{D}_{\text{test}}} \left[ \Phi(\theta_{\mathcal{S}}, x, b, y) \right] \right\}    ,
\end{equation}
where \( \theta_{\mathcal{S}} \) denotes the model parameters learned from the subset \( \mathcal{S} \),  
\( \mathcal{D}_{\text{test}} \) is the test dataset,  
\( \Phi \) is a task-specific evaluation metric such as mean Average Precision (mAP) or IoU,  
and \( \lambda > 0 \) is a trade-off parameter controlling the importance of performance.

Similar to the previous work~\cite{lee2024coreset}, our approach prioritizes the number of images over the number of annotations, given that the storage cost for images is typically several tens of times higher than that for annotations (Appendix Table~\ref{tab:oddataset}).

\subsection{Object-Level Attribution Problem}
\label{sec:defprediction}

In object detection tasks, assigning the most plausible prediction to each ground truth object is a fundamental challenge. 
Unlike image classification, where the model outputs a single vector of class-wise logits (or probabilities) for the entire image, object detection models produce multiple predictions per image, each consisting of bounding box coordinates, class scores, and confidence values.
This multiplicity of outputs makes attribution inherently complex.

To address this, we propose a matching procedure for assigning a model prediction to each ground truth object, as detailed in Algorithm~\ref{alg:prediction_matching_od}, which we refer to as \textbf{Class-Prioritized IoU-Aware Prediction Assignment (CIPA)}.
This algorithm returns, for each object \( j \) in image \( i \), the best-matching prediction \( p_{ij} \) at each epoch, forming the set \( \mathcal{P}_i = \{P_{i@1}, \dots, P_{i@T}\} \). 
For each ground truth object, the prediction with the highest IoU among those sharing the same class is selected; otherwise, the prediction with the highest IoU across all classes is chosen. 
If no valid prediction exists, the match is deemed undefined.
This procedure ensures that each ground-truth object is assigned its most representative prediction, enabling object-level evaluation in detection tasks.

\begin{algorithm}[H]
\caption{Class-Prioritized IoU Aware Prediction Assignment}
\label{alg:prediction_matching_od}
\textbf{Input:} Ground truth objects for image $i$: $GT_i = \{gt_{i1}, \ldots, gt_{in_i}\}$, predicted $m$ bounding boxes at epoch $t$: $Pred_{i@t} = \{pred_{i1}, \dots, pred_{im}\}$, where $gt_{ij} = (b_{gt_{ij}}, y_{gt_{ij}})$ and $pred_{ik} = (b_{pred_{ik}}, y_{pred_{ik}})$. Here, $b$ denotes the bounding box and $y$ denotes the class label.\\
\textbf{Output:} Selected predictions for each object in image $i$ over all epochs: $\mathcal{P}_i = \{P_{i@1}, \dots, P_{i@T}\}$, where $P_{i@t} = \{p_{ij}\}_{j=1}^{n_i}$ is the set of assigned predictions at epoch $t$.

\begin{algorithmic}[1]
\STATE $\mathcal{P}_i \gets \emptyset$
\FOR{$t = 1$ to $T$}
    \STATE $P_{i@t} \gets \emptyset$
    \FOR{each $gt_{ij} \in GT_i$}
        \STATE $CandPred \gets \{ pred_{ik} \in Pred_{i@t} \mid \text{IoU}(b_{gt_{ij}}, b_{pred_{ik}}) > 0 \}$
        \IF{$CandPred \neq \emptyset$}
            \STATE $ClassPredSameClass \gets \{ pred_{ik} \in CandPred \mid y_{pred_{ik}} = y_{gt_{ij}} \}$
            \IF{$ClassPredSameClass \neq \emptyset$}
                \STATE $p_{ij} \gets \arg\max_{pred_{ik} \in ClassPredSameClass} \text{IoU}(b_{gt_{ij}}, b_{pred_{ik}})$    
            \ELSE
                \STATE $p_{ij} \gets \arg\max_{pred_{ik} \in CandPred} \text{IoU}(b_{gt_{ij}}, b_{pred_{ik}})$
            \ENDIF
        \ELSE
            \STATE $p_{ij} \gets \emptyset$ \Commentx{Assign empty if no candidate predictions are available}
        \ENDIF
        \STATE $P_{i@t} \gets P_{i@t} \cup \{p_{ij}\}$ \Commentx{Store the best prediction for $gt_{ij}$}
    \ENDFOR
    \STATE $\mathcal{P}_i \gets \mathcal{P}_i \cup \{P_{i@t}\}$ \Commentx{Store predictions for epoch $t$}
\ENDFOR
\STATE \textbf{return} $\mathcal{P}_i$
\end{algorithmic}
\end{algorithm}

\subsection{Applying Traditional Score to Object Detection}
\label{sec:traditional}

In dataset pruning for classification tasks, a common approach relies on model predictions, particularly logits~\cite{paul2021deep, pleiss2020identifying, toneva2019empirical}. 
To naturally extend this methodology to object detection, we leverage the logits from the matched predictions for each ground truth object, as described in Section~\ref{sec:defprediction}. 
Specifically, for each object \( j \) in image \( i \), we extract the logit from the corresponding prediction \( p_{ij} \in \mathcal{P}_i \) and compute the traditional classification score, assigning it as the object-level score \( s_{ij} \).
These object-level scores are then aggregated using the function \( \mathcal{A} \), as defined in Section~\ref{sec:aggscore}, to produce an image-level score analogous to those used for ranking or selection in classification tasks. 
This adaptation enables the direct application of classification-based pruning techniques to detection, thereby opening up new avenues for extending dataset pruning to more complex tasks.

\subsection{Variance-based Prediction Score}
\label{sec:VPS}
Object detection models, which are widely used, typically output the coordinates of bounding boxes, along with associated class probabilities and confidence scores. For a more detailed discussion on object detection models, refer to Appendix~\ref{sec:morerelatedwork}. 
In contrast to traditional scoring methods that focus primarily on class logits, we introduce a scoring mechanism that leverages the unique properties of object detection tasks. 
Recent studies~\cite{cho2025lightweightdatasetpruningtraining, he2024large} have highlighted that the variability in model predictions of class logits, especially across multiple training epochs, can provide a valuable indicator for pruning tasks. 
This variability is also relevant in object detection, where the nature of outputs—such as confidence scores and IoU—can vary significantly across different objects and training stages.

We define the \textbf{Variance-based Prediction Score (VPS)} to capture this variability in model outputs, tailored specifically for object detection. 
The score is formulated using the series of prediction values \( v^{(ij)} \), corresponding to object \( j \) in image \( i \), and is defined as follows:
\begin{equation}
\text{VPS}(v^{(ij)}) = \sqrt{\frac{1}{T} \sum_{k=1}^{T} \left( v_k^{(ij)} - \overline{v}^{(ij)} \right)^2}, \quad \overline{v}^{(ij)} = \frac{1}{T} \sum_{k=1}^{T} v_k^{(ij)}.
\end{equation}
Here, \( v_k^{(ij)} \in \{\mathrm{IoU}_k^{(ij)}, \mathrm{Conf}_k^{(ij)}\} \) represents the value of each output prediction in epoch \( k \), where \( \mathrm{IoU} \) refers to the Intersection over Union (IoU) score and \( \mathrm{Conf} \) refers to the confidence score. 
\( T \) is the total number of training epochs. The VPS is specifically designed for object detection tasks, where the outputs exhibit unique variations that are crucial for model evaluation and pruning.

\subsection{Analysis of Variance-based Prediction Score}
\label{sec:VPSAnalyze}

\begin{figure}[t]
    \centering
    \begin{subfigure}[b]{0.245\textwidth}
        \centering
        \includegraphics[width=\textwidth]{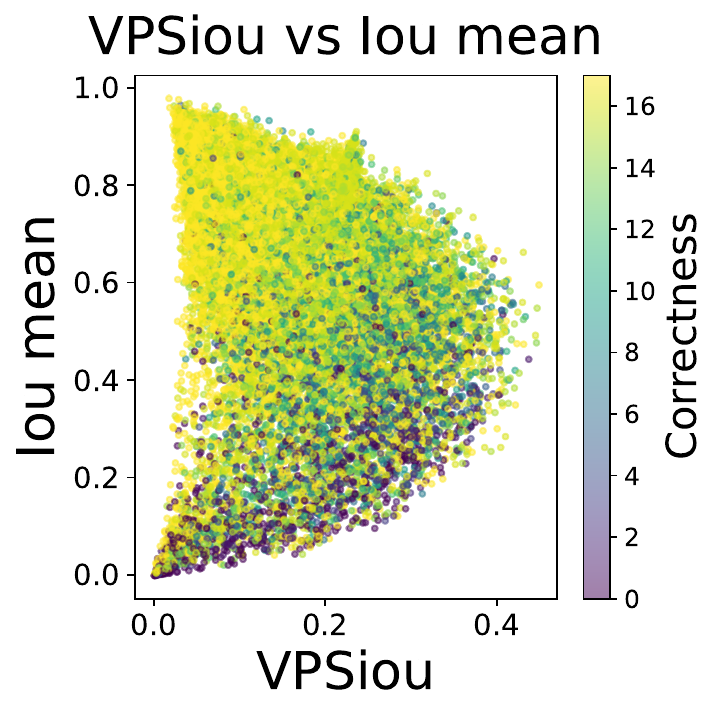}
        \caption{VPS\textsubscript{iou}, correctness}
        \label{fig:vps_scoresa}
    \end{subfigure}
    \begin{subfigure}[b]{0.245\textwidth}
        \centering
        \includegraphics[width=\textwidth]{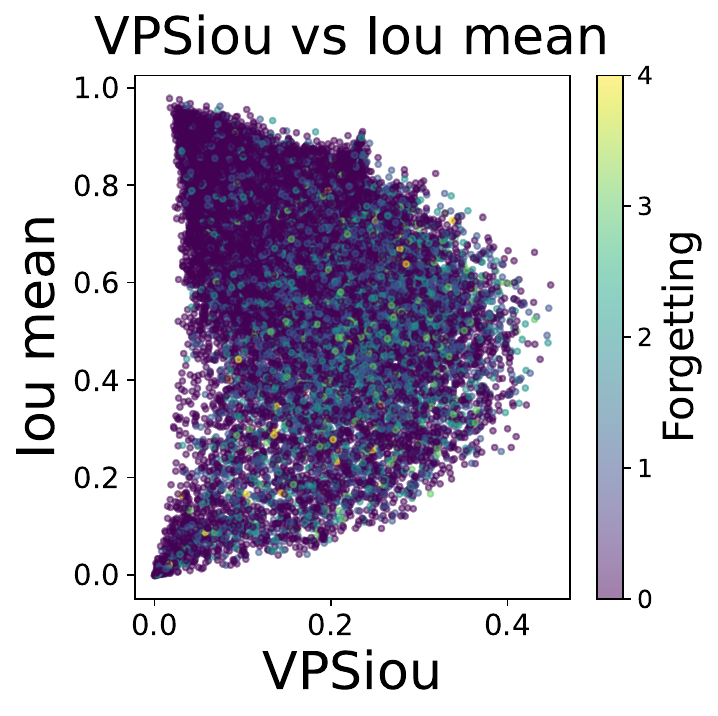}
        \caption{VPS\textsubscript{iou}, forgetting}
        \label{fig:vps_scoresb}
    \end{subfigure}
    \begin{subfigure}[b]{0.245\textwidth}
        \centering
        \includegraphics[width=\textwidth]{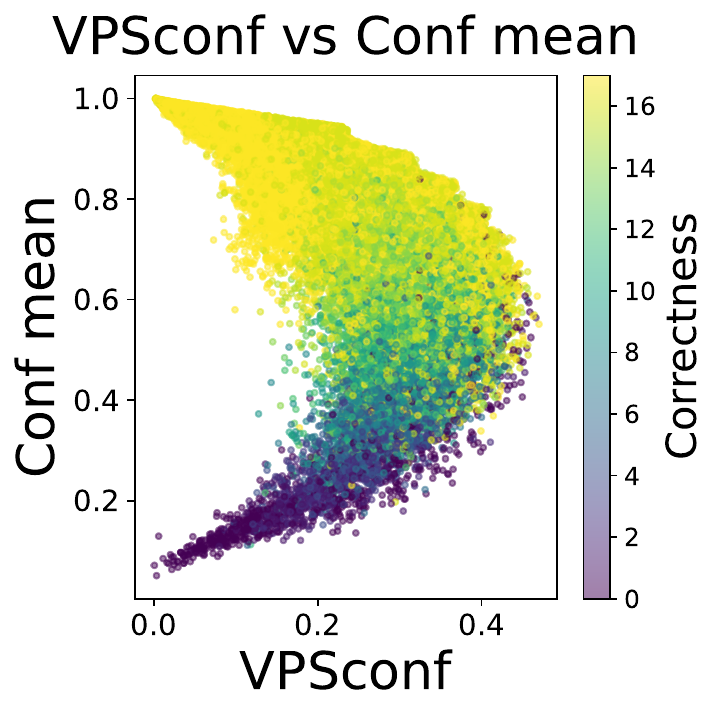}
        \caption{VPS\textsubscript{conf}, correctness}
        \label{fig:vps_scoresc}
    \end{subfigure}
    \begin{subfigure}[b]{0.245\textwidth}
        \centering
        \includegraphics[width=\textwidth]{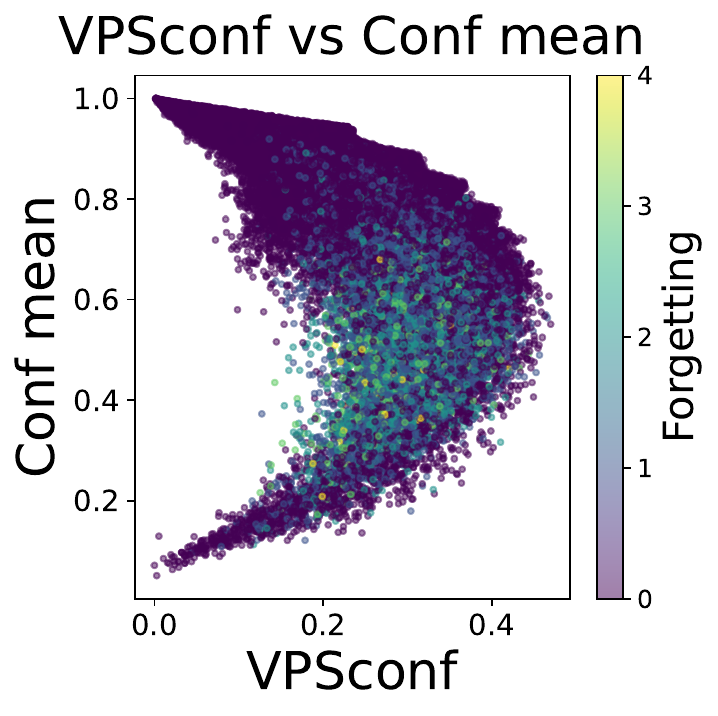}
        \caption{VPS\textsubscript{conf}, forgetting}
        \label{fig:vps_scoresd}
    \end{subfigure}
    \caption{
        Visualization of VPS scores (IoU, confidence) for object in PASCAL VOC dataset. The subplots display average score (y-axis) vs. standard deviation (x-axis), colored by correctness or forgetting events. (a, b) show VPS\textsubscript{iou}, and (c, d) show VPS\textsubscript{conf}. (a, c) are colored by correctness, while (b, d) are colored by forgetting events.
        }
    \label{fig:vps_scores}
\end{figure}

\begin{figure}[t]
    \centering
    \begin{subfigure}[b]{0.245\textwidth}
        \centering
        \includegraphics[width=\textwidth]{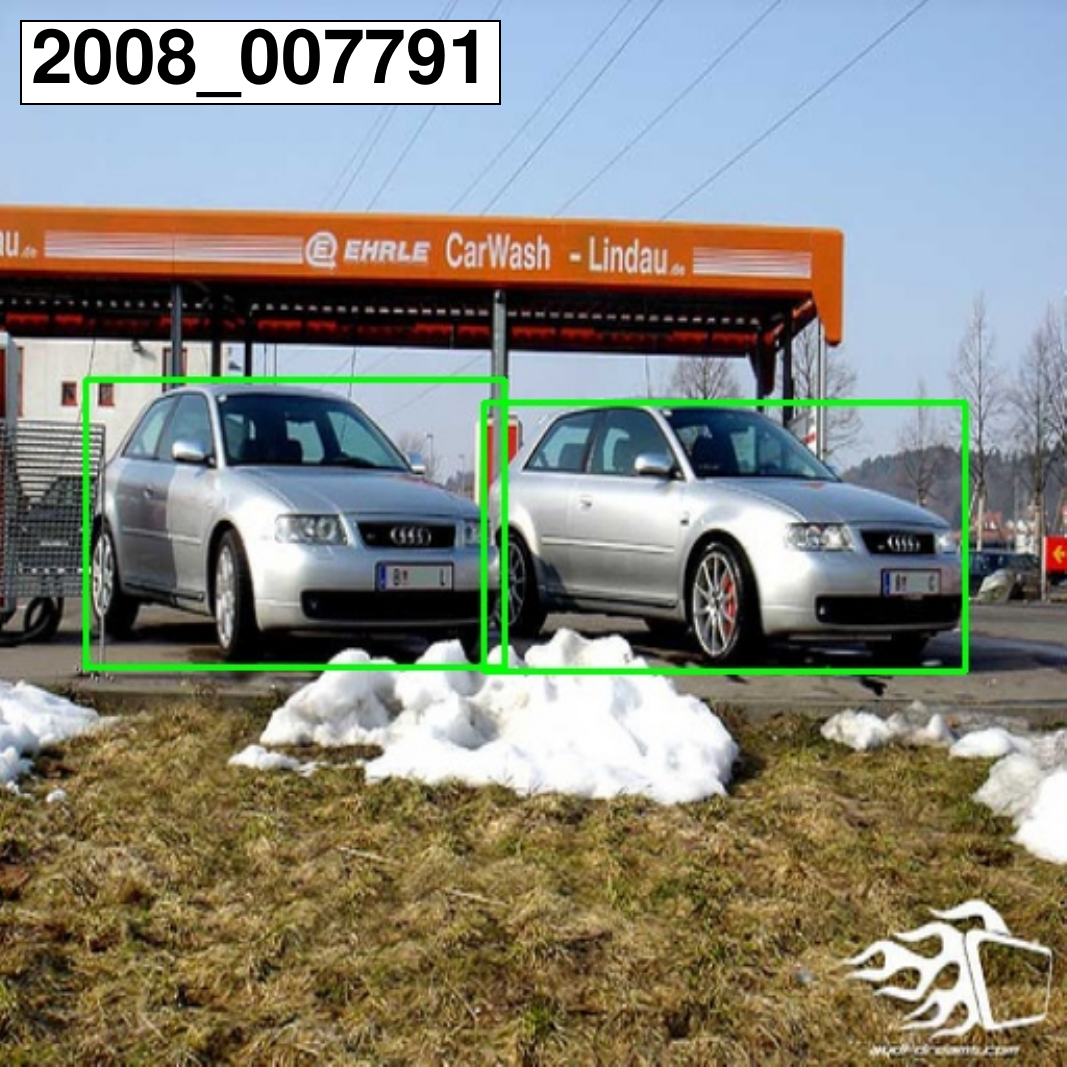}
        \caption{Easy Sample}
        \label{fig:four_imagesa}
    \end{subfigure}
    \begin{subfigure}[b]{0.245\textwidth}
        \centering
        \includegraphics[width=\textwidth]{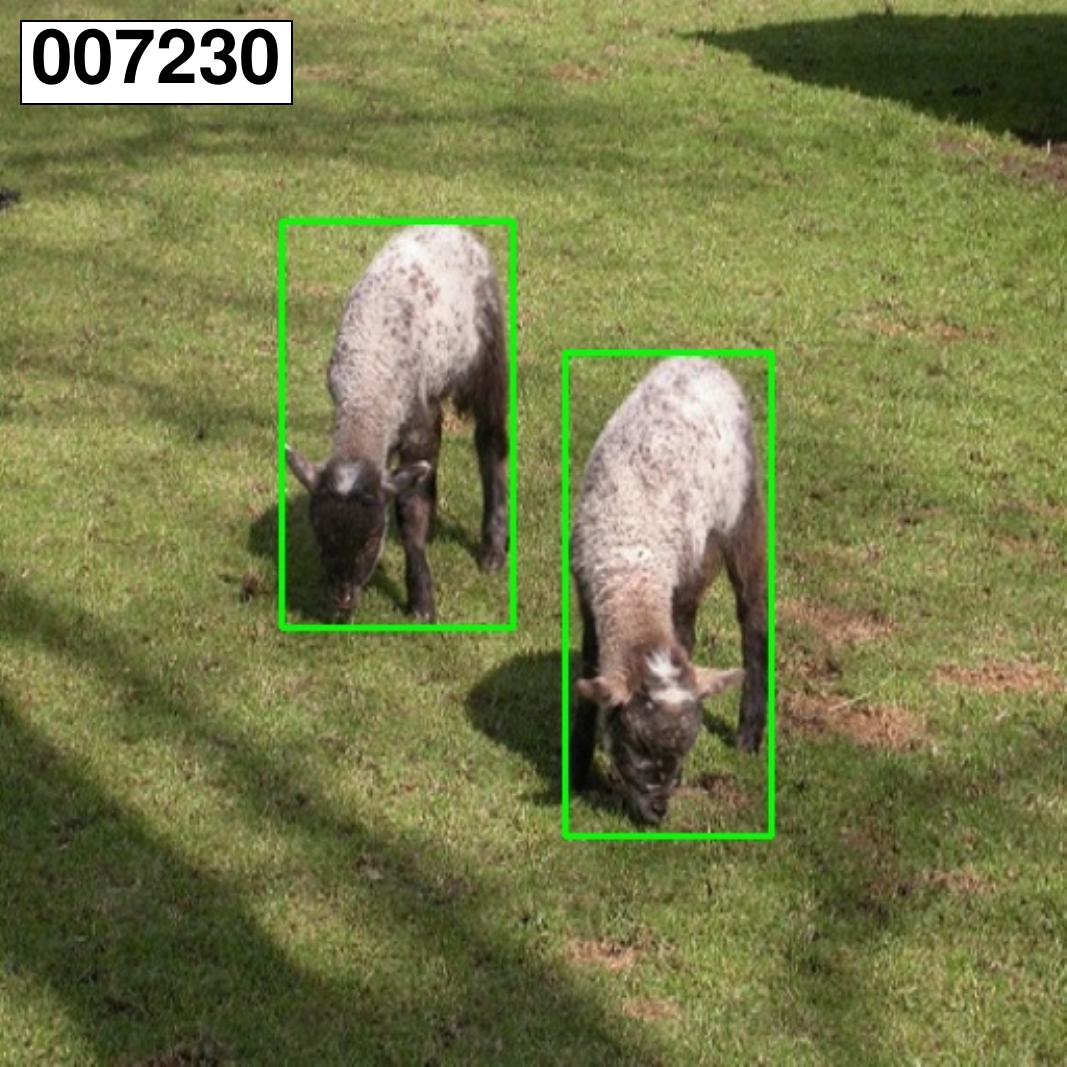}
        \caption{Easy Sample}
        \label{fig:four_imagesb}
    \end{subfigure}
    \begin{subfigure}[b]{0.245\textwidth}
        \centering
        \includegraphics[width=\textwidth]{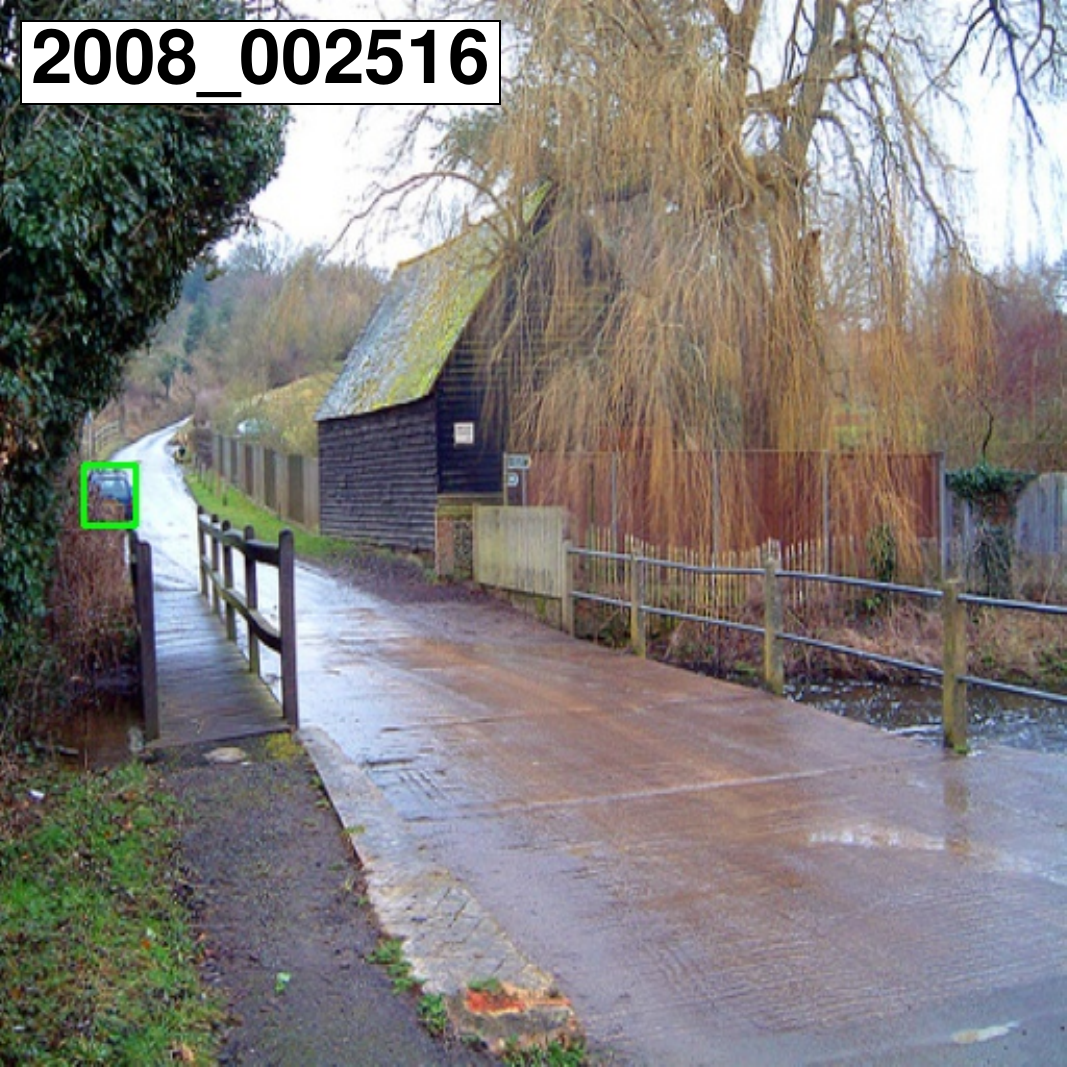}
        \caption{Hard Sample}
        \label{fig:four_imagesc}
    \end{subfigure}
    \begin{subfigure}[b]{0.245\textwidth}
        \centering
        \includegraphics[width=\textwidth]{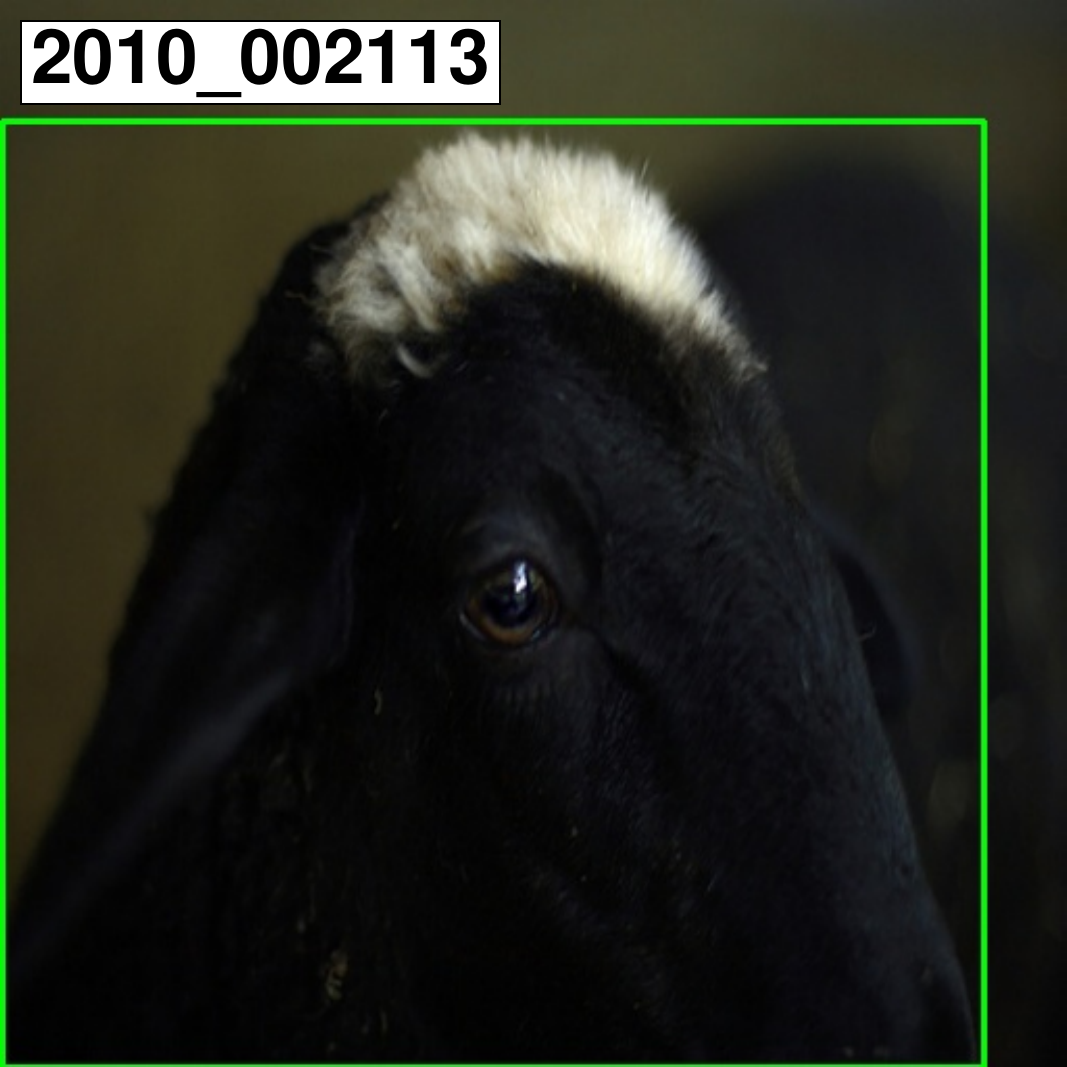}
        \caption{Hard Sample}
        \label{fig:four_imagesd}
    \end{subfigure}
    \caption{
        Pruned samples with ground truth bounding box based on low VPS\textsubscript{iou} scores using max aggregation from PASCAL VOC~\cite{everingham2010PASCAL}.
        (a, b) show examples with high IoU\textsubscript{mean} scores (easy), where foreground and background are clearly separable.
        (c, d) show difficult cases with low IoU\textsubscript{mean}; (c) has a distant car, (d) a black sheep head against a dark background.
    }
    \label{fig:four_images}
\end{figure}

In this subsection, we further analyze the prediction scores proposed in the previous part. 
Figure~\ref{fig:vps_scores} presents the mean and variance of time-series predictions (IoU and confidence) for each object, colored by correctness and Forgetting scores~\cite{toneva2019empirical}. 
The correctness score denotes the number of correct class predictions over $T$ training epochs.
Notably, consistent with prior studies~\cite{cho2025lightweightdatasetpruningtraining, he2024large, swayamdipta2020dataset}, we observe the characteristic “moon-shaped” distribution when visualizing these metrics—previously reported in logit-space—now emerging in the context of IoU and confidence predictions.

From Figures~\ref{fig:vps_scoresa} and~\ref{fig:vps_scoresc}, which are colored by correctness, we observe that low VPS scores correspond to samples that are either consistently misclassified (i.e., hard examples) or consistently correctly classified (i.e., easy examples). 
In contrast, high VPS scores are predominantly associated with samples of intermediate difficulty.
Furthermore, Figures~\ref{fig:vps_scoresb} and~\ref{fig:vps_scoresd} reveal an overlap between Forgetting scores and intermediate-to-high VPS values, with this pattern being more pronounced in Figure~\ref{fig:vps_scoresd}. 
This suggests that VPS offers a continuous and informative perspective on a traditionally discrete Forgetting metric~\cite{toneva2019empirical}.
To qualitatively support these observations, Figure~\ref{fig:four_images} illustrates four pruned examples of low VPS samples: the easy examples (Figures~\ref{fig:four_imagesa} and~\ref{fig:four_imagesb}) contain clearly visible and easily detectable objects, while the hard examples (Figures~\ref{fig:four_imagesc} and~\ref{fig:four_imagesd}) feature small or occluded objects, further highlighting the relationship between VPS and detection difficulty. For additional visualizations, please refer to Appendix~\ref{app:visu}.

\subsection{Image-Level Aggregation Problem}
\label{sec:aggscore}

After assigning scores to individual objects, the final challenge is to determine a representative score for the entire image based on these object-level scores. 
To address this, we propose a simple yet effective solution: aggregating object-level scores using a statistical function. The image-level score \( S_i \) for image \( i \) is defined as follows:
\begin{equation}
S_i = \mathcal{A}(s_{i1}, s_{i2}, \dots, s_{in_i}).
\end{equation}
Here, \( \mathcal{A} \) is an aggregation function such as the \textbf{mean} (arithmetic average), the \textbf{sum} (total score), or the \textbf{max} (maximum score), 
among others that can summarize the object-level scores. In this equation, \( s_{ij} \) denotes the score assigned to object \( j \), and \( n_i \) represents the total number of objects in image \( i \).
This formulation provides a flexible yet interpretable way to summarize object-level scores into a single image-level score, making it adaptable for use in traditional classification tasks.

\section{Experiment}
In this section, we provide a comprehensive evaluation of our proposed pruning strategies on PASCAL VOC~\cite{everingham2010PASCAL} and MS COCO~\cite{lin2014microsoft}, as described in Section~\ref{sec:mainresults}. 
Furthermore, we analyze the effects of aggregation techniques (Section~\ref{sec:aggregation}), generalization across architectures (Section~\ref{sec:crossarch}), annotation density (Section~\ref{sec:analannotation}), and class imbalance (Section~\ref{sec:analclass}) on pruning effectiveness.

\subsection{Baseline}
\label{sec:baseline}
We compare our method with several established score-based pruning techniques, including Random, AUM~\cite{pleiss2020identifying},  Entropy~\cite{coleman2019selection}, EL2N~\cite{paul2021deep}, and Forgetting~\cite{toneva2019empirical}. We also include a loss-based baseline, where images are pruned based on their loss values from a well-trained model.
To ensure a fair comparison, we introduce \textbf{Instance-Density Pruning (IDP)}, which prioritizes images by object count, resolving ties (i.e., instances with the same count) randomly. 
Additionally, we include two baselines that prioritize harder samples with lower average IoU or confidence scores, serving as references for evaluating the effectiveness of our proposed VPS method.

\subsection{Implementation Details and Experiment Setup}
\label{sec:implementationdetailsinmain}
\paragraph{Collecting Statistics.}
To collect data statistics for each data point throughout the training process, we use Faster R-CNN-C4~\cite{ren2015faster} with a ResNet-50~\cite{he2016deep} backbone. 
This choice is driven by its compatibility with prior work~\cite{lee2024coreset} and its balanced trade-off between accuracy and computational efficiency. 
Our implementation follows the default configuration provided by Detectron2~\cite{wu2019detectron2}.

\paragraph{Scores and Training Setup.}
To ensure fairness, we use an equal number of statistics per object instance when calculating scores. 
The number of training epochs used for score calculation is set to 17 for VOC and 12 for COCO.
All baselines follow similar procedures, except for EL2N, which uses the first 10 epochs~\cite{paul2021deep}. 
After calculating per-object scores, we apply an aggregation function to obtain a final score for an image. 
In the main paper, results are primarily reported using max aggregation, with mAP, mAP@50, and mAP@75 as evaluation metrics. 
Training iterations are adjusted based on the pruning ratio to ensure fair comparisons. 
Further implementation details, including hyperparameters, pruning settings, and dataset information, are provided in Appendix~\ref{app:implementationdetails}.

\subsection{Main Results}
\label{sec:mainresults}
\paragraph{PASCAL VOC}

\begin{table}[t]
\centering
\caption{Comparison of traditional and proposed pruning methods in terms of mAP, mAP@75, and mAP@50 across various pruning ratios on PASCAL VOC~\cite{everingham2010PASCAL}. Scores are aggregated using the max function.
The best result is shown in \textbf{bold}, and the second best is \underline{underlined}.
Performance on the full dataset is shown in parentheses in the top row.
For the definitions of IDP, IoU, and Confidence, please refer to Section~\ref{sec:baseline}.}
\resizebox{\textwidth}{!}{
\begin{tabular}{lcccccccccccccc}
\toprule
\multirow{2}{*}{Method} & \multicolumn{4}{c}{mAP (\%) (\textit{52.16})} & \multicolumn{4}{c}{map@75 (\%) (\textit{56.92})} & \multicolumn{4}{c}{map@50 (\%) (\textit{80.63})} \\
\cmidrule(lr){2-5} \cmidrule(lr){6-9} \cmidrule(lr){10-13}
& 30\% & 50\% & 70\% & 90\% & 30\% & 50\% & 70\% & 90\% & 30\% & 50\% & 70\% & 90\% \\
\midrule
Random & 48.87 & 45.98 & 40.81 & 28.09      & 51.75 & 48.21 & 40.41 & 22.53       & 78.86 & 76.53 & 72.82 & 59.05 \\
IDP & 49.55 & 46.47 & 40.93 & 27.88         & 52.97 & 49.02 & 39.97 & 23.61       & 79.21 & 77.17 & 72.27 & 56.11 \\
Loss & 48.82 & 45.56 & 40.60 & 28.98        & 52.54 & 47.51 & 40.28 & 23.60       & 78.54 & 76.09 & 71.10 & 59.58\\
AUM~\cite{pleiss2020identifying} & 48.17 & 43.52 & 35.27 & 16.98          & 50.85 & 44.24 & 32.30 & 10.88       & 78.67 & 75.12 & 66.44 & 39.66 \\
Entropy~\cite{coleman2019selection} & 49.27 & 45.84 & 39.76 & 24.66       & 52.35 & 47.24 & 38.39 & 18.57       & 79.29 & 77.52 & 71.92 & 54.12 \\
EL2N~\cite{paul2021deep} & 49.56 & 45.72 & 39.96 & 27.24                  & 53.33 & 46.96 & 38.79 & 21.98       & \underline{79.67} & 77.38 & 72.81 & 56.19 \\
Forgetting~\cite{toneva2019empirical} & 48.96 & 46.34 & \underline{41.71} & 28.14       & 52.03 & 48.28 & \underline{40.88} & 21.99       & 78.96 & 77.37 & \underline{74.04} & \underline{59.61} \\
IoU & 48.64 & 43.93 & 36.39 & 16.50             & 52.21 & 45.35 & 33.24 & 8.82        & 79.17 & 76.13 & 69.73 & 42.38 \\
Confidence & 47.78 & 43.09 & 33.80 & 15.97      & 50.31 & 43.14 & 30.82 & 10.41       & 78.42 & 74.77 & 64.00 & 37.32 \\
\midrule
\textbf{VPS\textsubscript{conf} (Ours)} & \underline{49.78} & \underline{46.65} & 40.99 & \underline{29.25}       & \textbf{53.36} & \underline{49.55} & 39.88 & \underline{24.21} &       \textbf{79.69} & \textbf{77.72} & 73.10 & 59.45 \\
\textbf{VPS\textsubscript{iou} (Ours)} & \textbf{49.81} & \textbf{47.08} & \textbf{42.75} & \textbf{30.05}        & \underline{53.33} & \textbf{49.77} & \textbf{43.02} & \textbf{25.84}       & 79.45 & \underline{77.54} & \textbf{74.17} & \textbf{60.77} \\
\bottomrule
\end{tabular}
}
\label{tab:voc}
\end{table}

We evaluated the effectiveness of our proposed pruning methods, VPS\textsubscript{iou} and VPS\textsubscript{conf}, on the PASCAL VOC dataset~\cite{everingham2010PASCAL} across various pruning ratios, 30\%, 50\%, 70\%, and 90\%. 
Table~\ref{tab:voc} shows a detailed comparison with several baseline methods.
The primary evaluation metric is averaged mAP, which summarizes detection performance by averaging IoU thresholds from 0.5 to 0.95 in 0.05 increments.

Our results show that both VPS-based criteria consistently outperform competing methods across all metrics. 
In particular, VPS\textsubscript{iou} achieved the highest mAP and mAP@75 at all pruning ratios, except for mAP@75 at the 30\% pruning ratio. 
Notably, VPS\textsubscript{iou} surpassed all baselines by over one percentage point at a very high pruning ratio (90\%), demonstrating its robustness under extreme compression. 
Meanwhile, VPS\textsubscript{conf} performed best in mAP@50, highlighting its advantage under relaxed IoU thresholds.

Among the baselines, Forgetting and EL2N delivered competitive performance.  
Notably, Forgetting ranked second in all metrics at the 70\% pruning level, demonstrating robustness under moderate pruning.  
EL2N achieved the second-best mAP@50 at the 30\% pruning level, suggesting its utility when minimal pruning is required.  
In contrast, AUM, IoU, and Confidence suffered notable performance drops under high pruning ratios ($\geq$70\%). 
In particular, using average IoU or confidence as selection criteria resulted in a 5--15\% mAP drop compared to our proposed methods.
These findings challenge the common assumption that harder examples are inherently more informative for training, and underscore the importance of accounting for variance in object-level characteristics when designing pruning strategies.

\paragraph{MS COCO}

To further assess the effectiveness of our proposed pruning strategies, we conducted experiments on the MS COCO dataset. 
Due to the larger scale of COCO compared to VOC, we limited our comparison to representative baseline methods: Random, IDP, and two strong baselines (EL2N and Forgetting) that demonstrated strong performance on the VOC benchmark.
Table~\ref{tab:coco} presents the results of our methods (VPS\textsubscript{iou} and VPS\textsubscript{conf}) compared to several baselines across pruning ratios of 60\%, 70\%, 80\%, and 90\%. 

In terms of the most important metric, mAP, both VPS\textsubscript{iou} and VPS\textsubscript{conf} consistently achieved the top performance across all pruning levels.  
Notably, VPS\textsubscript{conf} slightly outperformed VPS\textsubscript{iou} in mAP, although the margin was generally less than one percent, suggesting that both criteria are comparably effective.  
The gap was more noticeable in mAP@75, with our proposed methods consistently achieving higher performance than traditional scoring methods.

One notable finding is the performance of the Forgetting score at high pruning rates in terms of mAP@50.
This result suggests that the Forgetting metric remains a strong indicator of sample importance even in large-scale object detection tasks.  
However, its main limitation lies in its discrete-valued nature: samples with the same score are treated identically, which forces the selection process to fall back on random sampling. 
This lack of granularity may hinder precise ranking, potentially compromising the effectiveness of score-based pruning.

\begin{table}[t]
\centering
\caption{Comparison of traditional and proposed pruning methods in terms of mAP, mAP@75, and mAP@50 across various pruning ratios on MS COCO~\cite{lin2014microsoft}. 
The best result is shown in \textbf{bold}, and the second best is \underline{underlined}.
Performance on the full dataset is shown in parentheses in the top row.}
\resizebox{\textwidth}{!}{
\begin{tabular}{lcccccccccccccc}
\toprule
\multirow{2}{*}{Method} & \multicolumn{4}{c}{mAP (\%) (\textit{35.55})} & \multicolumn{4}{c}{map@75 (\%) (\textit{37.92})} & \multicolumn{4}{c}{map@50 (\%) (\textit{	55.81})} \\
\cmidrule(lr){2-5} \cmidrule(lr){6-9} \cmidrule(lr){10-13}
& 60\% & 70\% & 80\% & 90\% & 60\% & 70\% & 80\% & 90\% & 60\% & 70\% & 80\% & 90\% \\
\midrule
Random & 30.94 & 29.07 & 26.37 & 21.56 & 32.50 & 30.17 & 26.63 & 20.46 & 51.39 & 49.13 & 46.32 & 40.83 \\
IDP & 31.26 & 28.85 & 25.25 & 19.28 & 32.72 & 30.05 & 25.62 & 18.04 & 51.74 & 46.68 & 44.58 & 36.94 \\
EL2N~\cite{paul2021deep} & 30.85 & 28.81 & 25.49 & 20.03 & 32.00 & 29.63 & 25.64 & 18.54 & 52.11 & 49.52 & 45.64 & 38.84 \\
Forgetting~\cite{toneva2019empirical} & 31.40 & 29.62 & 26.90 & 21.93 & 33.06 & 30.59 & 27.33 & 20.64 & \underline{52.43} & \textbf{50.54} & \textbf{47.52} & \textbf{41.66} \\
\midrule
\textbf{VPS\textsubscript{conf} (Ours)} & \textbf{31.74} & \textbf{30.00} & \textbf{27.31} & \underline{22.45} & \underline{33.18} & \textbf{31.46} & \textbf{28.26} & \underline{21.93} & 52.00 & 50.16 & 46.80 & 41.26 \\
\textbf{VPS\textsubscript{iou} (Ours)} & \underline{31.72} & \underline{29.89} & \underline{27.30} & \textbf{22.49} & \textbf{33.43} & \underline{31.25} & \underline{27.90} & \textbf{22.08} & \textbf{52.46} & \underline{50.23} & \underline{47.44} & \underline{41.39} \\
\bottomrule
\end{tabular}
}
\label{tab:coco}
\end{table}

\begin{figure}[t]
  \begin{minipage}{0.50\textwidth}
    \centering
    \begin{subfigure}{0.5\linewidth}
      \centering
      \includegraphics[width=\linewidth]{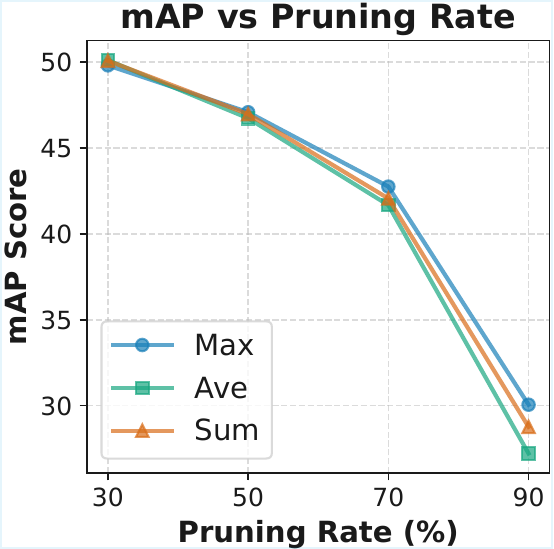}
      \caption{VPS\textsubscript{iou}}
    \end{subfigure}\hfill
    \begin{subfigure}{0.5\linewidth}
      \centering
      \includegraphics[width=\linewidth]{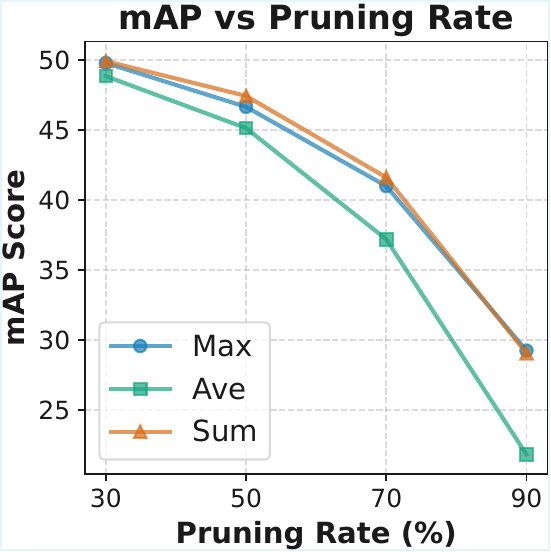}
      \caption{VPS\textsubscript{conf}}
    \end{subfigure}
    \caption{mAP comparison of different aggregation methods (max, average, sum) on the PASCAL VOC~\cite{everingham2010PASCAL} dataset across varying pruning rates.}
    \label{fig:compagg}
  \end{minipage}\hfill
  \begin{minipage}{0.49\textwidth}
    \centering
    \captionof{table}{Cross-architecture evaluation of mAP on PASCAL VOC~\cite{everingham2010PASCAL} using YOLOv5m~\cite{yolov5}. The best result is highlighted in \textbf{bold}, and the second best is \underline{underlined}.}
    \label{tab:cross-arch}
    \begin{tabular}{lcccc}
      \toprule
      \multirow{2}{*}{Method} & \multicolumn{4}{c}{mAP (\%)} \\
      \cmidrule(lr){2-5}
                              & 30\%  & 50\%  & 70\%  & 90\%  \\
      \midrule
      Random                  & 51.52 & 47.61 & \underline{39.51} & \underline{22.87} \\
      IDP                     & 51.86 & \textbf{48.13} & 39.22 & 22.57 \\
      EL2N                    & \underline{51.98} & 46.90 & 36.83 & 18.44 \\
      Forgetting              & 51.37 & 47.04 & 39.05 & 20.37 \\
      \midrule
      \textbf{VPS\textsubscript{conf}} & \textbf{52.03} & 47.65 & 39.06 & 21.06 \\
      \textbf{VPS\textsubscript{iou}}  & 51.91 & \underline{48.05} & \textbf{40.73} & \textbf{23.08} \\
      \bottomrule
    \end{tabular}
  \end{minipage}
\end{figure}

\subsection{Analysis of the Aggregation Method}
\label{sec:aggregation}
We present additional results using the \textit{average} and \textit{sum} aggregation methods, as summarized in Figure~\ref{fig:compagg}.
VPS\textsubscript{iou} achieves the best performance with \textit{max} under high pruning rates, while remaining competitive with \textit{sum} and \textit{average} at lower pruning levels.
In contrast, VPS\textsubscript{conf} performs poorly with \textit{average} but shows a significant improvement when using \textit{sum} and \textit{max}.
As shown in Table~\ref{tab:compagg} in Appendix~\ref{app:agg}, the superiority of our proposed methods remains consistent regardless of the aggregation method used.
Notably, these results suggest that the effect of the aggregation method varies based on the scoring function.
Full results can be found in Table~\ref{tab:compagg} in Appendix~\ref{app:agg}.

\subsection{Cross Architecture Evaluation}
\label{sec:crossarch}
To evaluate generalization across architectures, we employed YOLOv5m~\cite{yolov5}, a representative object detection model with a one-stage architecture, distinct from Faster R-CNN.
YOLOv5m was selected for its balance between inference speed and detection performance.
In this evaluation, we reused the data points selected by Faster R-CNN with YOLOv5m.
As shown in Table~\ref{tab:cross-arch}, while improvements were less pronounced than with Faster R-CNN, the VPS\textsubscript{iou}-based selection method consistently outperformed the Random baseline.
These results indicate that our approach maintains generalizability across model architectures, though it may also be seen as a limitation.

\begin{figure}[t]
  \centering
  \includegraphics[width=\linewidth]{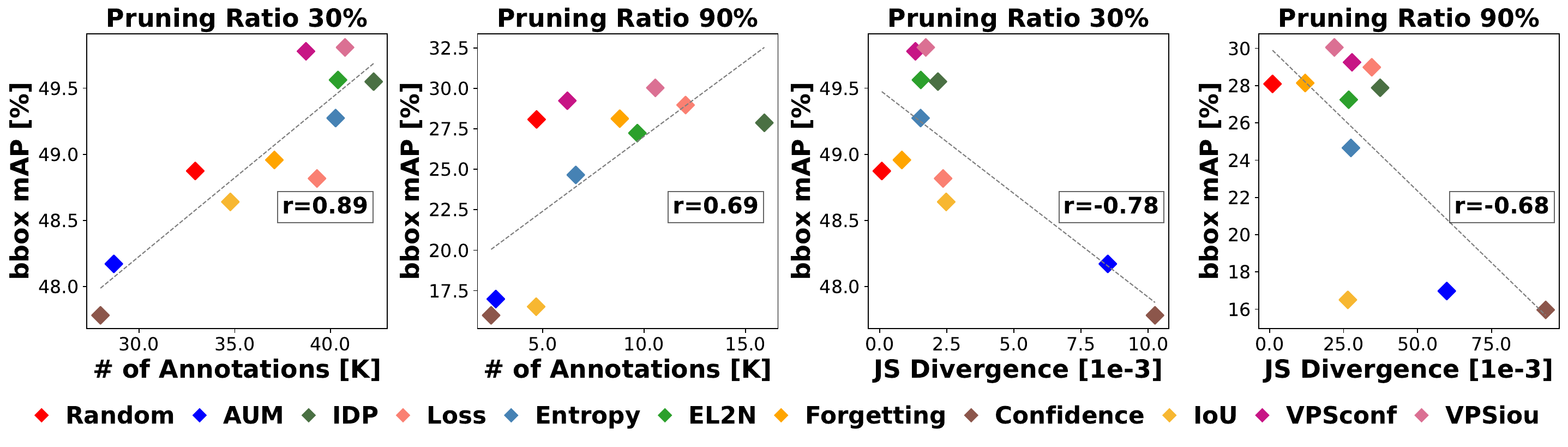}
  \caption{
  Plots showing averaged mAP vs. number of annotations and JS Divergence under high (90\%) and low (30\%) pruning on PASCAL VOC, with correlation coefficients (r) annotated.
  }
  \label{fig:relation}
\end{figure}

\subsection{Analysis of the Number of Annotations}
\label{sec:analannotation}
As shown in Table~\ref{tab:voc} and Table~\ref{tab:coco}, the IDP baseline introduced earlier showed limited effectiveness at high pruning ratios (70\%--90\%), but consistently outperformed random pruning under pruning ratios of 60\% or lower.  
Furthermore, as illustrated in Figure~\ref{fig:relation}, there is a strong positive correlation between the number of annotations and detection accuracy, which is particularly prominent under low pruning ratios with a coefficient of 0.89, while it becomes approximately 0.69 under high pruning ratios.
However, it is also noteworthy that IDP does not always achieve the best performance, indicating that increasing the number of annotations alone does not guarantee improved results; rather, the careful selection of informative samples is crucial for enhancing performance in object detection.

\subsection{Analysis of Class Distribution Shift}
\label{sec:analclass}
We analyze how shifts in class distribution between the original and pruned datasets affect detection performance.
To quantify this shift, we use the Jensen-Shannon (JS) Divergence.
For each pruning ratio, we computed the normalized class frequencies of the pruned datasets and calculated the JS Divergence relative to the original dataset. The results are shown in Figure~\ref{fig:relation}.
Our findings reveal a negative correlation between JS Divergence and detection accuracy, with correlation values of 0.78 under low pruning ratios and 0.68 under high ratios.
However, compared to the nearly-zero JS value of Random, our proposed method shows relatively larger JS values. 
This suggests that smaller shifts in class distribution are not necessarily prioritized factors for improving detection performance.

\section{Conclusions}
\label{sec:conclusions}

In this paper, we defined dataset pruning for object detection by identifying and addressing Object-Level Attribution, Scoring Strategy, and Image-Level Aggregation problems. 
We proposed the Variance-based Prediction Score (VPS), which selects informative training examples by combining IoU and confidence score variance. 
Our experiments on PASCAL VOC and MS COCO show that our approach outperforms existing pruning methods in terms of mean Average Precision (mAP) and generalizes well across different detection architectures. We also found that while annotation count and class distribution shift impact performance, the key factor is selecting informative examples rather than focusing on dataset size or balance. This work bridges pruning for classification and object detection, enabling efficient training in complex vision tasks, fostering research, and promoting sustainable AI with social impact via reduced resource use.

\newpage

\newpage

\appendix

\section{Dataset and Implementation Details}
\label{app:implementationdetails}

\subsection{Datasets}
We employ two widely used benchmark datasets for object detection: PASCAL VOC 2007+2012~\cite{everingham2010PASCAL} and MS COCO~\cite{lin2014microsoft}. 
For PASCAL VOC, we use the combined trainval sets from 2007 and 2012 for training, which include 20 categories and 16,551 datapoints. Evaluation is conducted on the VOC 2007 test set, which contains 4,952 images.
The MS COCO dataset consists of 80 object categories, with 118,287 training images and 5,000 validation images.
For the pruning experiments, we use 117,264 training images with annotations as the base set.
The storage sizes of each dataset are shown in Table~\ref{tab:oddataset}, clearly highlighting the sizes of images and annotations.
This supports our policy of prioritizing pruning at the image level rather than individual objects.

\begin{table}[htbp]
  \centering
  \caption{Storage size of images and annotations in main object detection datasets.}
  \label{tab:oddataset}
  \begin{tabular}{lcc}
    \hline
    Dataset & Images & Annotations \\
    \hline
    VOC2007 & 0.86\,GB & 0.040\,GB \\
    VOC2012 & 1.9\,GB & 0.068\,GB \\
    COCO2017 & 25.3\,GB & 0.47\,GB \\
    \hline
  \end{tabular}
\end{table}

\subsection{Evaluation Metrics} 
Following standard protocol, we report detection performance using Average Precision (AP), averaged over multiple IoU thresholds ranging from 0.5 to 0.95 in steps of 0.05. We additionally report AP at IoU = 0.5 (map@50) and IoU = 0.75 (map@75).

\subsection{Computational Resources}
\label{sec:computational}
All experiments were conducted using two NVIDIA GeForce RTX 4090 GPUs with 125 GiB of system memory. For Faster R-CNN, parallel computation was performed across both GPUs. YOLOv5m was trained using a single GPU. In the case of Faster R-CNN, training on the full PASCAL VOC dataset took approximately 4 hours, while training on the COCO dataset took about one day. For YOLOv5m, training on the full PASCAL VOC dataset took approximately one day.

\subsection{Models and Hyperparameter Settings}

\begin{table}[htbp]
\centering
\caption{Iteration schedules for PASCAL VOC~\cite{everingham2010PASCAL} and MS COCO~\cite{lin2014microsoft} datasets with different pruning ratios.}
\begin{tabular}{l l c c c c c}
\toprule
Dataset & Pruning Ratio       & Full   & 30\%     & 50\%     & 70\%     & 90\%     \\ \midrule
\multirow{3}{*}{PASCAL VOC} & BASE\_MAX\_ITER & 18{,}000 & 12{,}600 & 9{,}000  & 5{,}400  & 1{,}800  \\ 
                            & BASE\_STEP1      & 12{,}000 & 8{,}400  & 6{,}000  & 3{,}600  & 1{,}200  \\ 
                            & BASE\_STEP2      & 16{,}000 & 11{,}200 & 8{,}000  & 4{,}800  & 1{,}600  \\ \midrule
Dataset & Pruning Ratio       & Full   & 60\%     & 70\%     & 80\%     & 90\%     \\ \midrule
\multirow{3}{*}{MS COCO}    & BASE\_MAX\_ITER & 90{,}000 & 36{,}000 & 27{,}000 & 18{,}000 & 9{,}000  \\ 
                            & BASE\_STEP1      & 60{,}000 & 24{,}000 & 18{,}000 & 12{,}000 & 6{,}000  \\ 
                            & BASE\_STEP2      & 80{,}000 & 32{,}000 & 24{,}000 & 16{,}000 & 8{,}000  \\ \bottomrule
\end{tabular}
\label{tab:iterations_pruning}
\end{table}

We adopt Faster R-CNN with the C4 architecture~\cite{ren2015faster} and a ResNet-50 backbone~\cite{he2016deep} to collect data statistics.  
Training is performed using the default configuration of the Detectron2 framework~\cite{wu2019detectron2}, with 18,000 iterations and a batch size of 16 for PASCAL VOC, and 90,000 iterations with the same batch size for MS COCO.  
For experiments evaluating model performance using pruned data, we adjust the number of training iterations in proportion to the reduction in training dataset size.  
Specifically, the number of iterations is linearly reduced to maintain an equivalent number of epochs compared to the default setting.  
Table~\ref{tab:iterations_pruning} provides further details on the training schedules.  
This approach ensures that the model undergoes a comparable number of updates per epoch, enabling fair comparisons between the full and pruned datasets.

To evaluate cross-architecture generalization, we utilize YOLOv5, a widely adopted object detection framework.  
YOLOv5 is available in five variants: YOLOv5n (nano), YOLOv5s (small), YOLOv5m (medium), YOLOv5l (large), and YOLOv5x (extra-large).  
Among these, we select YOLOv5m, which offers a balanced trade-off between inference speed and detection accuracy.  
We set the default number of epochs to 300 with a batch size of 32.  
When using pruned data, the number of epochs is linearly adjusted to maintain a comparable total number of training steps.  
All other hyperparameters follow the official Ultralytics implementation~\cite{yolov5}.

\subsection{Details of Score Calculation}
This section provides detailed descriptions of how each score is computed. The ranking of data points for pruning is fundamentally based on their estimated difficulty. Please also refer to Section~\ref{sec:implementationdetailsinmain} for additional implementation details.

\textbf{Random}: Image IDs are randomly selected without any specific criterion.

\textbf{Instance-Density Pruning (IDP)}: Images are ranked based on the number of annotated instances they contain, in descending order. Images with the same number of objects are randomly sampled.

\textbf{Loss}: The total loss is computed for each image using a well-trained model. This loss serves as the score for ranking, where images with higher loss values are prioritized.

\textbf{AUM}~\cite{pleiss2020identifying}: The Area Under the Margin (AUM) is calculated using the logits output at each epoch. The score is defined as the average margin between the logit corresponding to the ground truth class and the highest non-ground-truth logit over a predefined number of epochs. As in previous work~\cite{zheng2022coverage}, we use probabilities instead of raw logits for implementation. Images are ranked in ascending order of AUM.
The logits used here correspond to the predictions defined in Section~\ref{sec:defprediction}.

\textbf{Entropy}~\cite{coleman2019selection}: Entropy is computed from the model's logits, following the same procedure as AUM. Images are ranked in descending order of entropy values.

\textbf{EL2N}~\cite{paul2021deep}: Similar to AUM and Entropy, EL2N scores are derived from the logits. In accordance with prior studies~\cite{paul2021deep}, only the first 10 epochs are used for score calculation. Images are ranked in descending order of EL2N scores.

\textbf{Forgetting}~\cite{toneva2019empirical}: A forgetting event is defined as a transition where the model's prediction changes from correct to incorrect across training epochs. The number of forgetting events is counted for each image and used as the score. Images are ranked in descending order. Due to the inherent nature of this metric, the scores are always discrete integer values. In cases where multiple images share the same score, selection is performed randomly.

\textbf{IoU}: For each predicted object, we compute the Intersection over Union (IoU) using the predicted bounding box coordinates defined in Section~\ref{sec:defprediction}. The IoU is averaged over the selected epochs to obtain the final score. Images are ranked in ascending order of this averaged IoU.

\textbf{Confidence}: Similar to IoU, the model’s confidence value for each predicted object is averaged over the selected epochs. Images are ranked in ascending order of the average confidence.

\textbf{VPS\textsubscript{iou} (Ours)}: For each predicted object, we compute the variance of the IoU values across selected epochs. Images are ranked in descending order of IoU variance.

\textbf{VPS\textsubscript{conf} (Ours)}: Following the same logic as VPS\textsubscript{iou}, we compute the variance of the confidence values across selected epochs. Images are ranked in ascending order of this variance.

\vspace{1em}
\noindent
\textbf{Note on Training Dynamics-Based Methods:}  
Recent studies have introduced methods based on training dynamics~\cite{cho2025lightweightdatasetpruningtraining, he2024large}, which compute scores over a window of several epochs (typically 10) and use the average as the final score. 
However, object detection tasks are computationally expensive, and collecting long training trajectories (spanning hundreds of epochs) is infeasible in our environments due to excessive time and storage requirements. 
For instance, as discussed in Section~\ref{sec:computational}, even training Faster R-CNN with two GPUs under the default settings takes approximately four hours on PASCAL VOC and about one day on MS COCO.
The number of available epochs is limited to at most 18 for PASCAL VOC and 13 for MS COCO.
Due to these practical constraints, we do not include training dynamics-based methods in our experiments.


\section{Related Work on Object Detection}
\label{sec:morerelatedwork}
\begin{figure}[t]
  \centering
  \includegraphics[width=\linewidth]{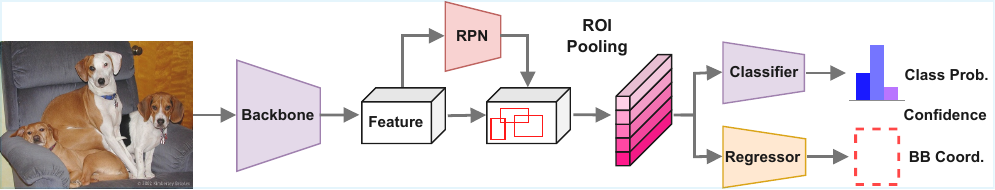}
  \caption{Overall pipeline of Faster R-CNN~\cite{ren2015faster}. The input image is first processed by a backbone network to extract feature maps. 
  These features are then fed into a Region Proposal Network (RPN) and Region of Interest (RoI) Pooling to generate RoI features.
  The resulting features are passed to a classifier and regressor to predict class probabilities and bounding box coordinates, respectively. 
  Our method selects predicted bounding boxes according to the method described in Section~\ref{sec:defprediction}.}
  \label{fig:fasterrcnn}
\end{figure}

Modern deep learning object detection frameworks can be broadly categorized into three paradigms: \textbf{two-stage detectors}, \textbf{one-stage detectors}, and \textbf{transformer-based} approaches.

\subsection{Two-Stage Detectors}
Two-stage detectors divide detection into region proposal generation, followed by classification and bounding box refinement. 
R-CNN~\cite{girshick2014rich} was a breakthrough, using Convolutional Neural Networks (CNNs) to extract features from region proposals generated by selective search, followed by classification and regression. 
Although accurate, R-CNN was computationally expensive due to repeated feature extraction for each region.
Fast R-CNN~\cite{girshick2015fast} mitigated this by introducing a shared backbone and RoI pooling, enabling end-to-end training and faster processing.
Faster R-CNN~\cite{ren2015faster} further improved efficiency by integrating Region Proposal Networks (RPNs) that share features with the detector, making proposals nearly cost-free. 
Figure~\ref{fig:fasterrcnn} illustrates this unified pipeline for efficient and accurate detection.
This approach outperforms one-stage methods in accuracy, particularly for small or overlapping objects.
Mask R-CNN~\cite{he2017mask} extended the framework to instance segmentation. 
These methods are preferred when accuracy is prioritized over speed.

\subsection{One-Stage Detectors}
One-stage detectors prioritize inference speed and architectural simplicity. 
YOLO (You Only Look Once)~\cite{redmon2018yolov3, redmon2016you} pioneered this approach by framing detection as a single regression problem, directly predicting bounding boxes and class probabilities from full images in one evaluation. 
SSD (Single Shot MultiBox Detector)~\cite{liu2016ssd} improved upon this idea by incorporating multi-scale feature maps and default boxes to handle objects of varying sizes more effectively. 
These architectures are particularly valued in resource-constrained environments and real-time applications due to their computational efficiency. 
While these methods enable high-speed detection, they may sacrifice accuracy in complex or crowded scenes, making them less suitable for tasks where fine-grained detection is necessary.

\subsection{Transformer-Based Detectors}
Transformer-based detectors have recently emerged, transferring the success of attention mechanisms from natural language processing to object detection. 
DETR (DEtection TRansformer)~\cite{carion2020endtoend} pioneered this approach by eliminating the need for many hand-designed components like non-maximum suppression, using a set-based global loss that forces unique predictions via bipartite matching. 
Deformable DETR~\cite{zhu2021deformable} addressed the slow convergence issues of DETR by introducing deformable attention mechanisms that attend to only a small set of key sampling points around a reference.
These approaches offer new perspectives on object detection by treating it as a direct set prediction problem, though they often require substantial computational resources for training.
Recent trends also include open-vocabulary object detection~\cite{zareian2021open}, which enables models to detect objects described by natural language beyond a fixed set of predefined classes, and self-supervised learning approaches that reduce the dependency on large annotated datasets~\cite{liu2021self}.

\subsection{Motivation for Dataset Pruning in Object Detection}
Since dataset pruning in the context of object detection remains relatively unexplored, we base our experiments on a two-stage detector, specifically Faster R-CNN, which provides a good balance between performance and speed. 
This choice is also motivated by prior work such as~\cite{lee2024coreset}.
Although one-stage detectors might benefit more immediately from reduced training data due to their inherent efficiency focus, Faster R-CNN's two-stage architecture presents a more challenging test case for our pruning methodology. Furthermore, insights gained from pruning for Faster R-CNN are likely to generalize well to other architectures given its foundational role in modern detection systems.

\section{Analysis on Score Relation}

\begin{figure}[t]
  \centering
  \includegraphics[width=\linewidth]{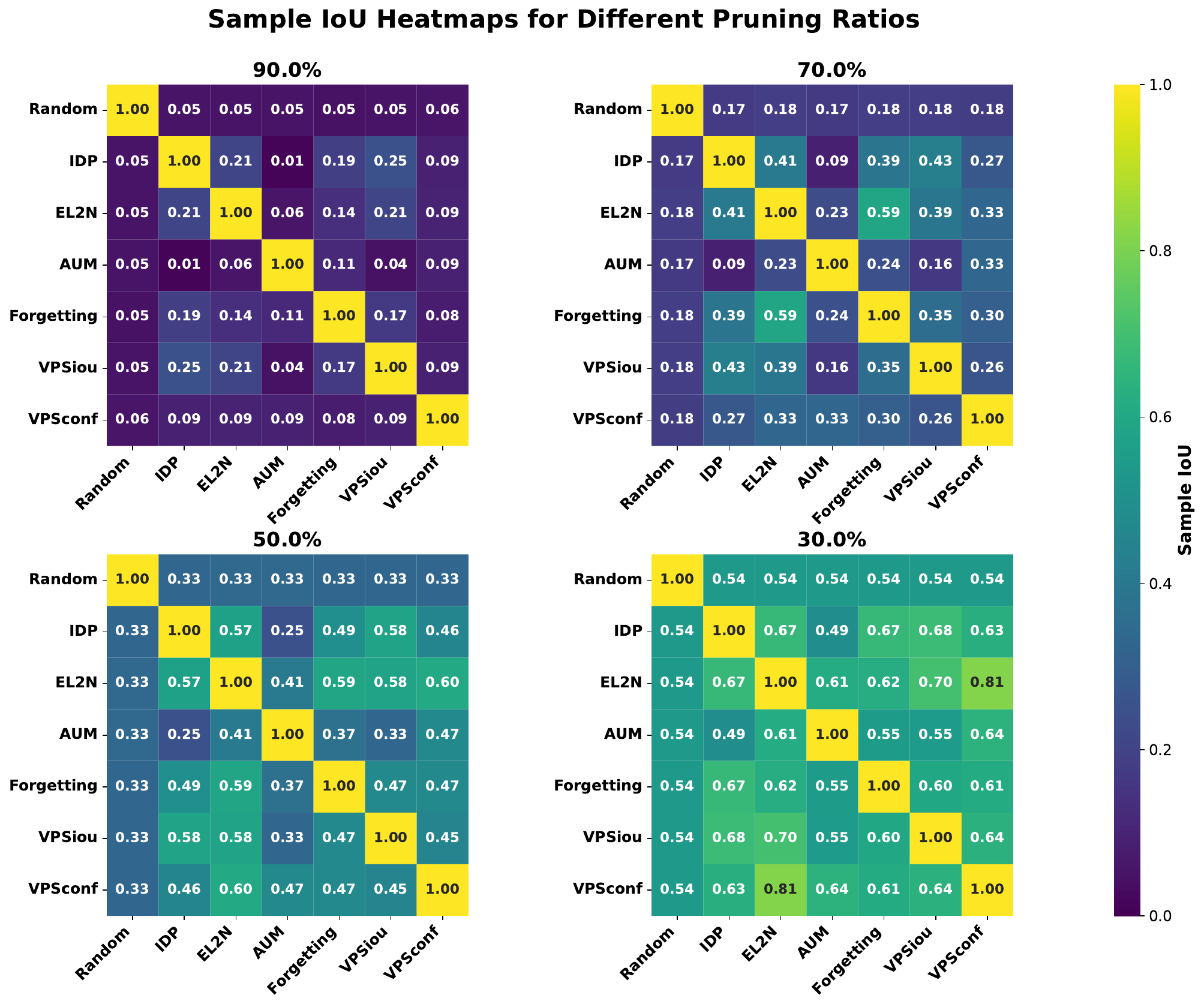}
  \caption{Heatmaps of sample IoU at different pruning rates on PASCAL VOC~\cite{everingham2010PASCAL}.
    The pruning rate for each heatmap is indicated at the top.
    A value close to 1 indicates that the two scoring methods select similar data points, while a value close to 0 indicates dissimilar selections.
    These results are obtained using max aggregation.
    }
  \label{fig:heatmap}
\end{figure}

To analyze the relationships between different scoring methods, we compare the sets of selected image IDs from each method. The degree of overlap is quantified using the Intersection over Union (IoU) metric, defined as:
\begin{equation}
\mathrm{IoU}(A, B) = \frac{|A \cap B|}{|A \cup B|}
\end{equation}
where \( A \) and \( B \) are the sets of selected image IDs by two different scores, \( |A \cap B| \) is the number of common elements, and \( |A \cup B| \) is the total number of unique elements across both sets.

Figure~\ref{fig:heatmap} shows the results. 
When comparing each score with the random baseline, we observe that the methods generally select moderately overlapping sets of samples, indicating partial agreement on sample importance. 
However, particularly at higher pruning rates, the overlaps remain relatively low (e.g., rarely exceeding 0.5 in Sample IoU), suggesting that the scoring methods capture different aspects of data utility.

A more detailed analysis reveals that the proposed method, VPS\textsubscript{iou} with max aggregation, exhibits a selection pattern similar to those of IDP and EL2N. In contrast, another proposed method, VPS\textsubscript{conf}, shows high similarity with EL2N at lower pruning rates, but relatively low Sample IoU with VPS\textsubscript{iou}. This suggests that each score emphasizes different criteria when evaluating image importance.
Leveraging this property, a promising future direction would be to integrate these diverse scoring methods into a unified framework for more robust sample selection.

\section{Further Analysis on Aggregation Methods}
\label{app:agg}

\begin{table}[t]
  \centering
  \caption{Comparison of different aggregation methods using Average and Sum strategies across various pruning rates on the PASCAL VOC dataset~\cite{everingham2010PASCAL}. Best results are indicated in bold, with second-best results underlined.}
  \label{tab:compagg}
  \begin{tabular}{lcccccccc}
    \toprule
    \multirow{2}{*}{Method} & \multicolumn{4}{c}{Average Aggregation: mAP (\%)} & \multicolumn{4}{c}{Sum Aggregation: mAP (\%)} \\
    \cmidrule(lr){2-5} \cmidrule(lr){6-9}
    & 30\% & 50\% & 70\% & 90\% & 30\% & 50\% & 70\% & 90\% \\
    \midrule
    AUM~\cite{pleiss2020identifying}       & 49.36 & 45.60 & 38.79 & 22.10 & 43.41 & 35.40 & 31.46 & 17.67 \\
    EL2N~\cite{paul2021deep}              & \underline{49.50} & 45.75 & 38.49 & 21.82 & 49.82 & 46.31 & 40.89 & 28.70 \\
    Entropy~\cite{coleman2019selection}   & 48.78 & 44.52 & 36.77 & 18.21 & 49.28 & 46.59 & 41.29 & 28.39 \\
    Forgetting~\cite{toneva2019empirical} & 48.92 & \underline{46.24} & \underline{40.15} & 22.22 & 49.14 & 46.16 & 40.96 & 28.06 \\
    IoU                          & 49.21 & 45.53 & 39.93 & \underline{25.03} & 42.77 & 35.46 & 28.94 & 15.67 \\
    Confidence                   & 49.34 & 45.24 & 38.25 & 21.06 & 43.03 & 35.32 & 29.66 & 16.22 \\
    \midrule
    \textbf{VPS\textsubscript{conf} (Ours)} & 48.84 & 45.13 & 37.19 & 21.83 & \underline{49.91} & \textbf{47.43} & \underline{41.61} & \textbf{29.05} \\
    \textbf{VPS\textsubscript{iou} (Ours)}  & \textbf{50.11} & \textbf{46.73} & \textbf{41.67} & \textbf{27.21} & \textbf{50.08} & \underline{46.97} & \textbf{42.06} & \underline{28.76} \\
    \bottomrule
  \end{tabular}
\end{table}

In this section, we present additional results using alternative aggregation strategies, namely \textit{average} and \textit{sum}, which were not included in the main part of this paper. The results are summarized in Table~\ref{tab:compagg}.
First, the performance of VPS\textsubscript{iou} using the \textit{average} aggregation method surpasses all other baselines. 
Even when using the \textit{sum} method, it consistently ranks either first or second, demonstrating the robustness of the proposed approach.
In contrast, the performance of VPS\textsubscript{conf} with the \textit{average} method is inferior to that of other baselines. 
However, when using the \textit{sum} strategy, it competes closely with VPS\textsubscript{iou} and often ranks among the best.

\section{Visualization of Top-Ranked and Pruned Data Points}
\label{app:visu}

In this section, we provide detailed visualizations of data points ranked using our proposed VPS (Variance-based Prediction Score) scoring strategy, which demonstrated strong performance in the main paper. The VPS score can be computed based on different criteria, including IoU (VPS\textsubscript{iou}) and confidence (VPS\textsubscript{conf}), using various aggregation methods. For the visualizations below, we adopt the max aggregation approach, which consistently performed well across datasets.

\begin{figure}[t]
    \centering

    \begin{subfigure}[t]{\textwidth}
        \centering
        \includegraphics[width=0.24\textwidth]{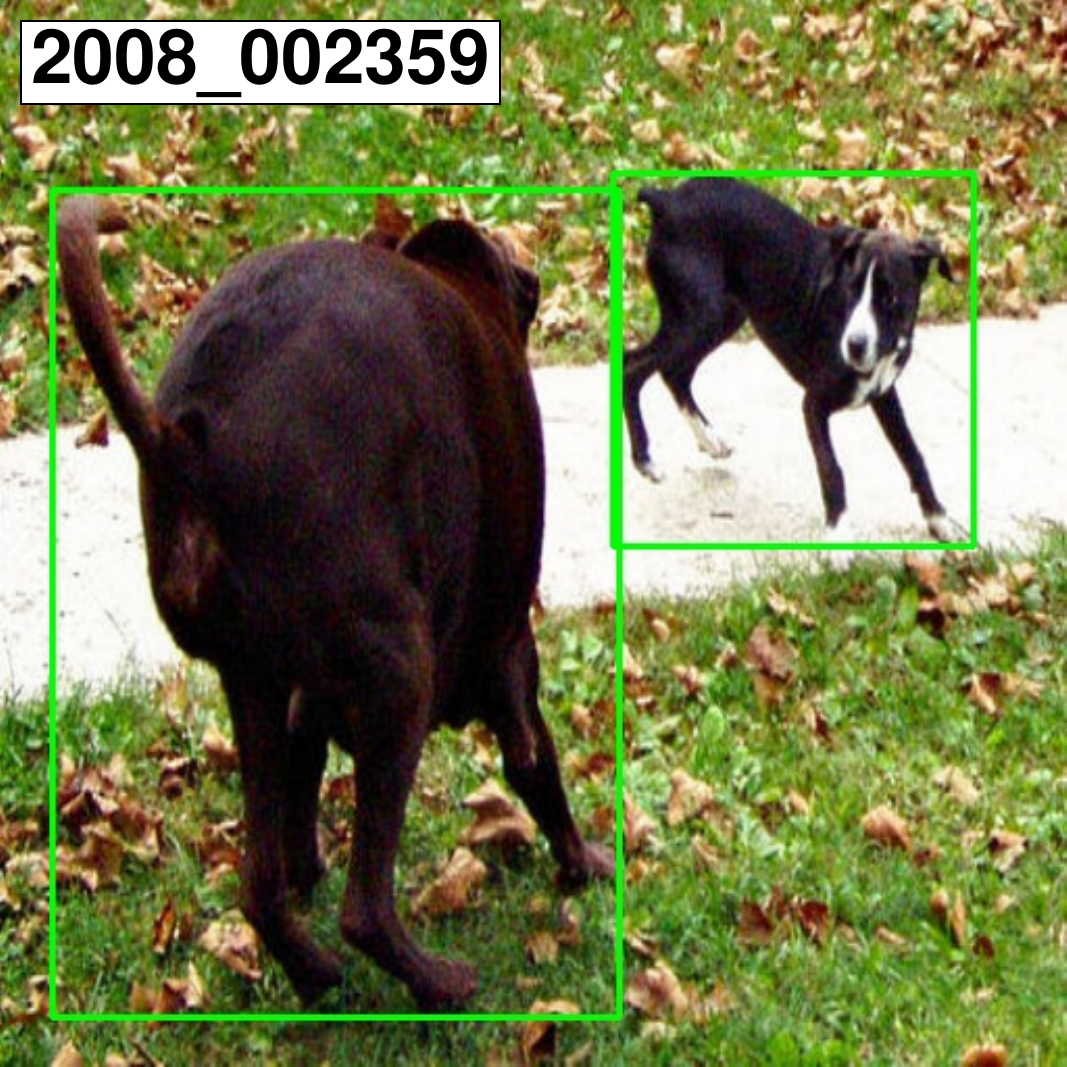}
        \includegraphics[width=0.24\textwidth]{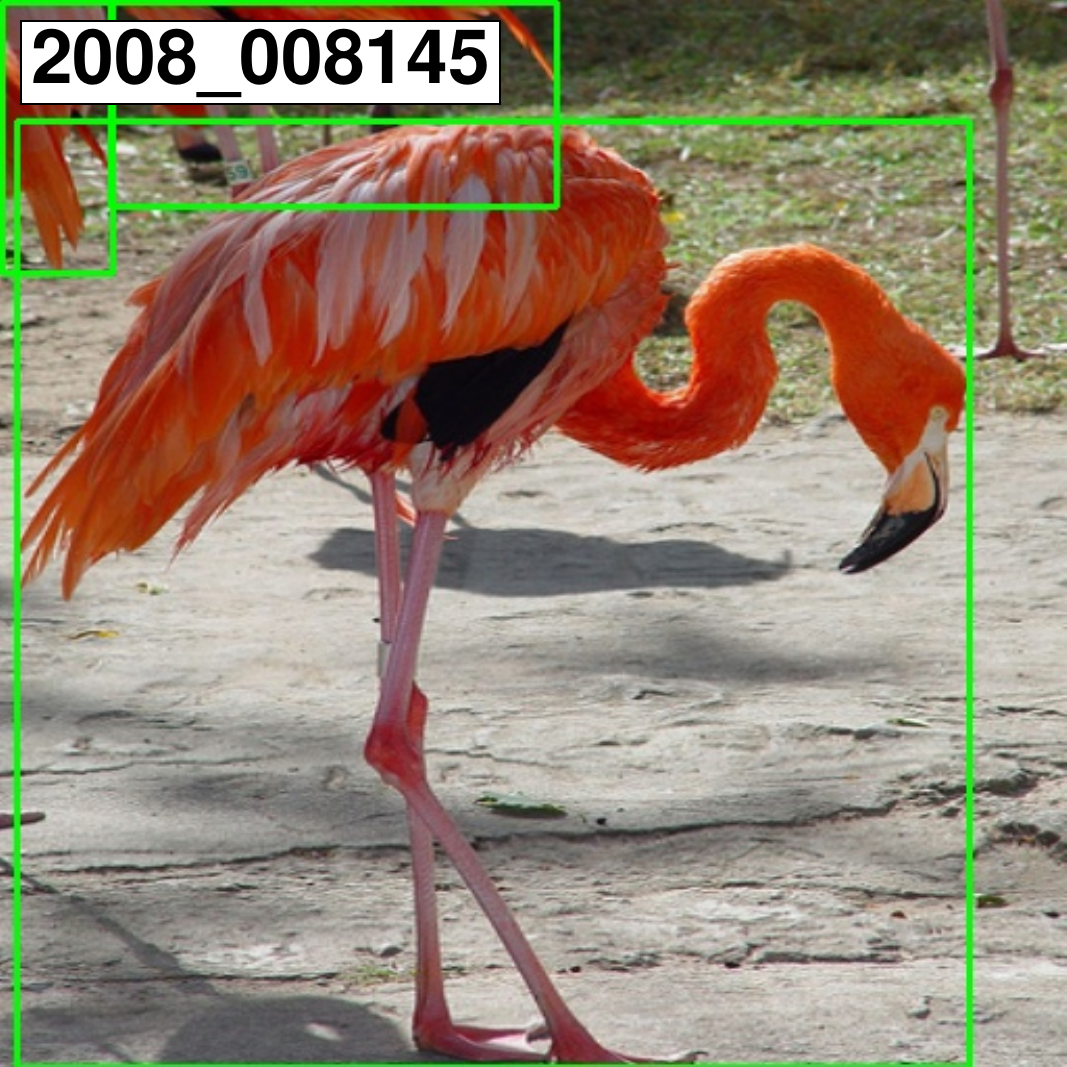}
        \includegraphics[width=0.24\textwidth]{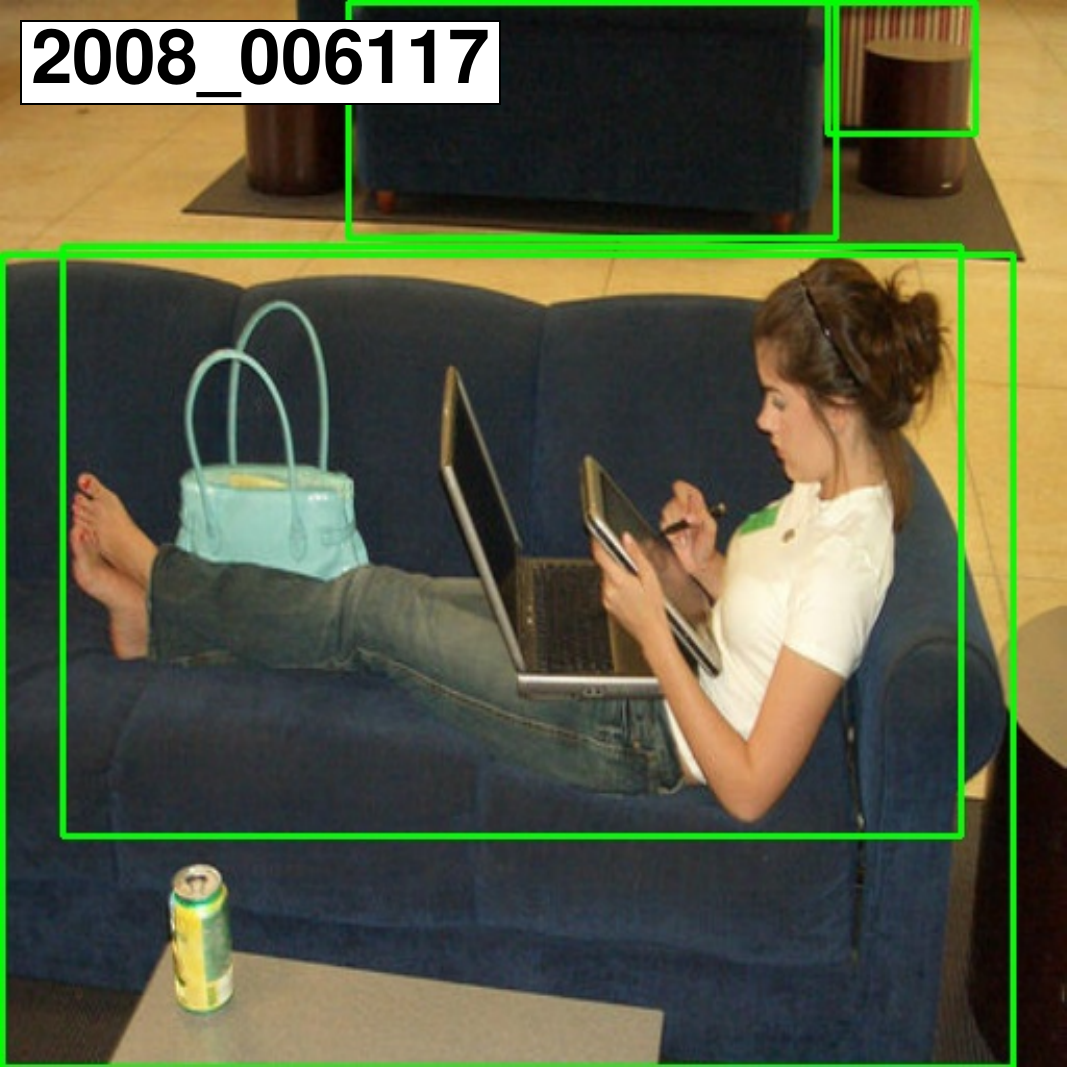}
        \includegraphics[width=0.24\textwidth]{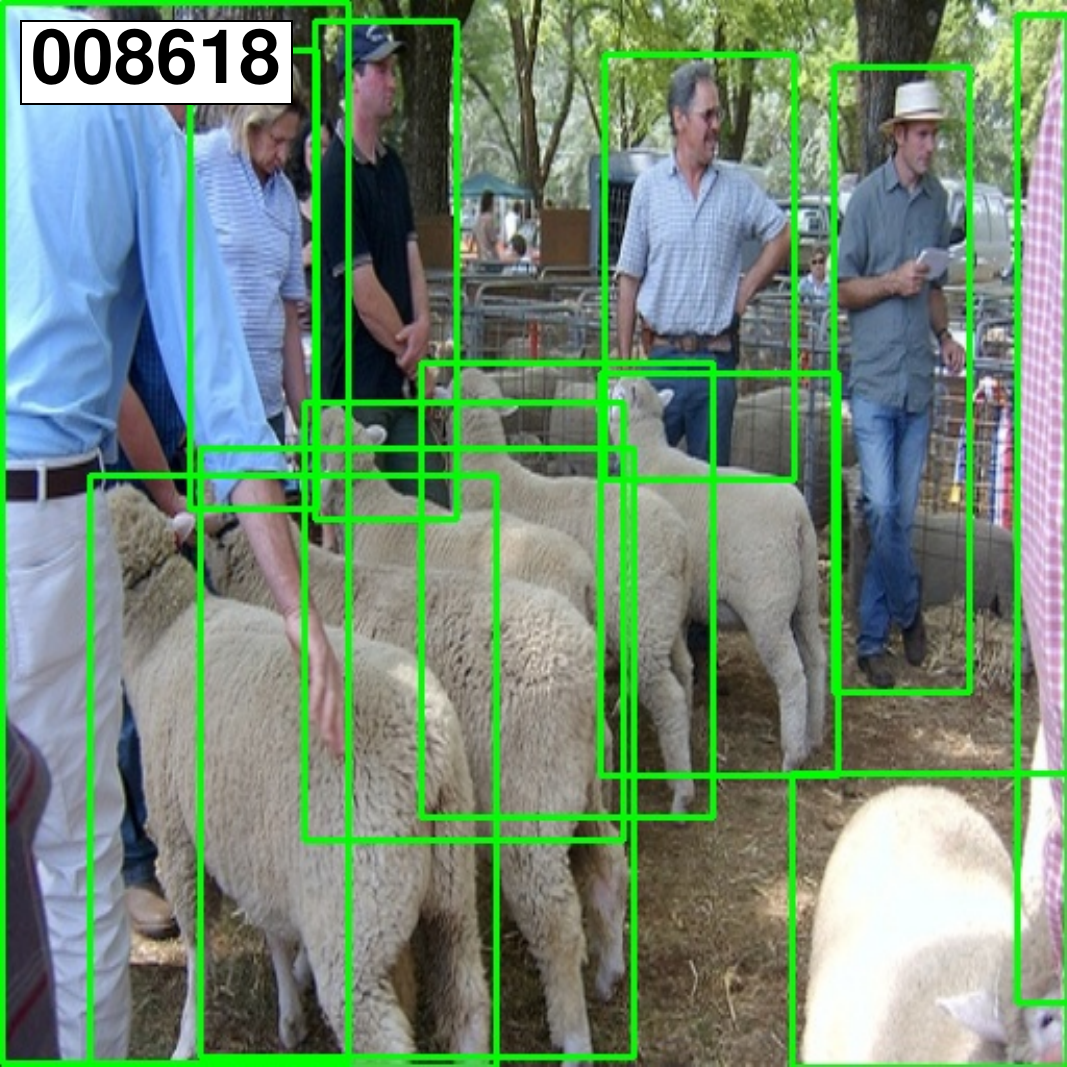}

        \includegraphics[width=0.24\textwidth]{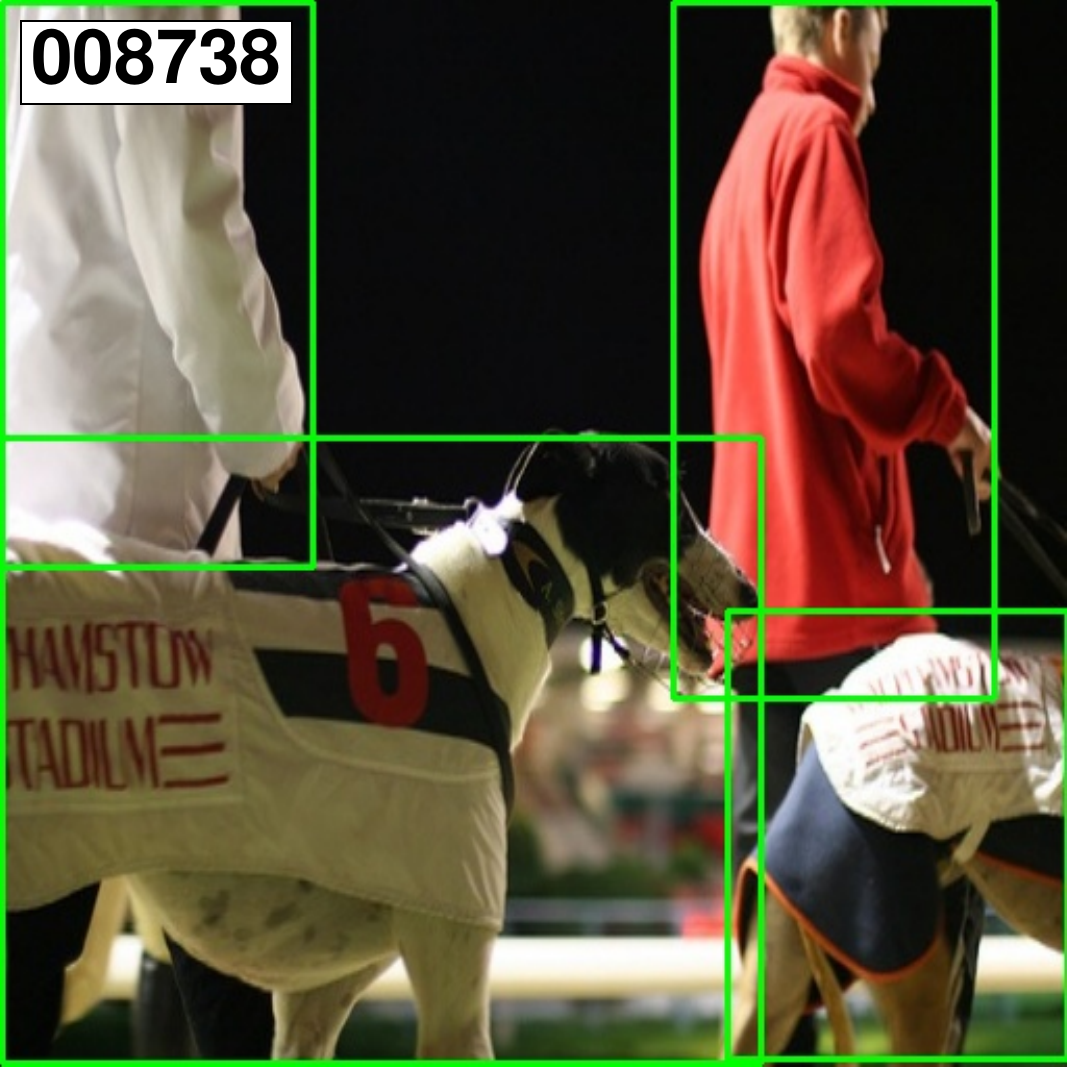}
        \includegraphics[width=0.24\textwidth]{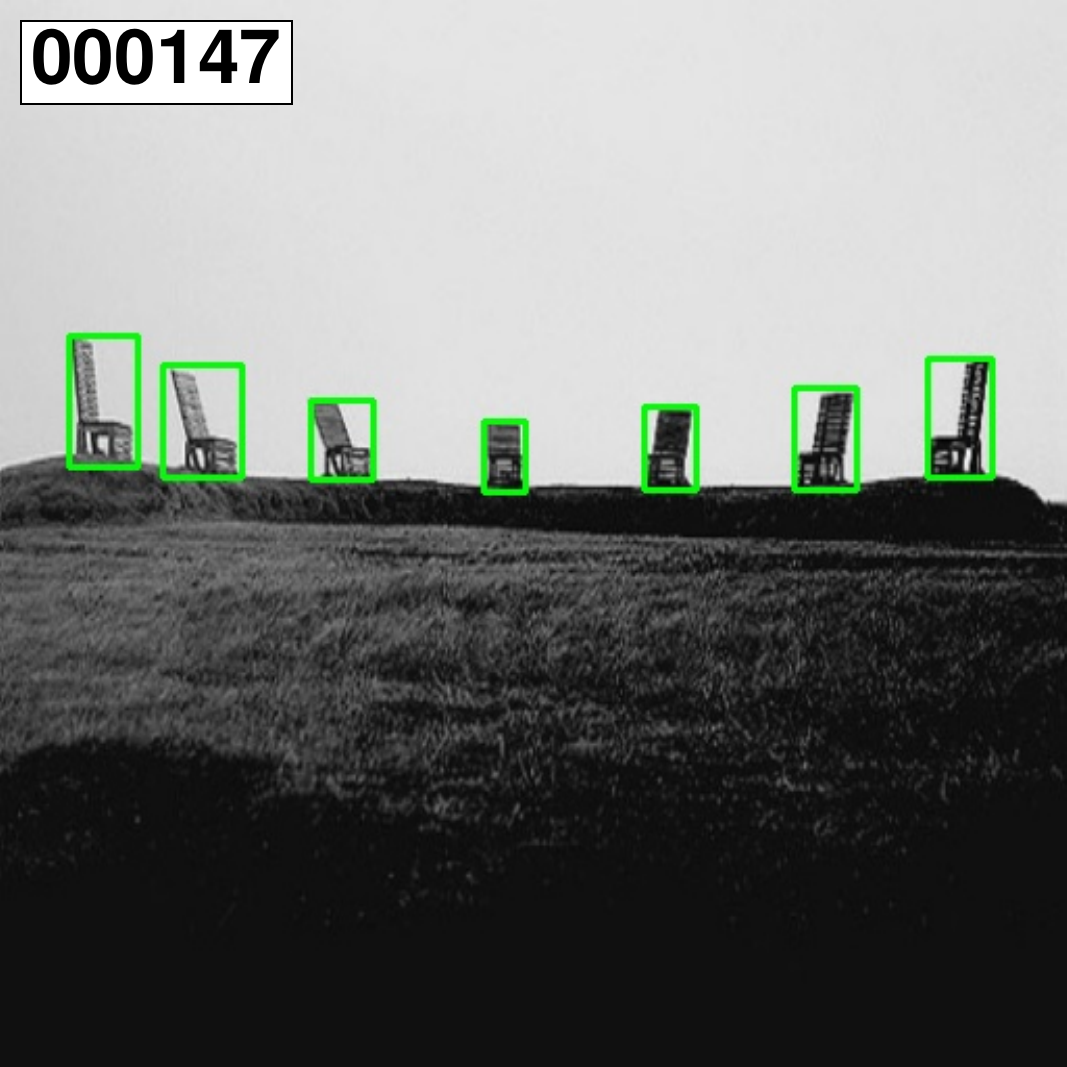}
        \includegraphics[width=0.24\textwidth]{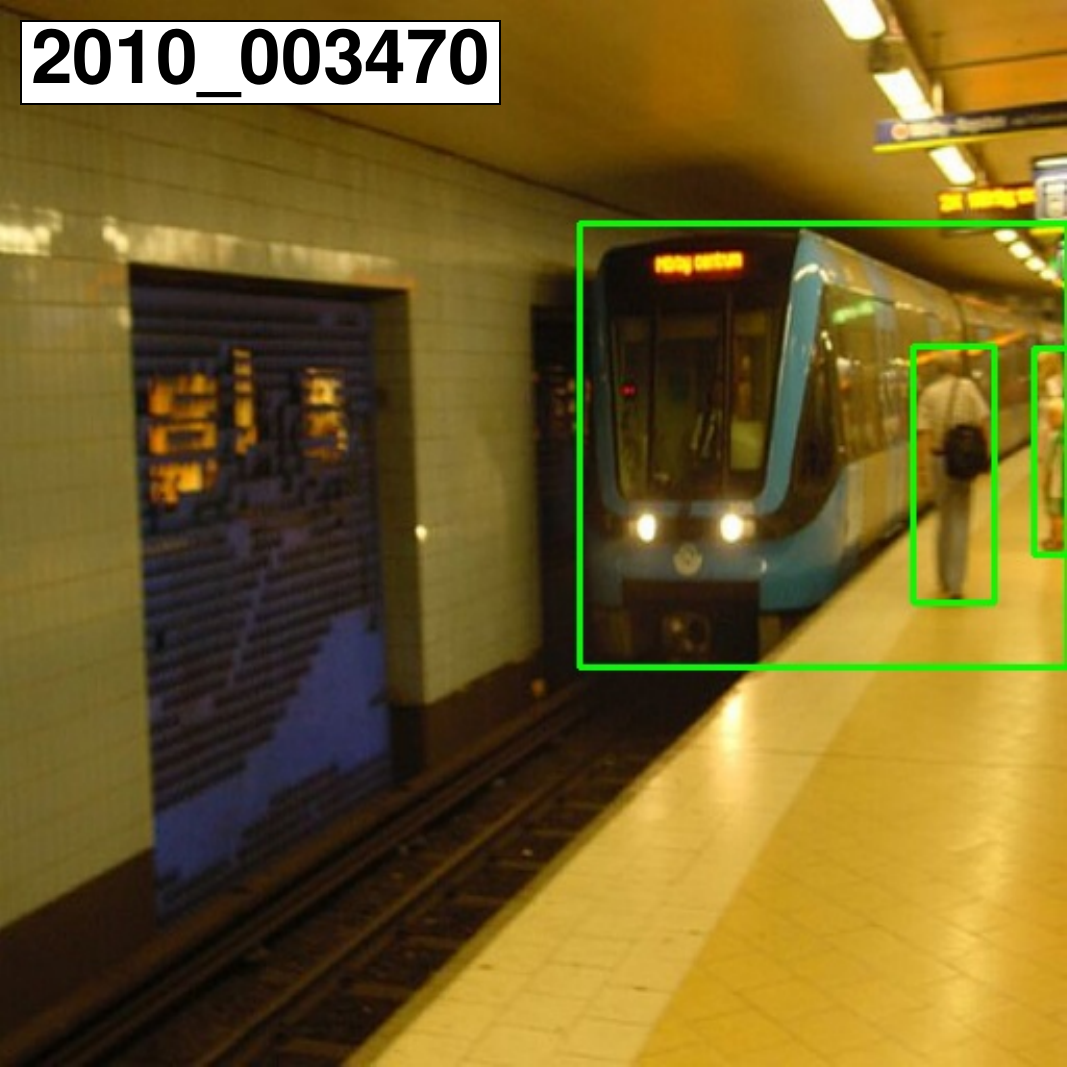}
        \includegraphics[width=0.24\textwidth]{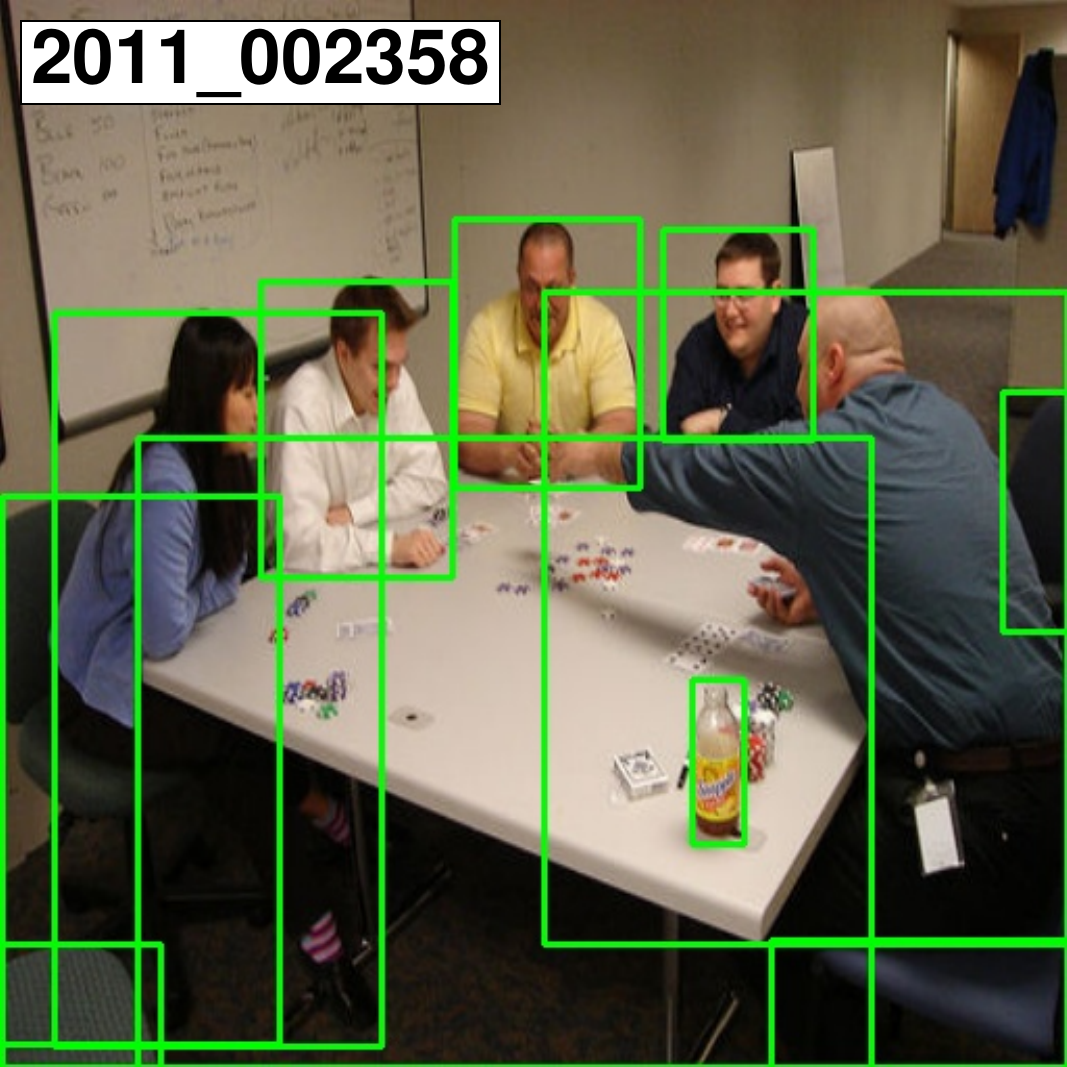}
        \subcaption{Selected samples from PASCAL VOC}
        \label{fig:vocselected}
    \end{subfigure}

    \vspace{2mm}

    \begin{subfigure}[t]{\textwidth}
        \centering
        \includegraphics[width=0.24\textwidth]{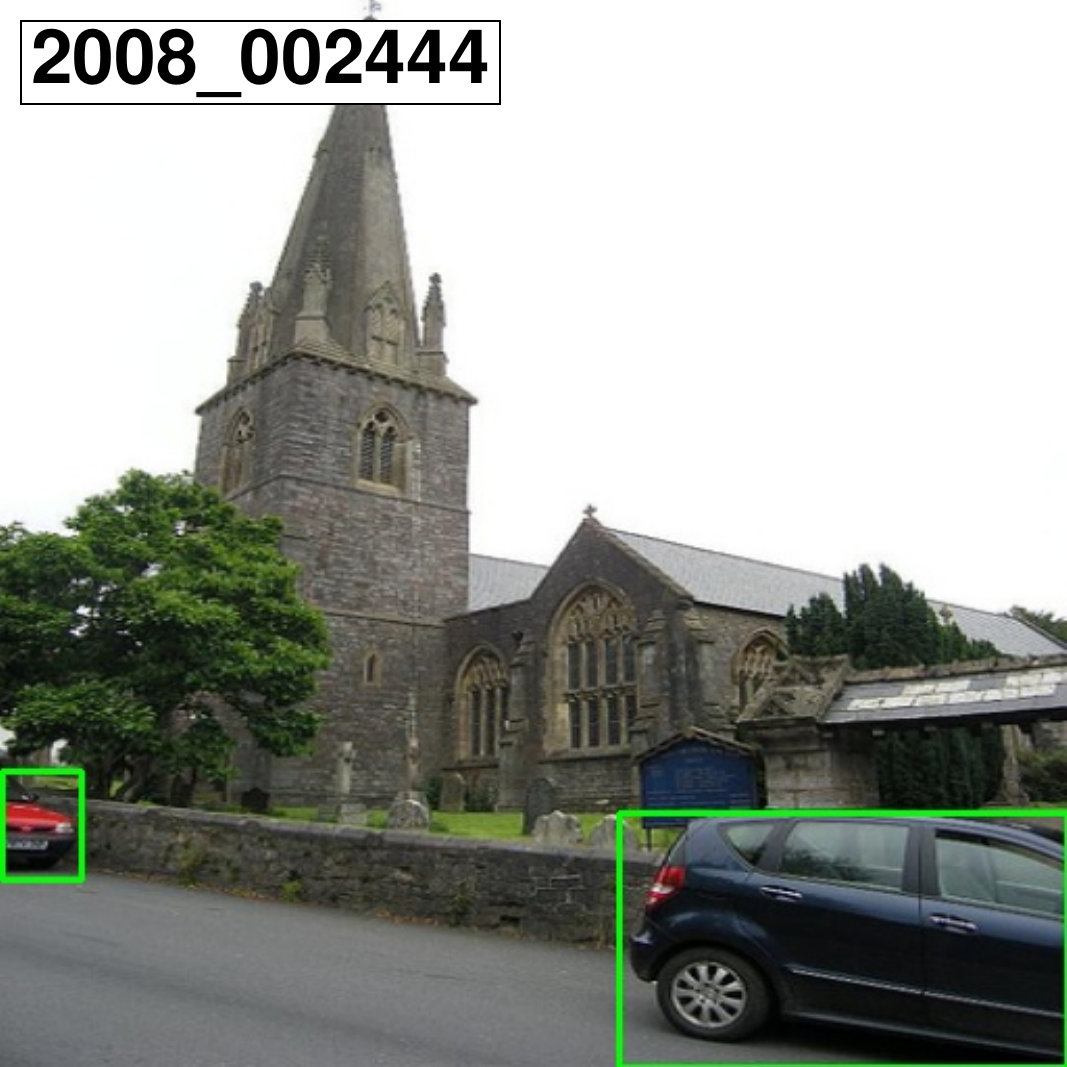}
        \includegraphics[width=0.24\textwidth]{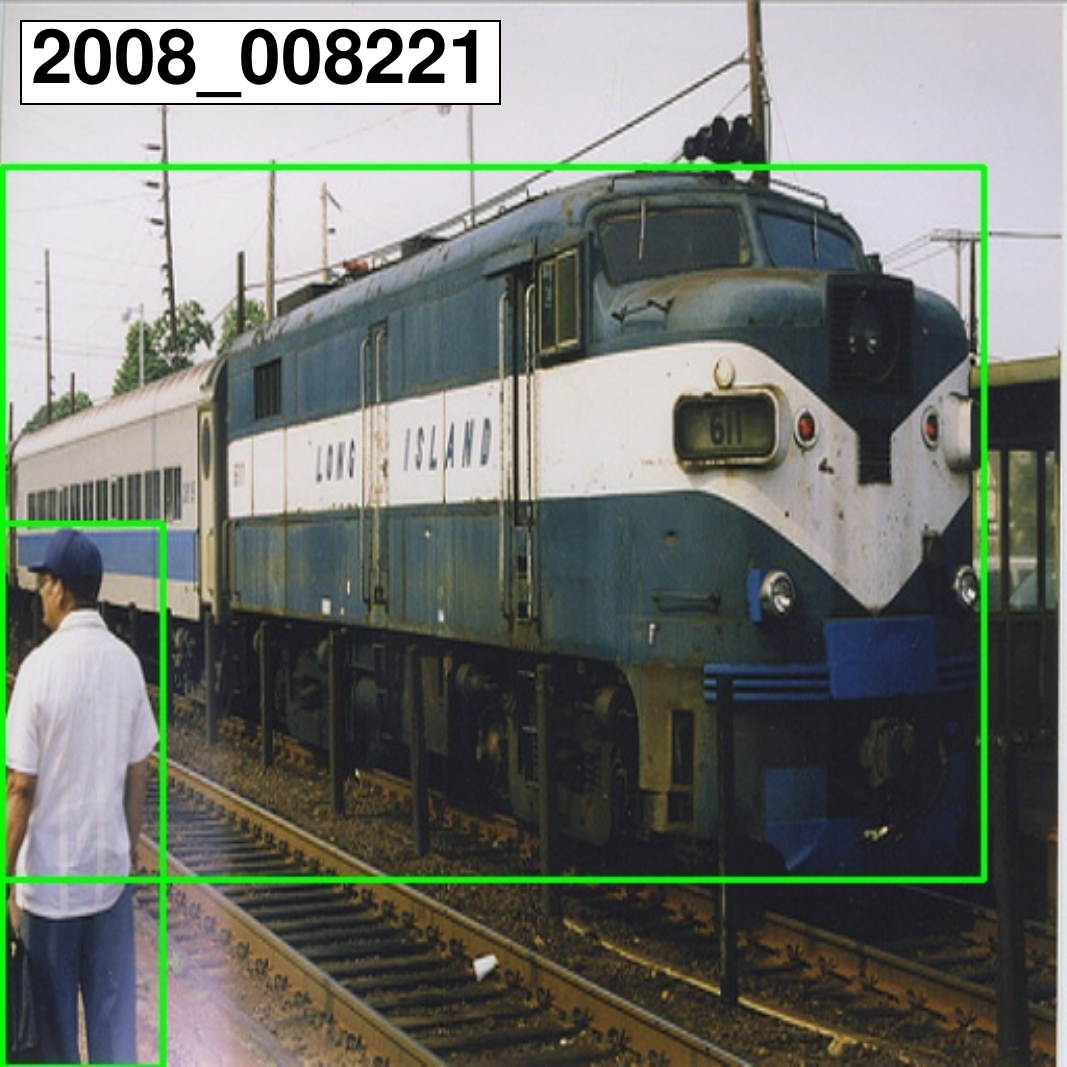}
        \includegraphics[width=0.24\textwidth]{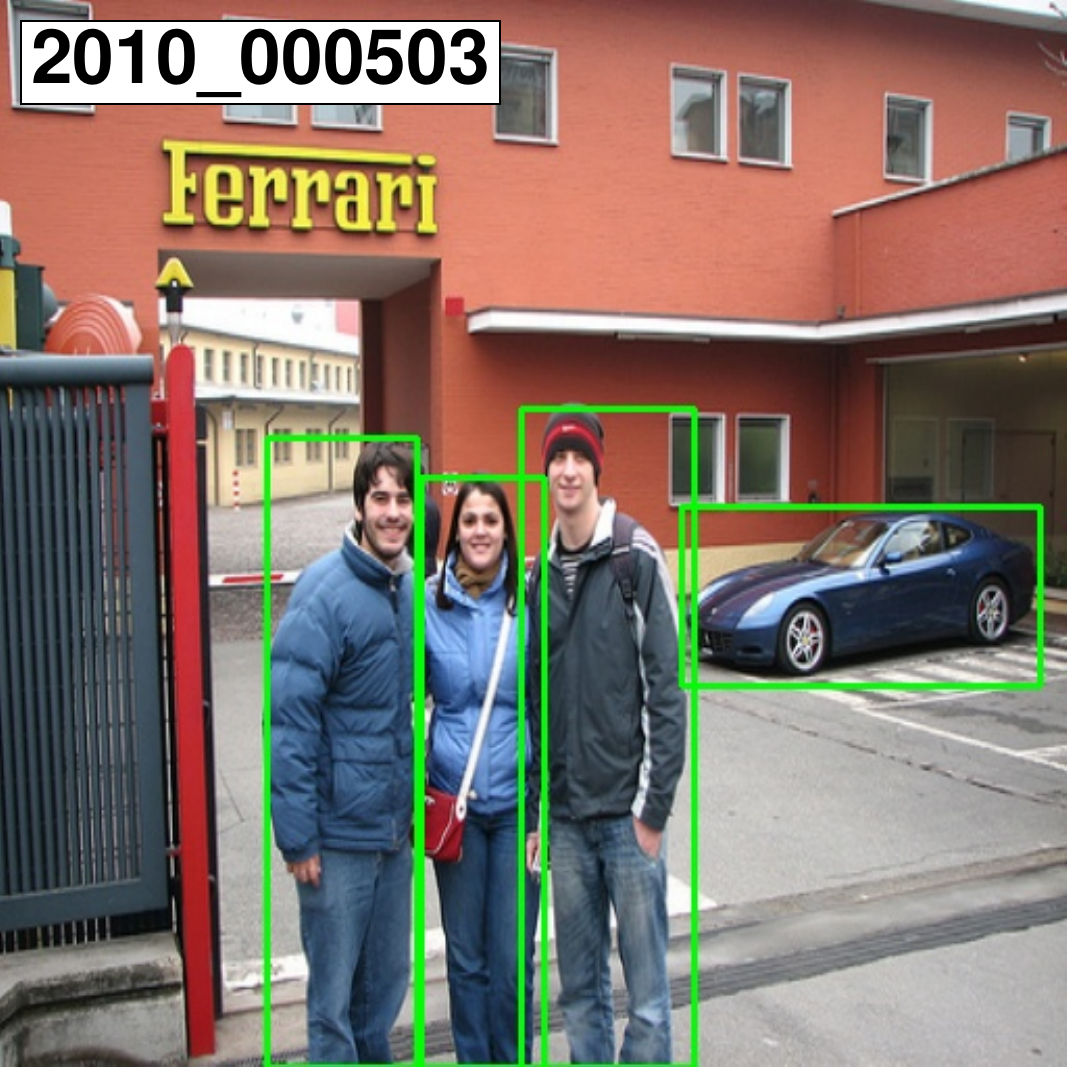}
        \includegraphics[width=0.24\textwidth]{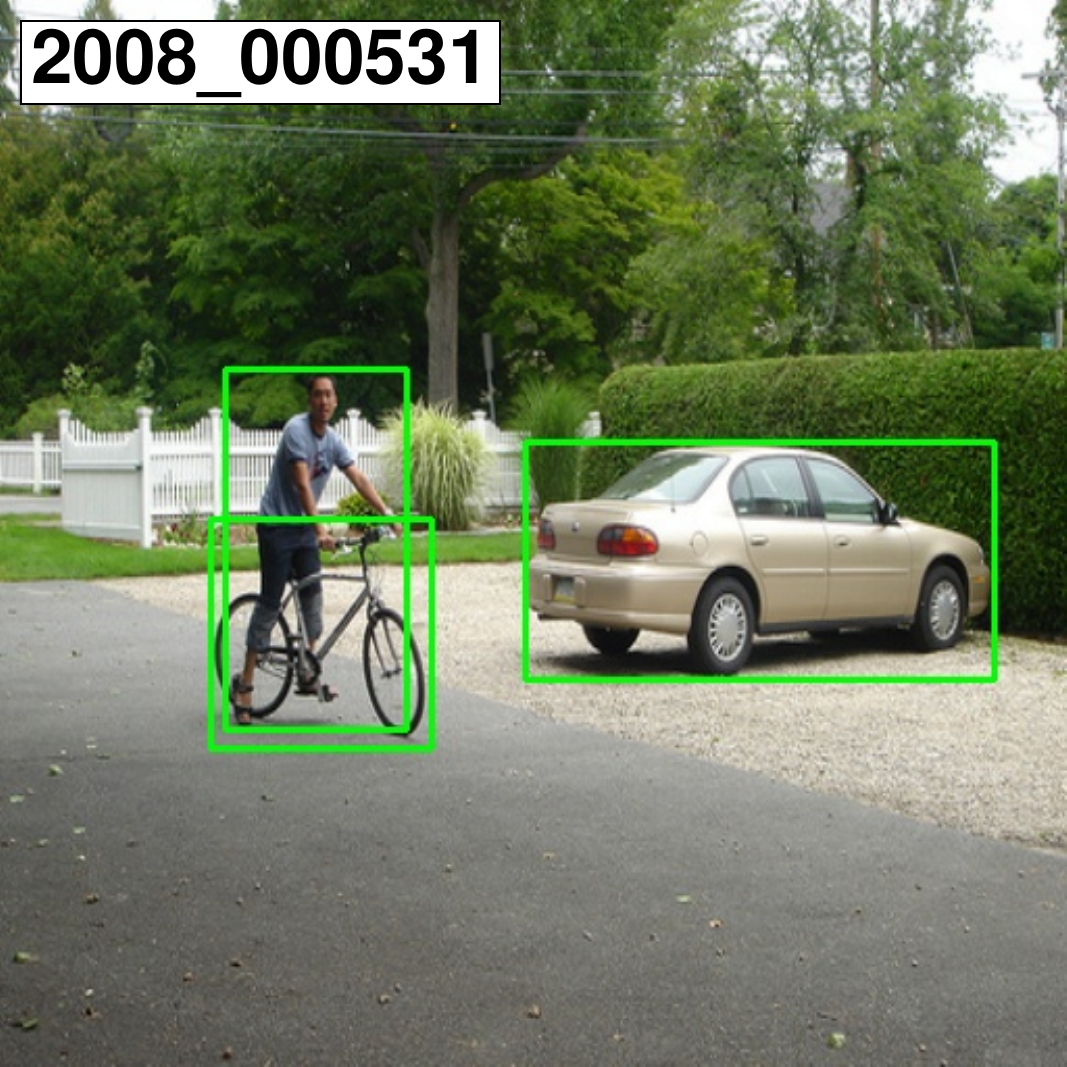}

        \includegraphics[width=0.24\textwidth]{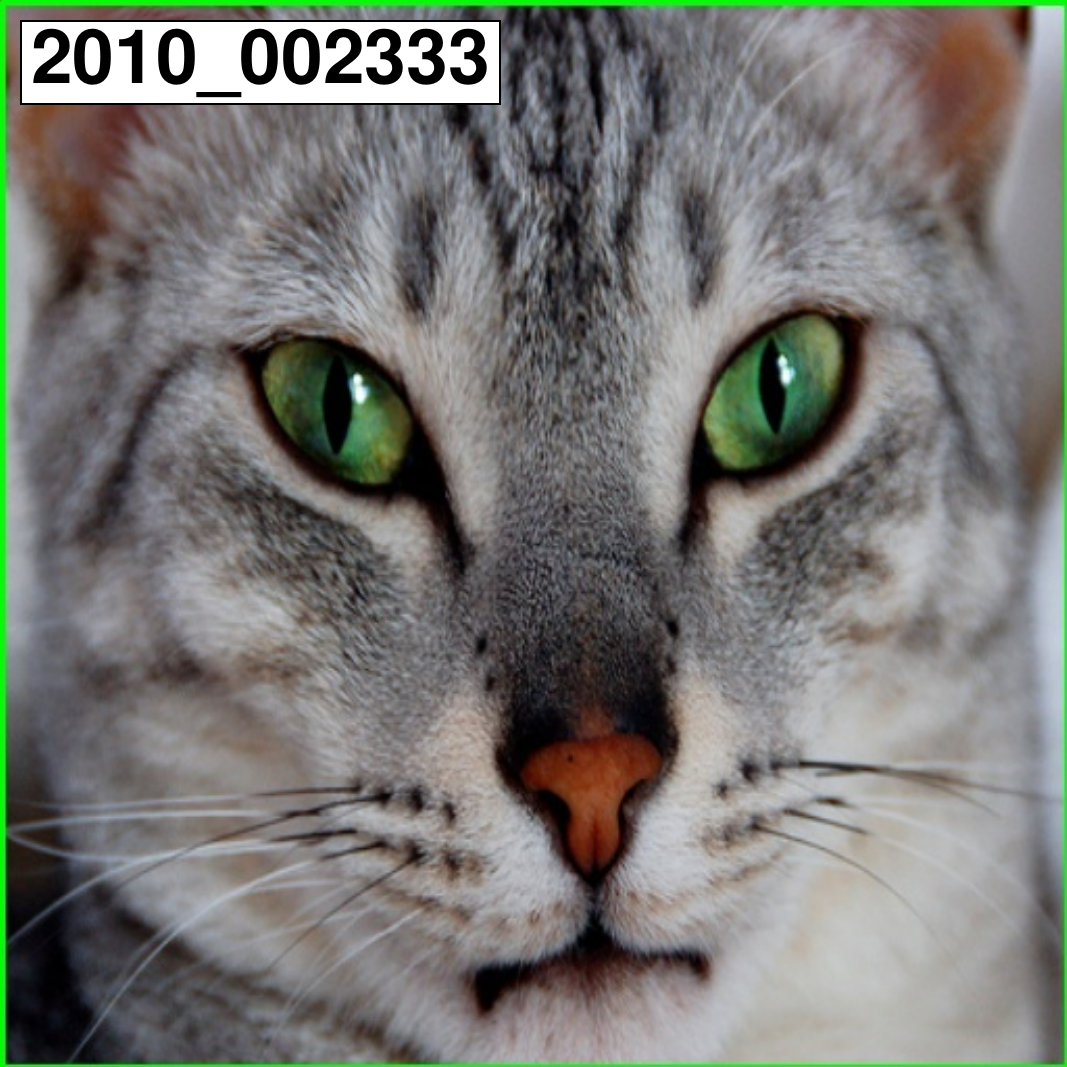}
        \includegraphics[width=0.24\textwidth]{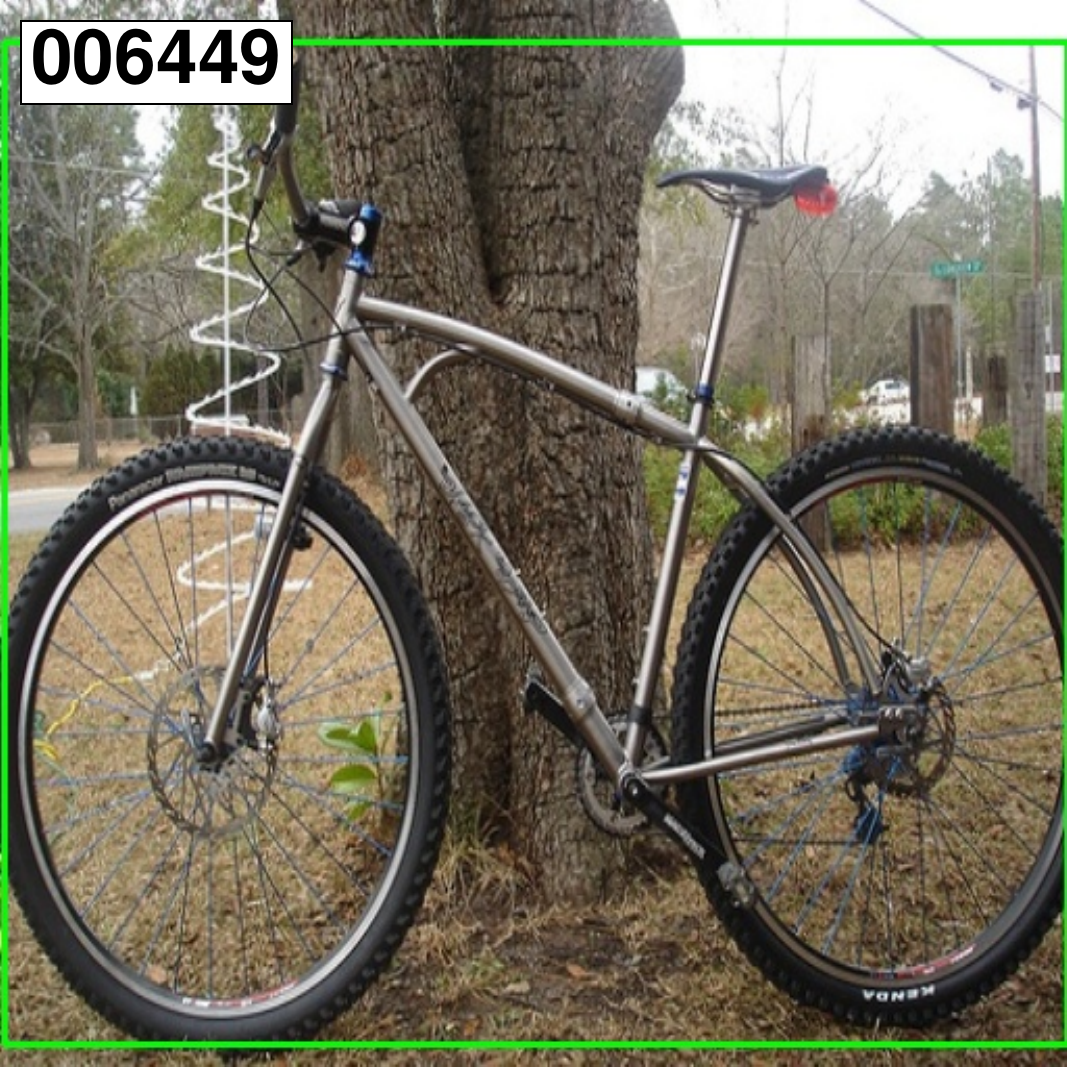}
        \includegraphics[width=0.24\textwidth]{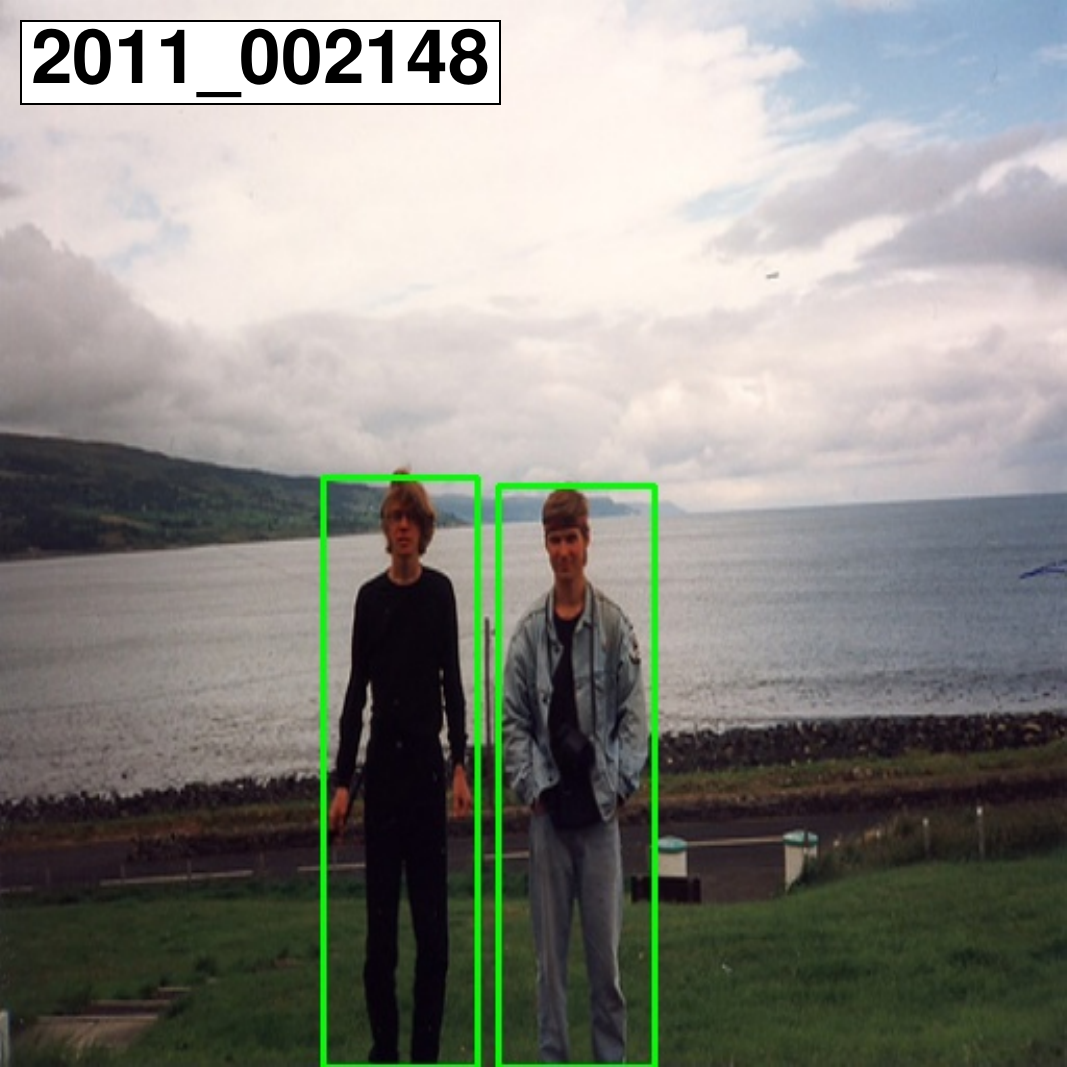}
        \includegraphics[width=0.24\textwidth]{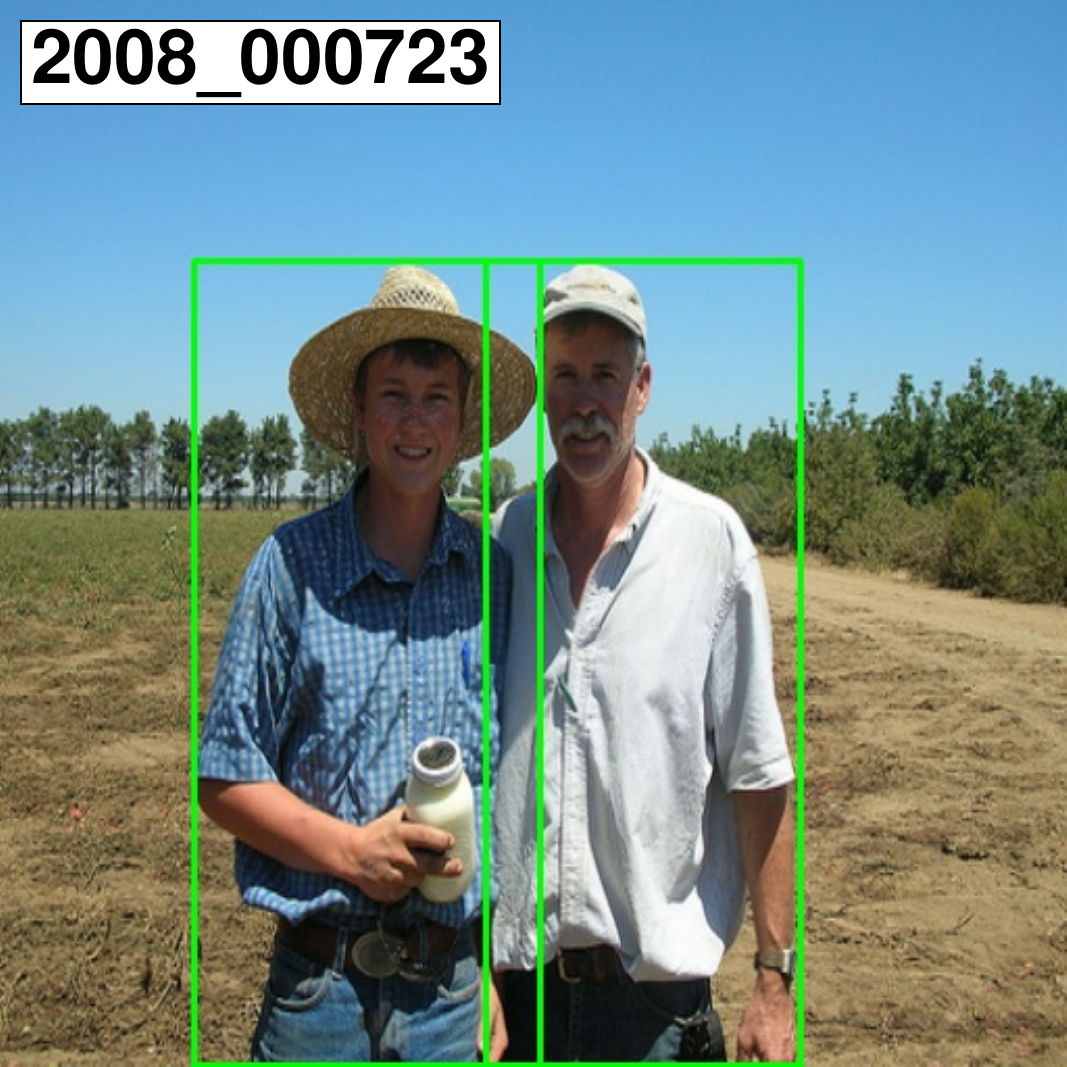}
        \subcaption{Easy pruned samples from PASCAL VOC}
        \label{fig:voceasy}
    \end{subfigure}

    \vspace{2mm}

    \begin{subfigure}[t]{\textwidth}
        \centering
        \includegraphics[width=0.24\textwidth]{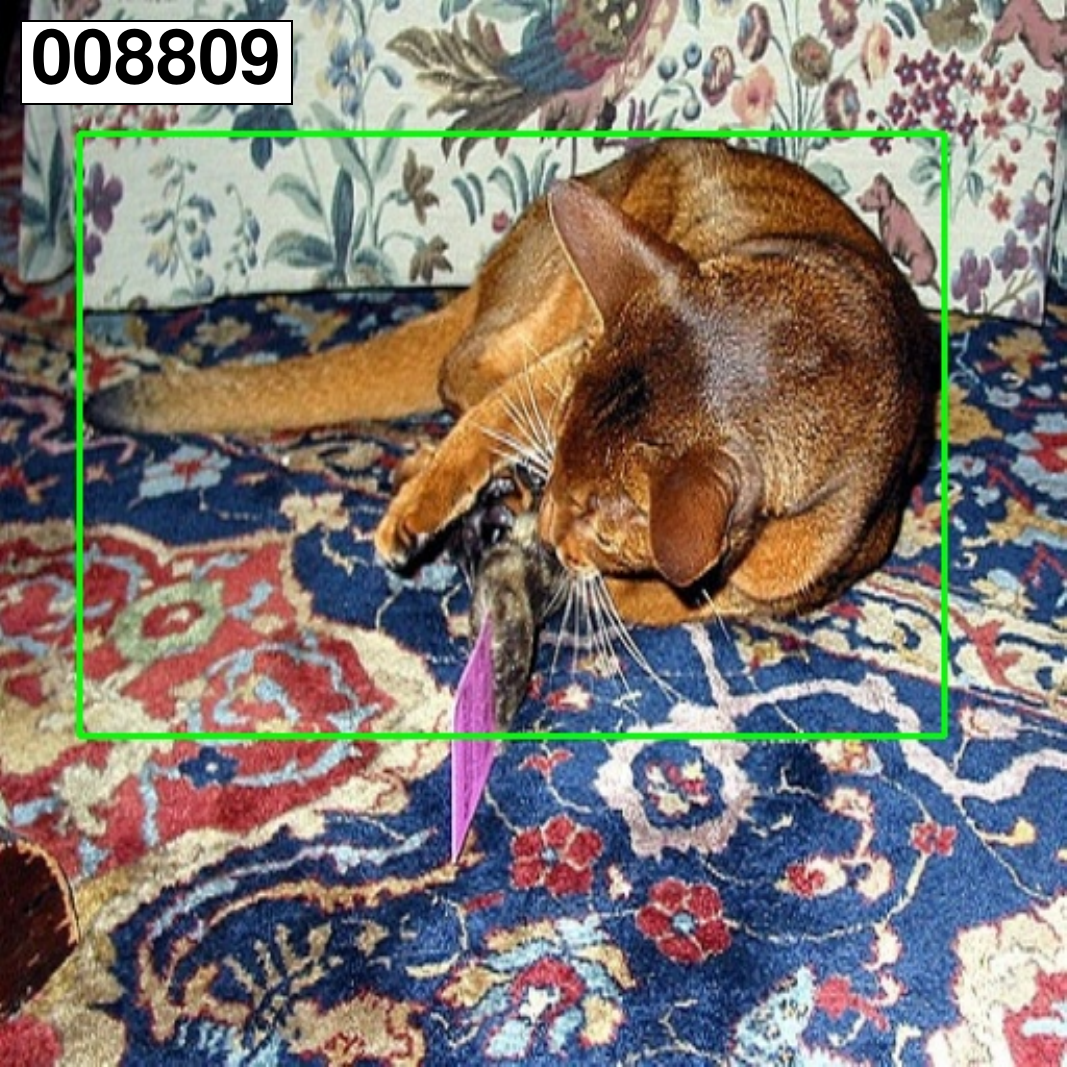}
        \includegraphics[width=0.24\textwidth]{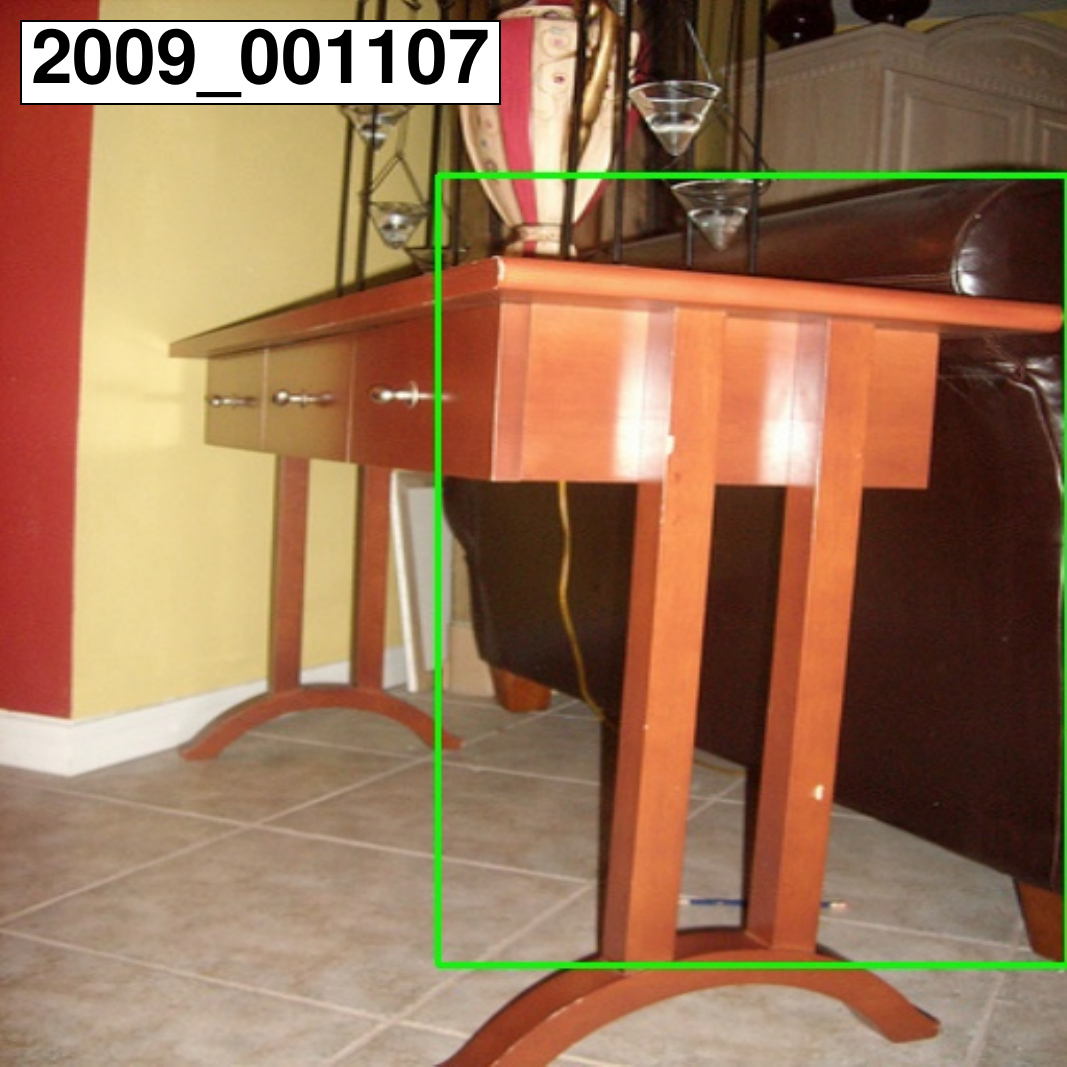}
        \includegraphics[width=0.24\textwidth]{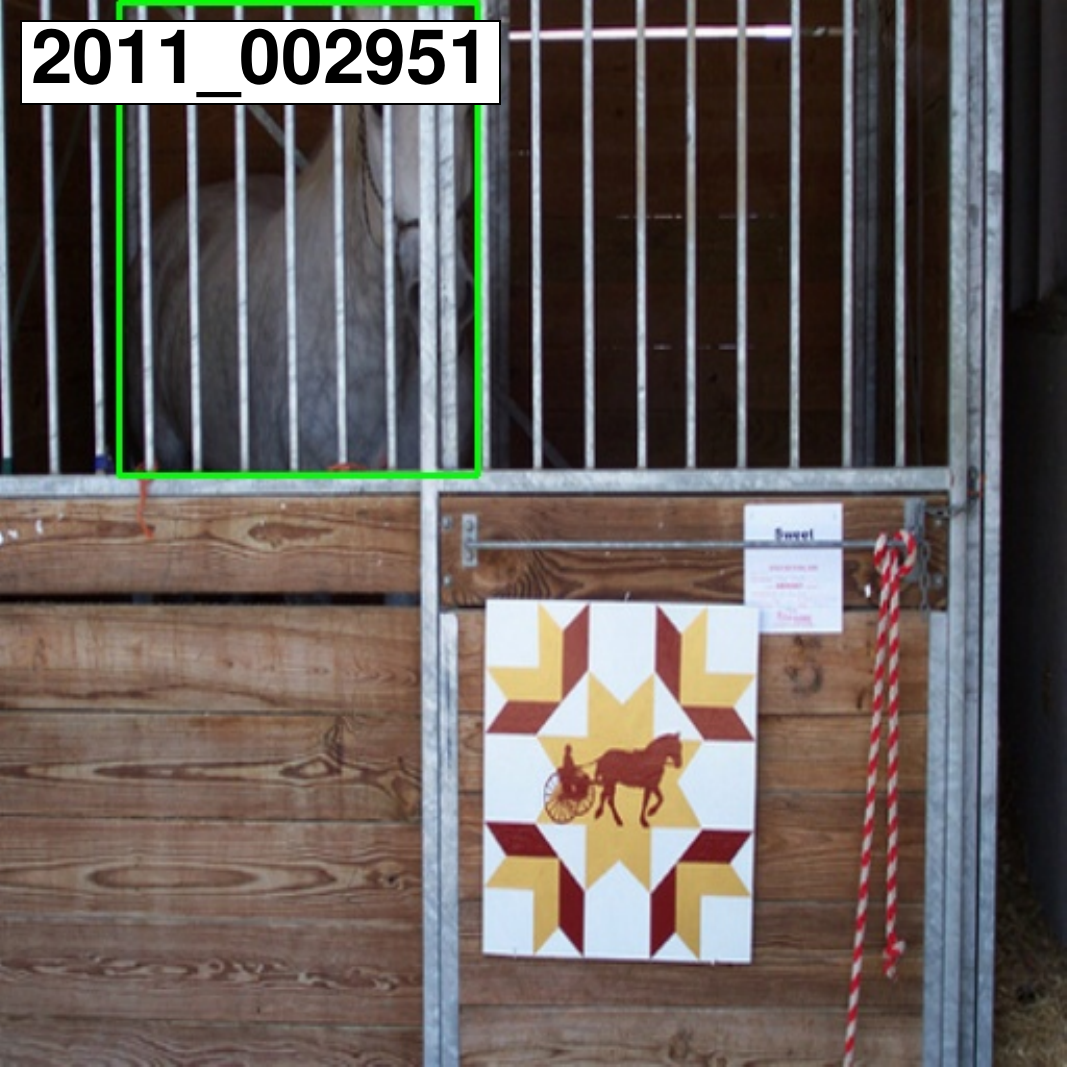}
        \includegraphics[width=0.24\textwidth]{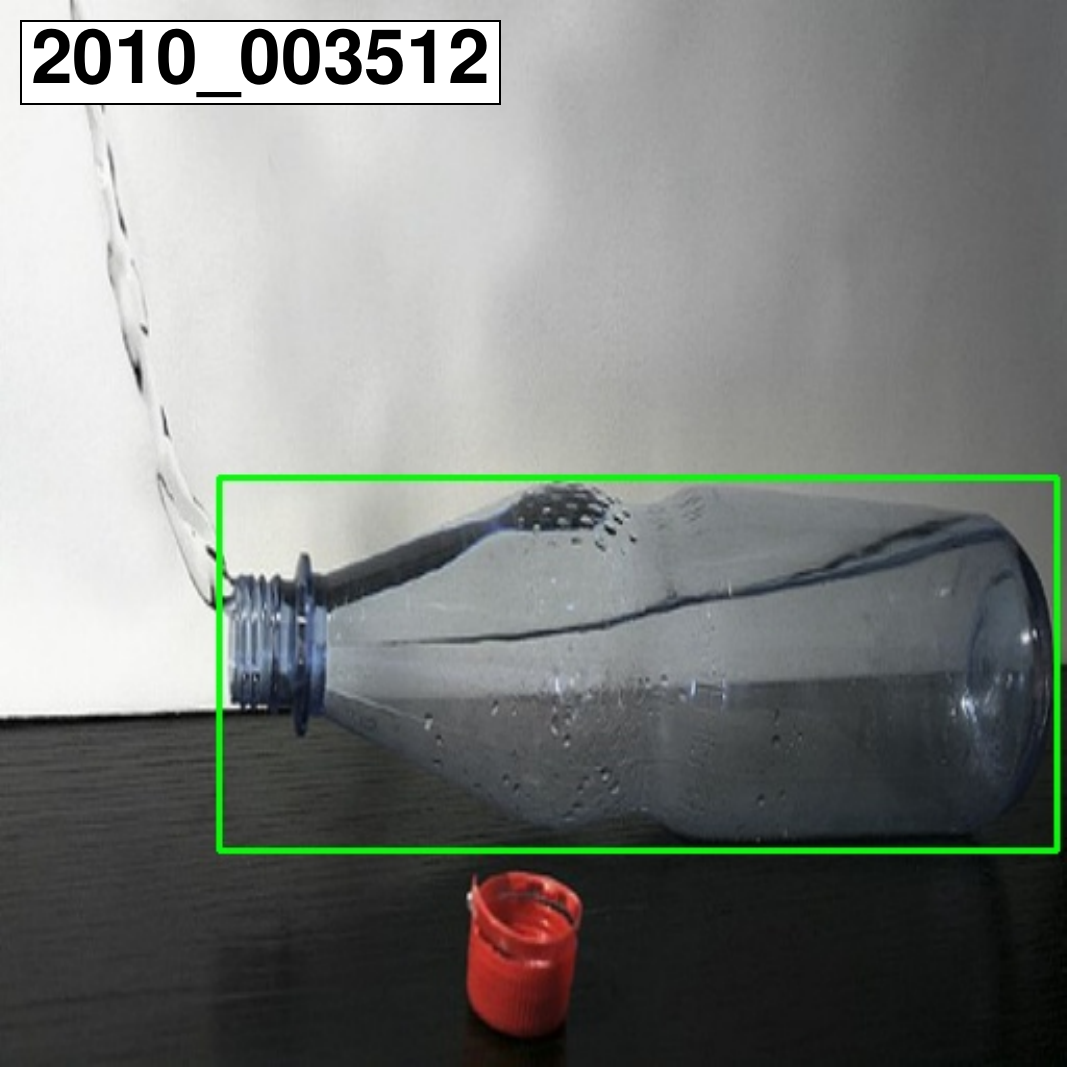}

        \includegraphics[width=0.24\textwidth]{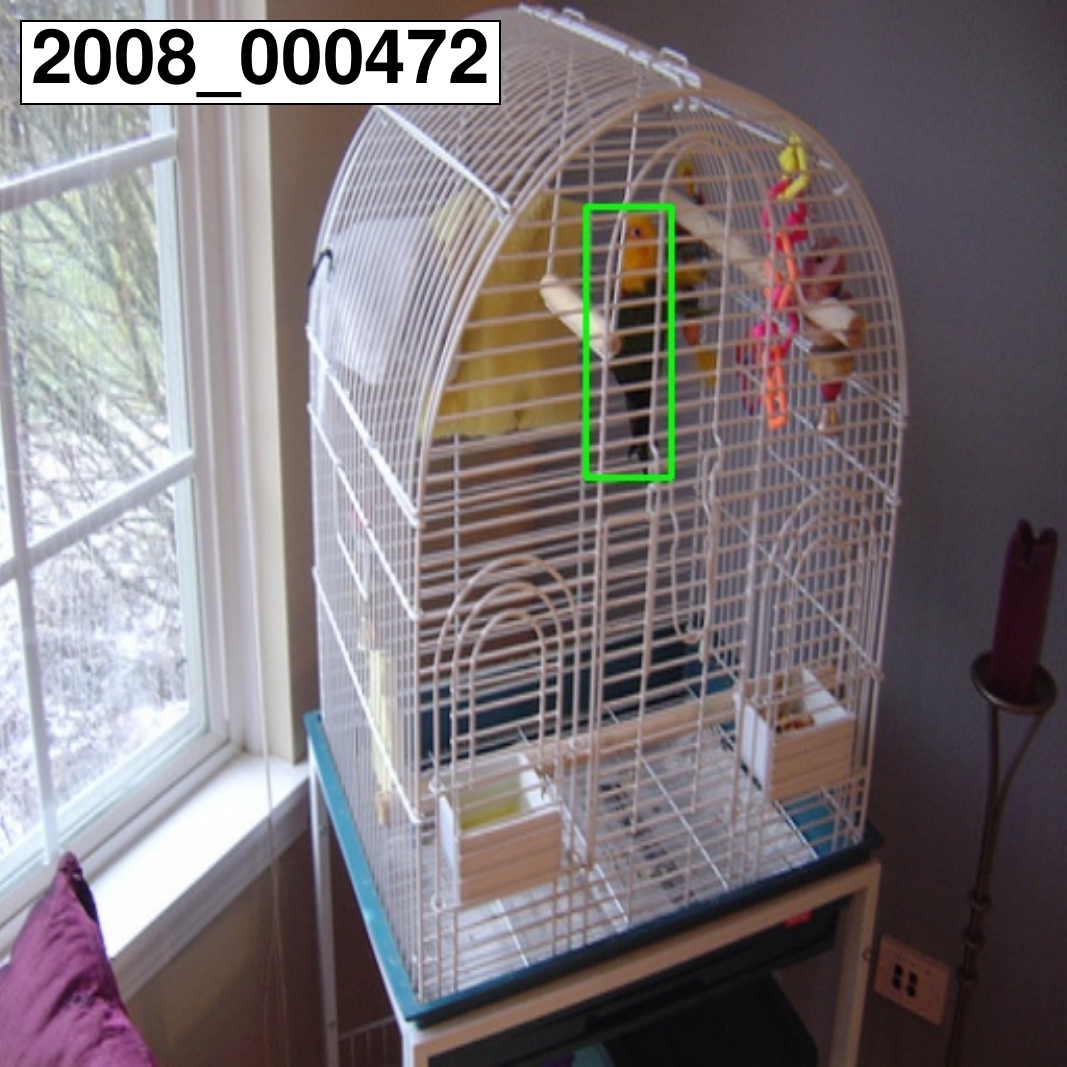}
        \includegraphics[width=0.24\textwidth]{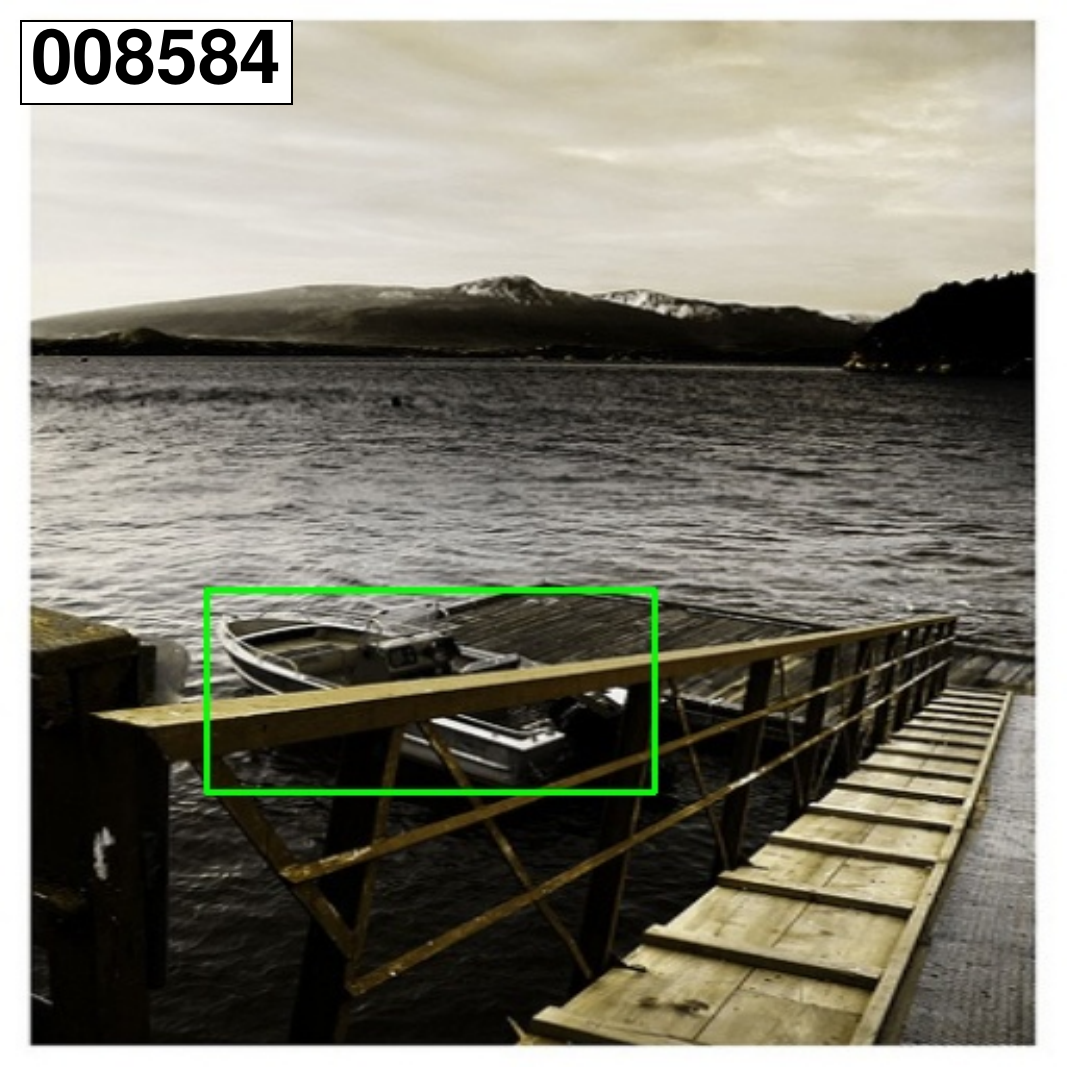}
        \includegraphics[width=0.24\textwidth]{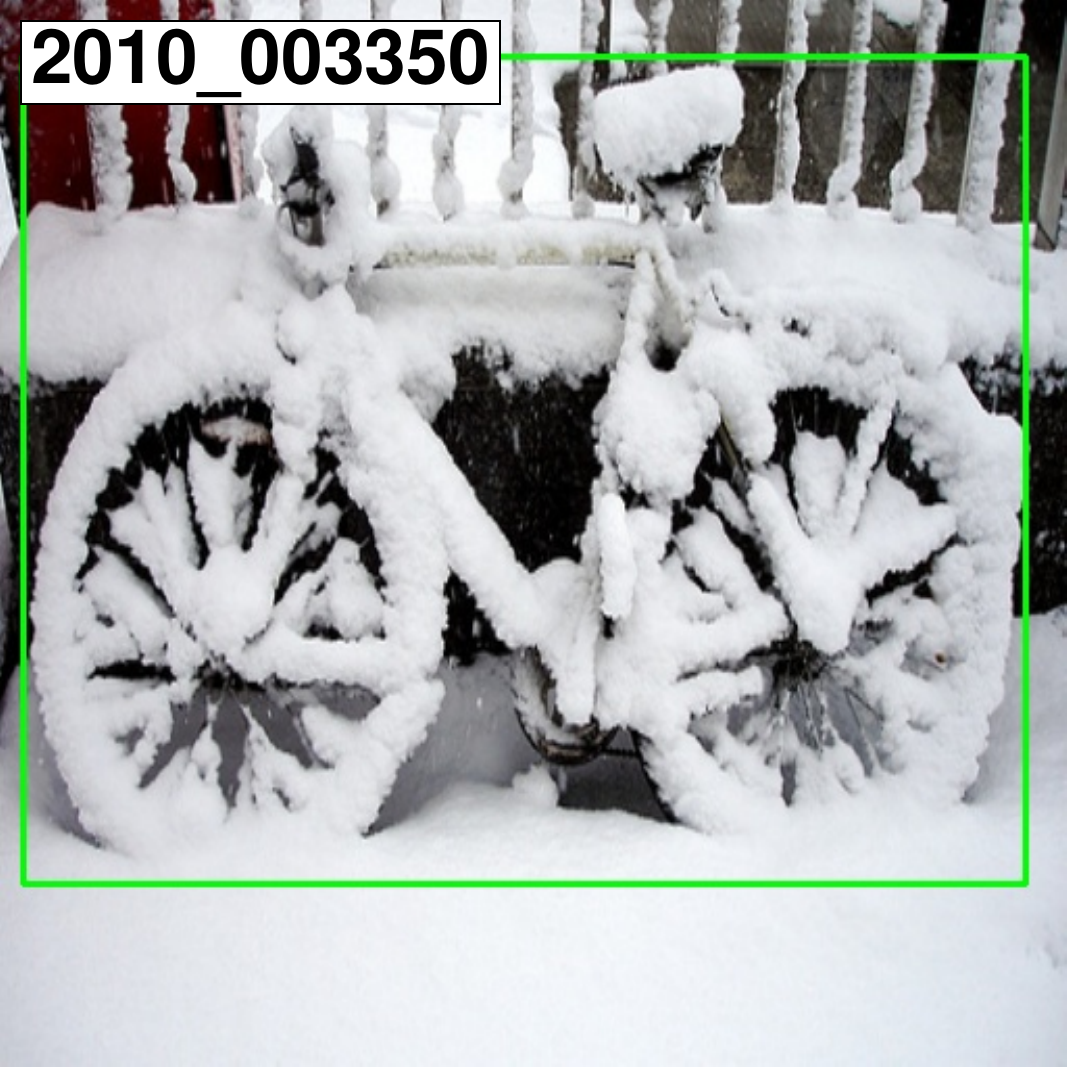}
        \includegraphics[width=0.24\textwidth]{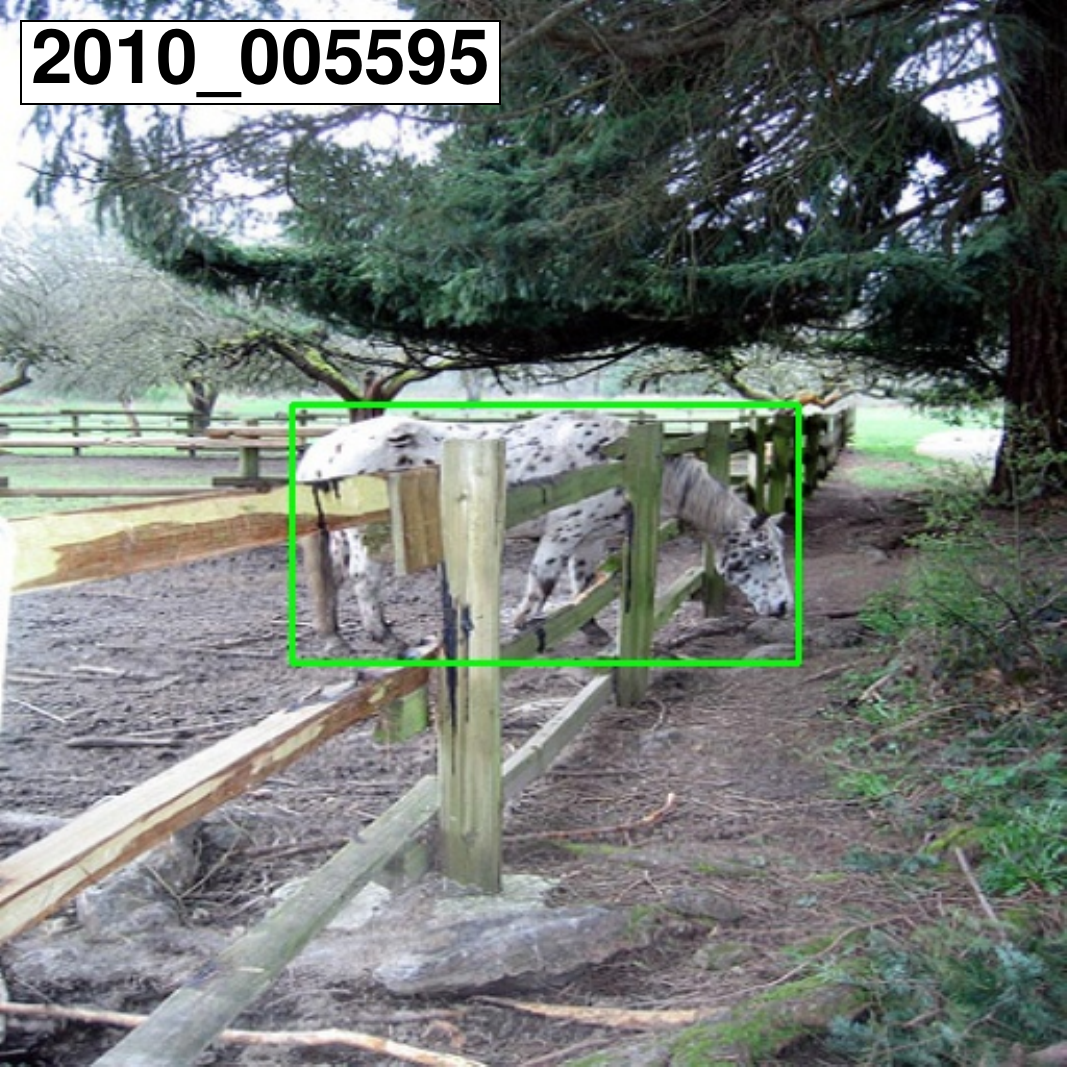}
        \subcaption{Hard pruned samples from PASCAL VOC}
        \label{fig:vochard}
    \end{subfigure}

    \caption{Comparison of sample selections from PASCAL VOC~\cite{everingham2010PASCAL}. (a) Selected samples, (b) Easy pruned samples, and (c) Hard pruned samples.}
    \label{fig:voc_all_samples}
\end{figure}

\begin{figure}[h]
    \centering

    \begin{subfigure}[t]{\textwidth}
        \centering
        \includegraphics[width=0.23\textwidth]{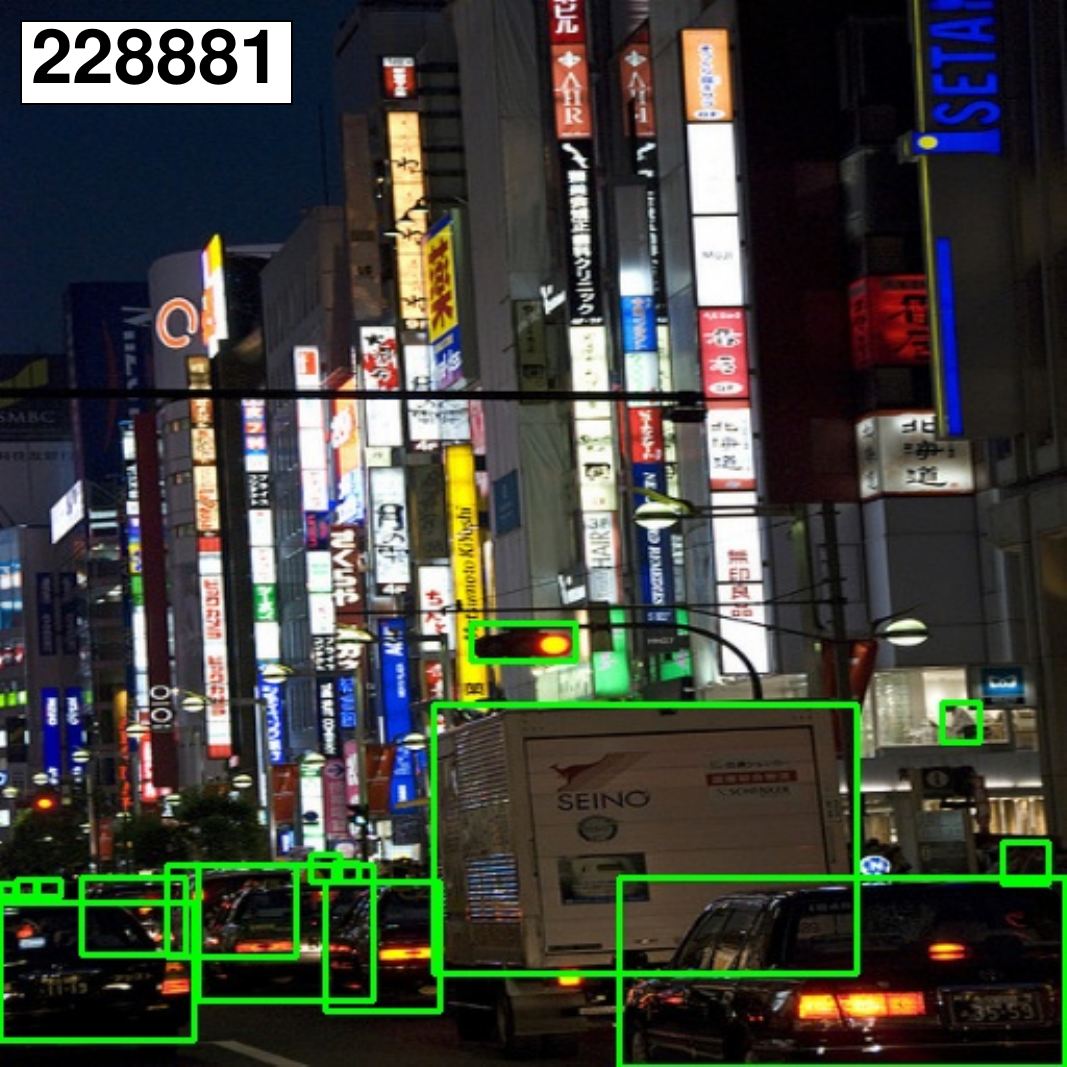}
        \includegraphics[width=0.23\textwidth]{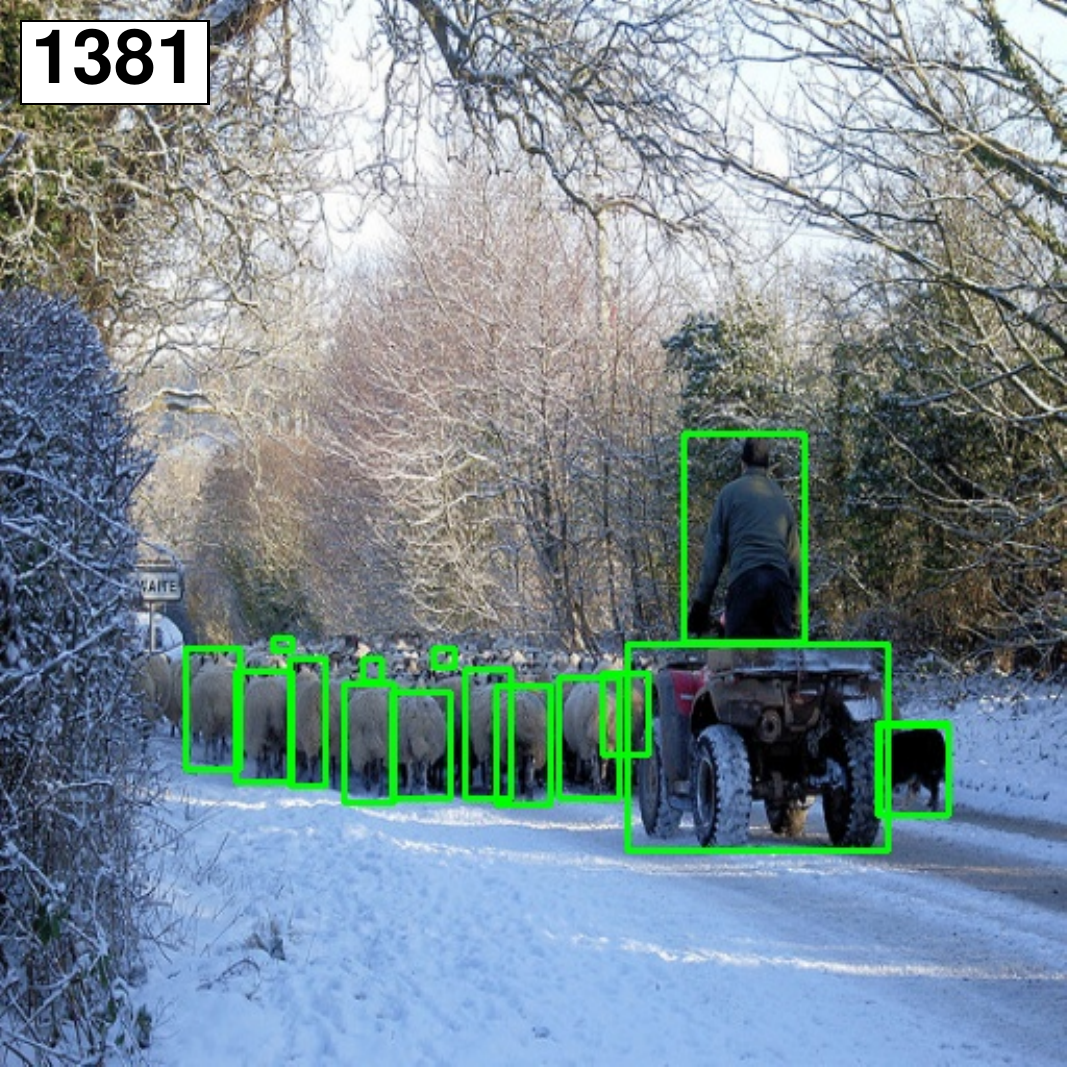}
        \includegraphics[width=0.23\textwidth]{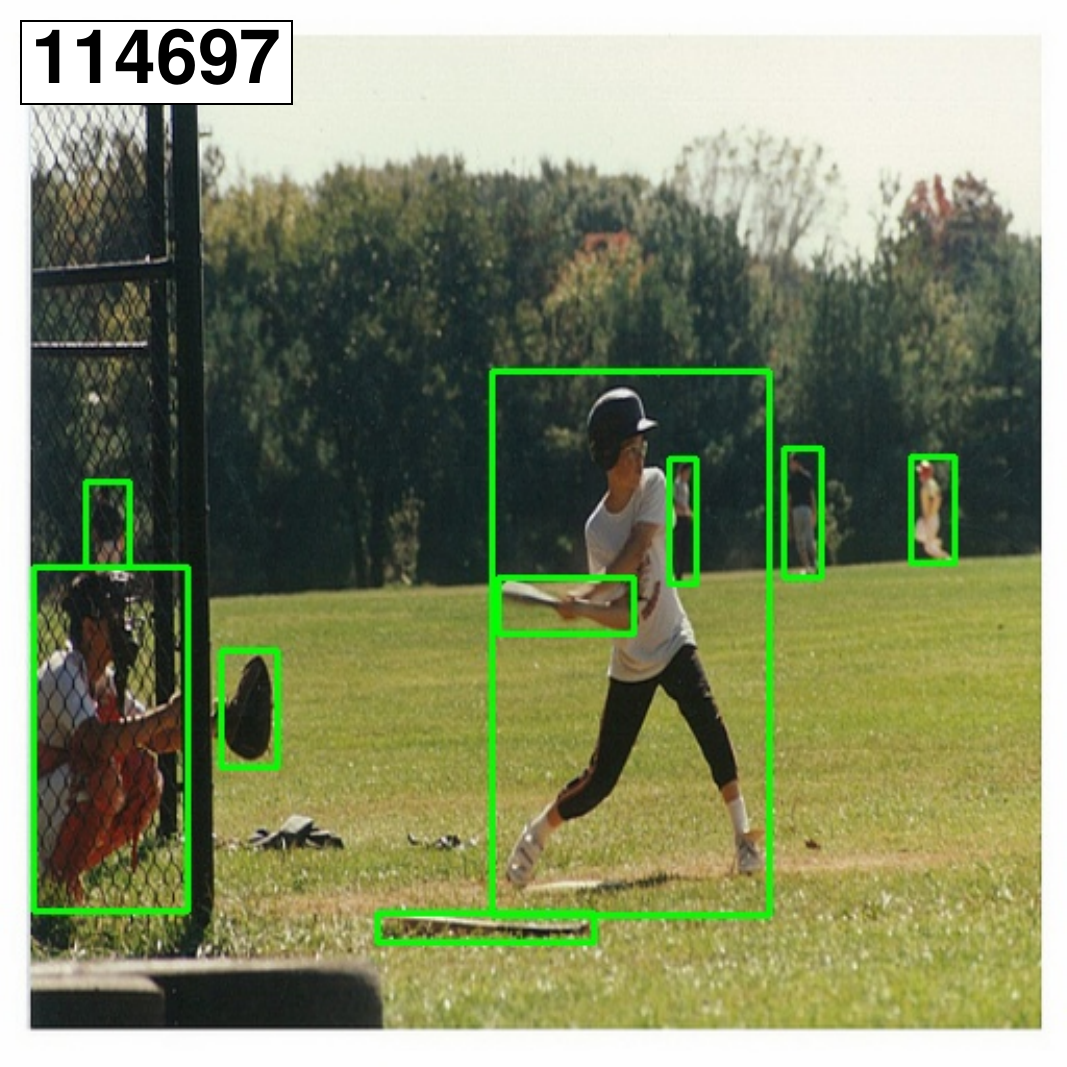}
        \includegraphics[width=0.23\textwidth]{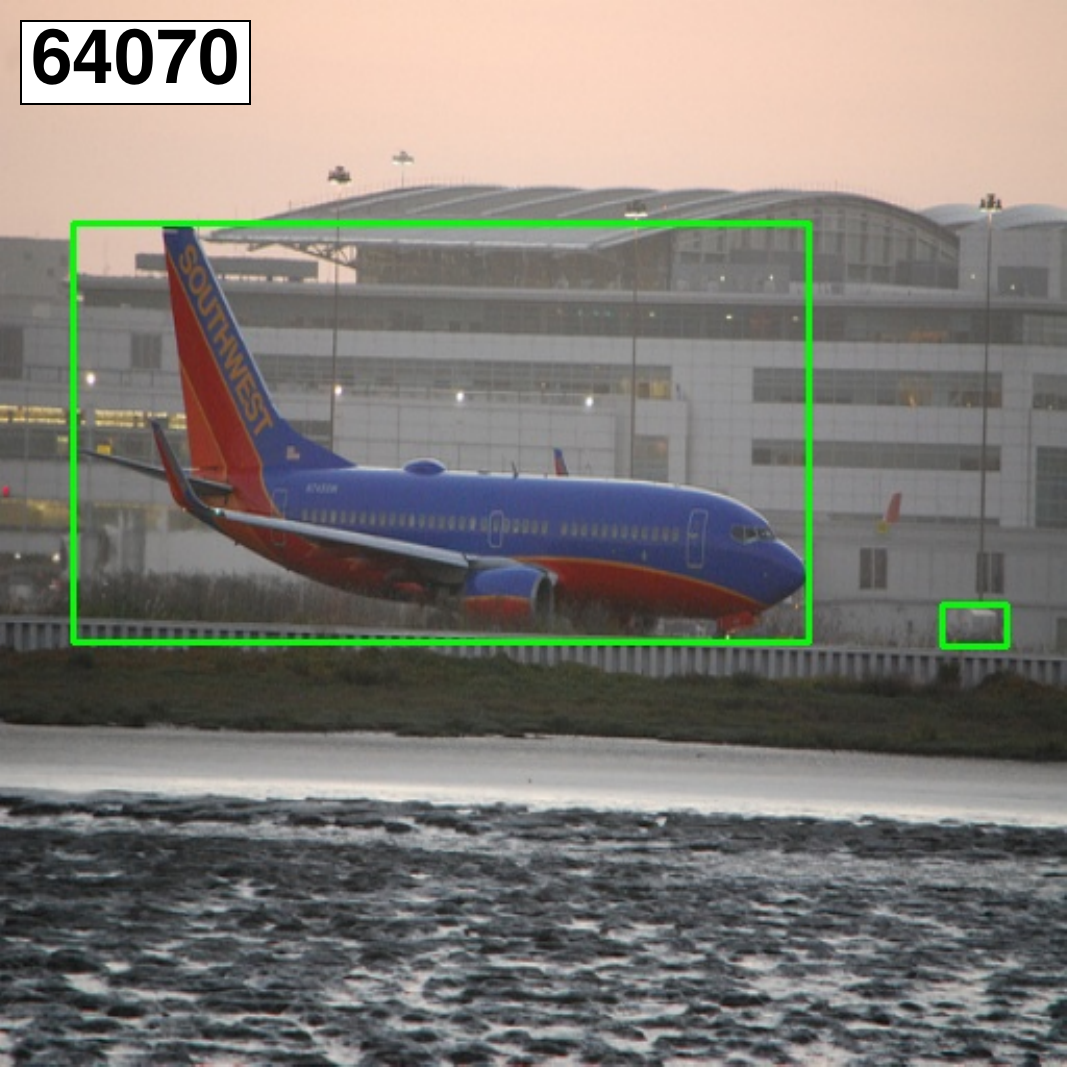}

        \includegraphics[width=0.23\textwidth]{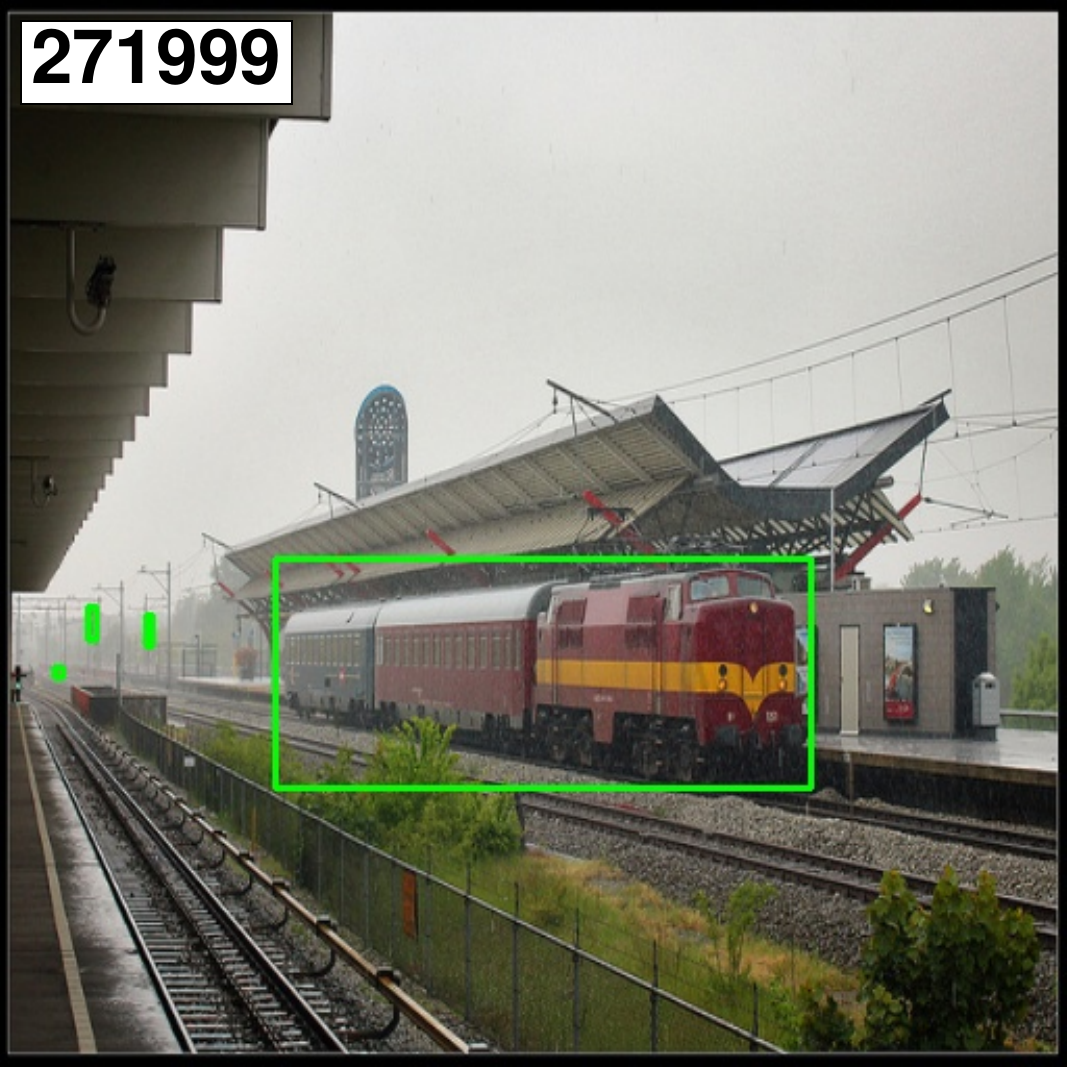}
        \includegraphics[width=0.23\textwidth]{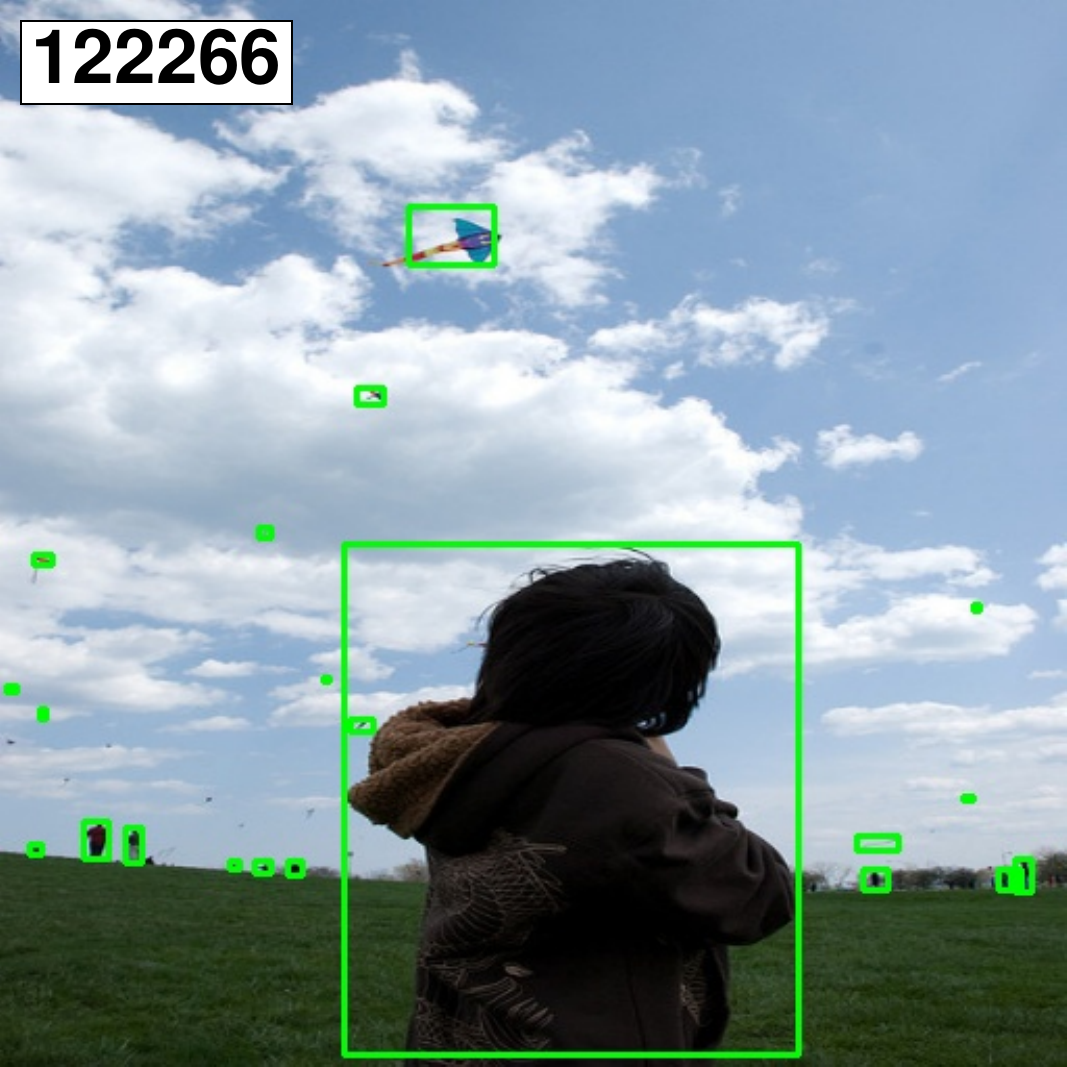}
        \includegraphics[width=0.23\textwidth]{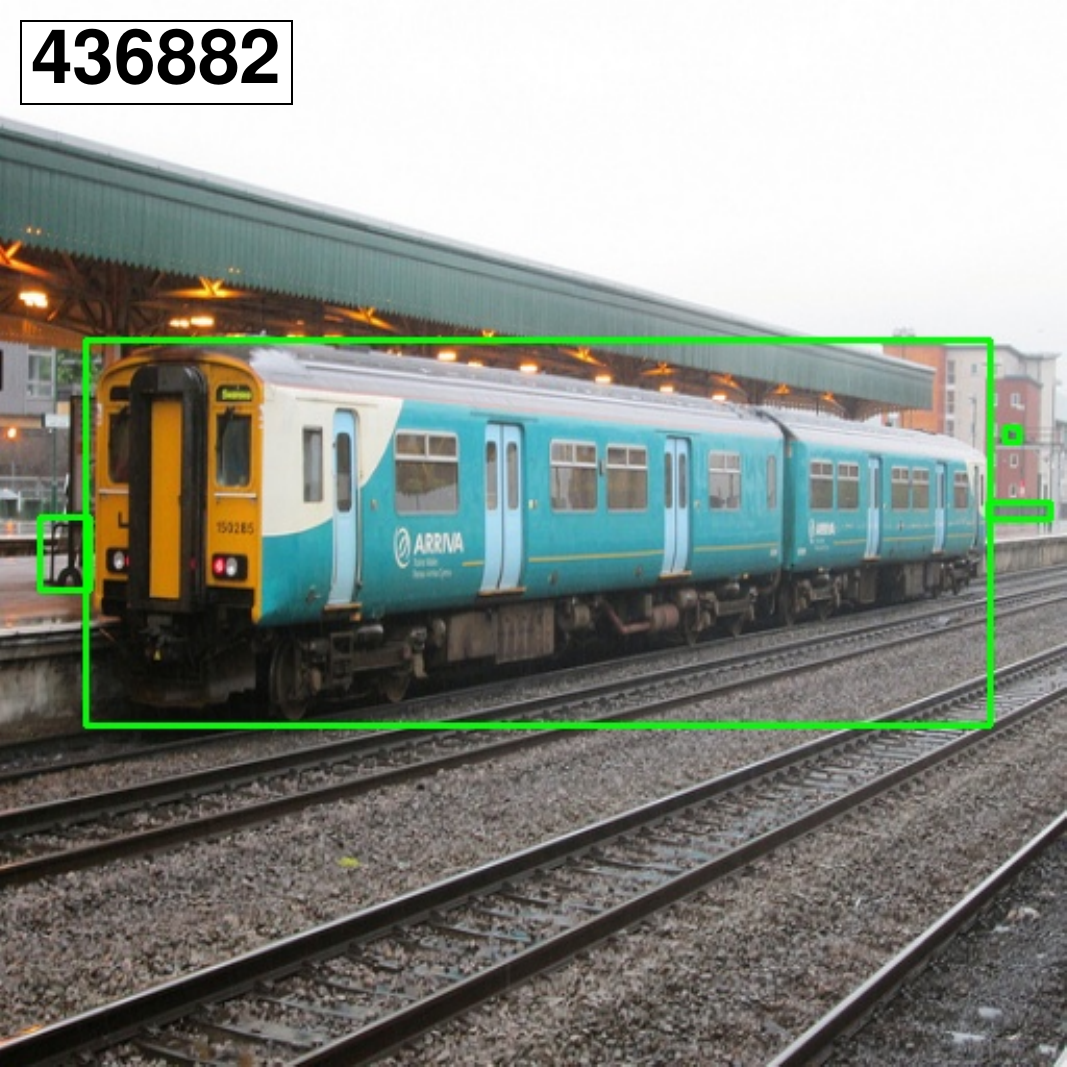}
        \includegraphics[width=0.23\textwidth]{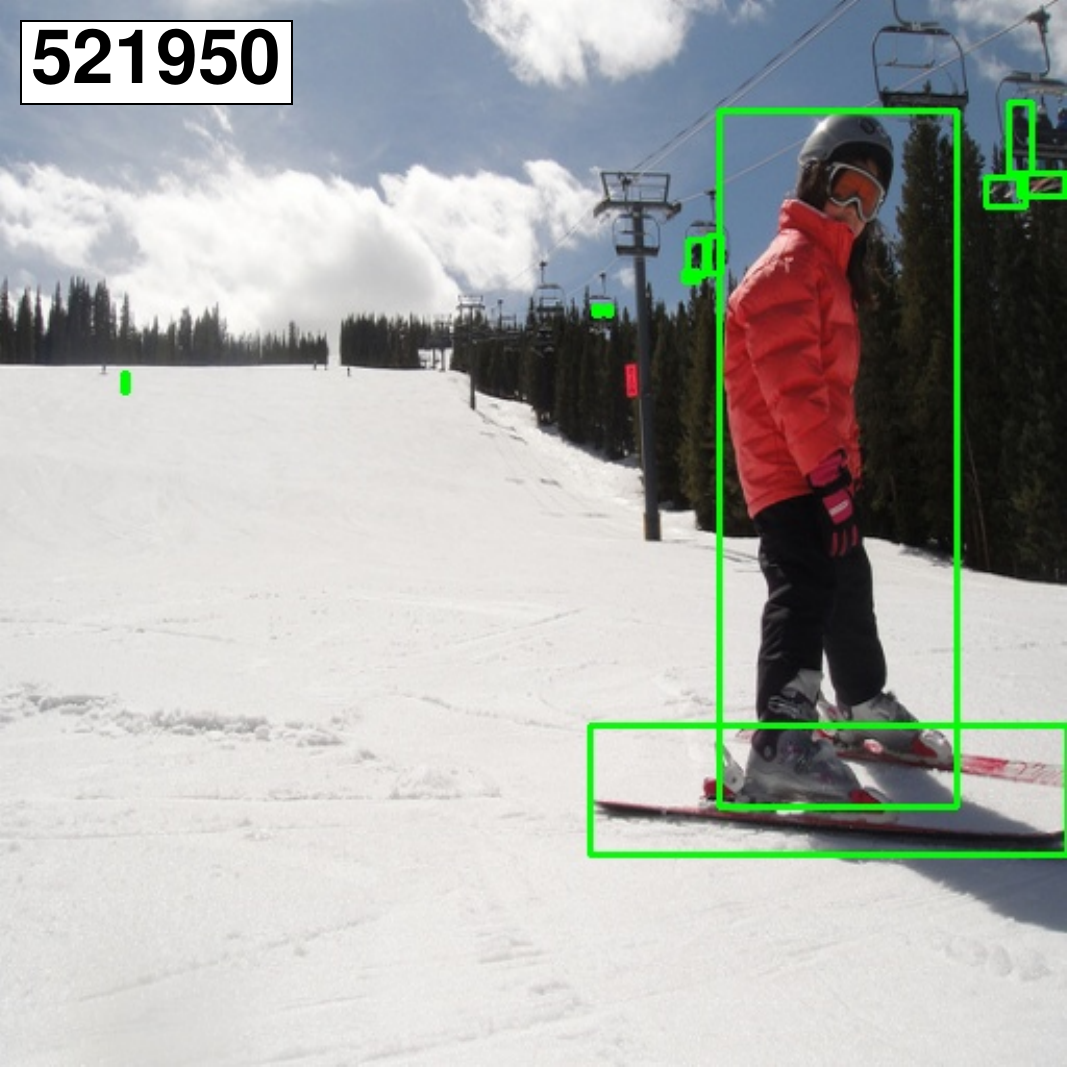}
        \subcaption{Selected samples from MS COCO}
        \label{fig:cocoselected}
    \end{subfigure}

    \vspace{2mm}

    \begin{subfigure}[t]{\textwidth}
        \centering
        \includegraphics[width=0.23\textwidth]{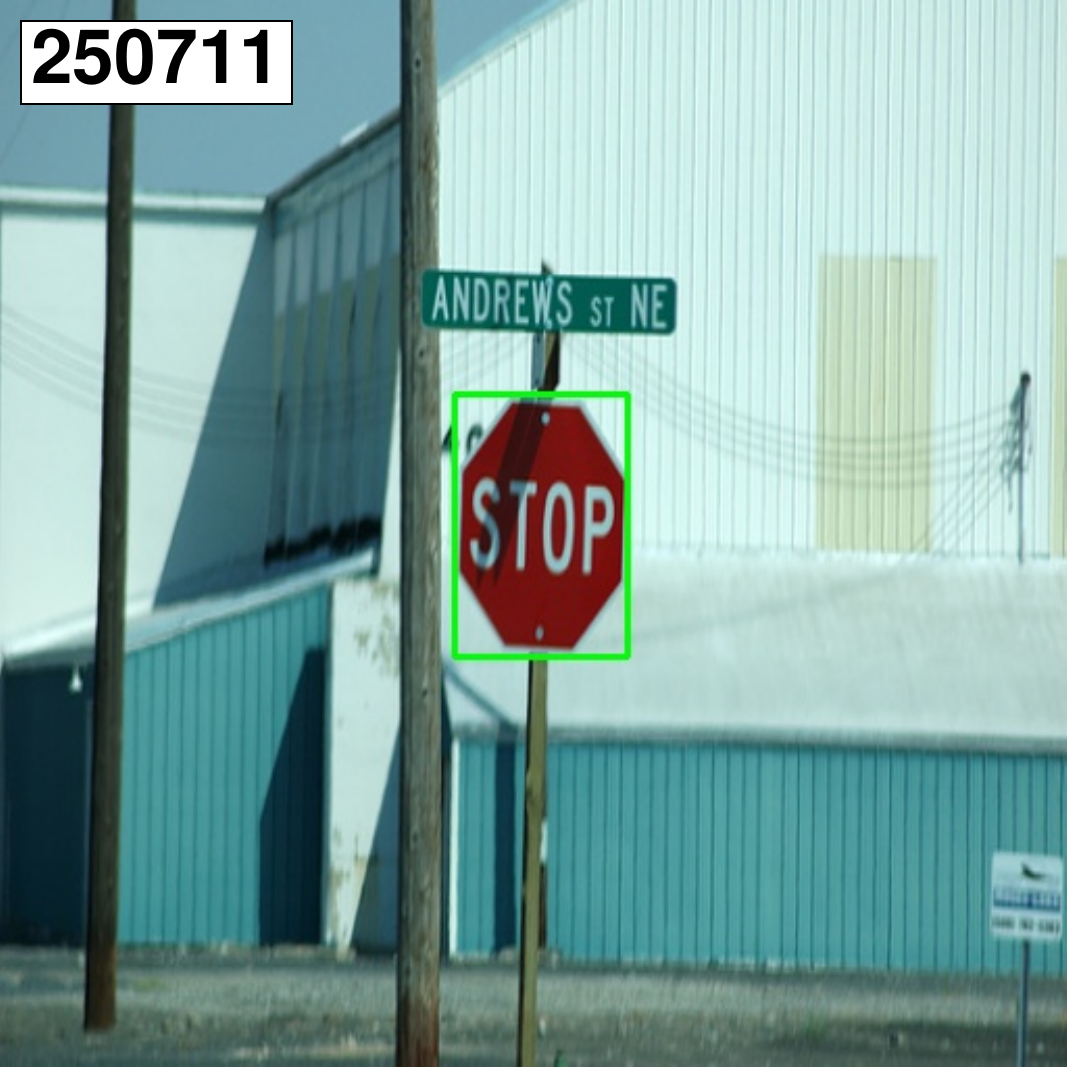}
        \includegraphics[width=0.23\textwidth]{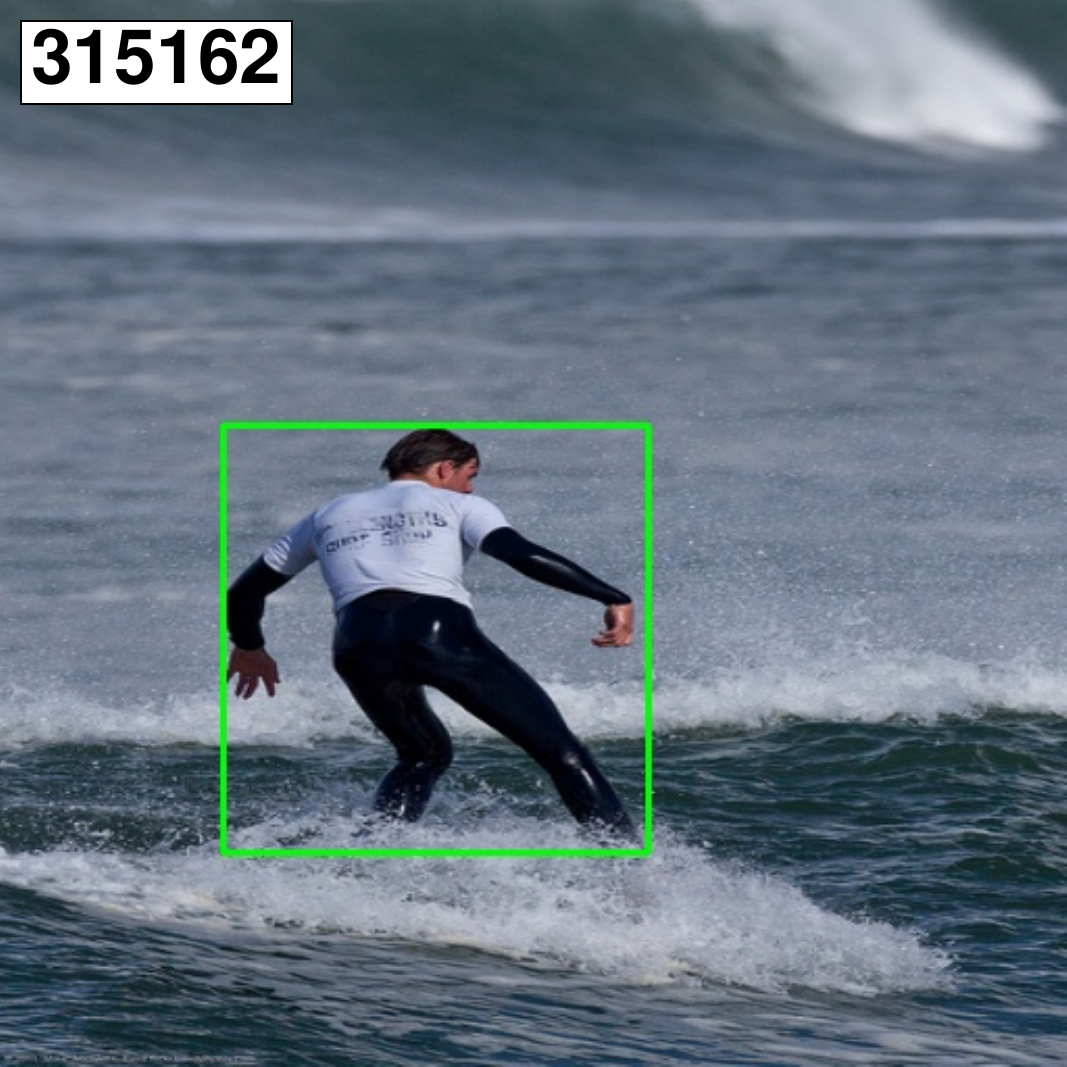}
        \includegraphics[width=0.23\textwidth]{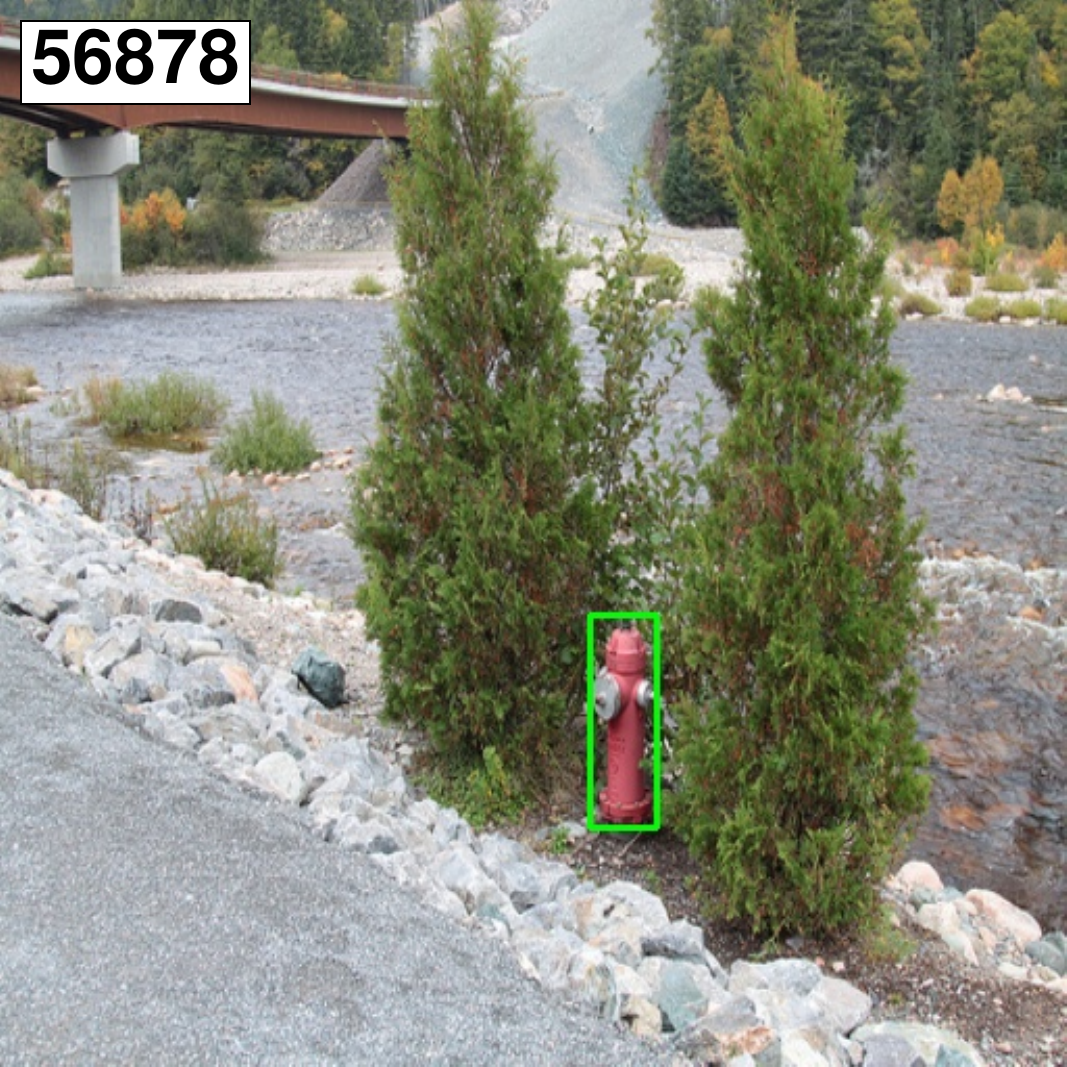}
        \includegraphics[width=0.23\textwidth]{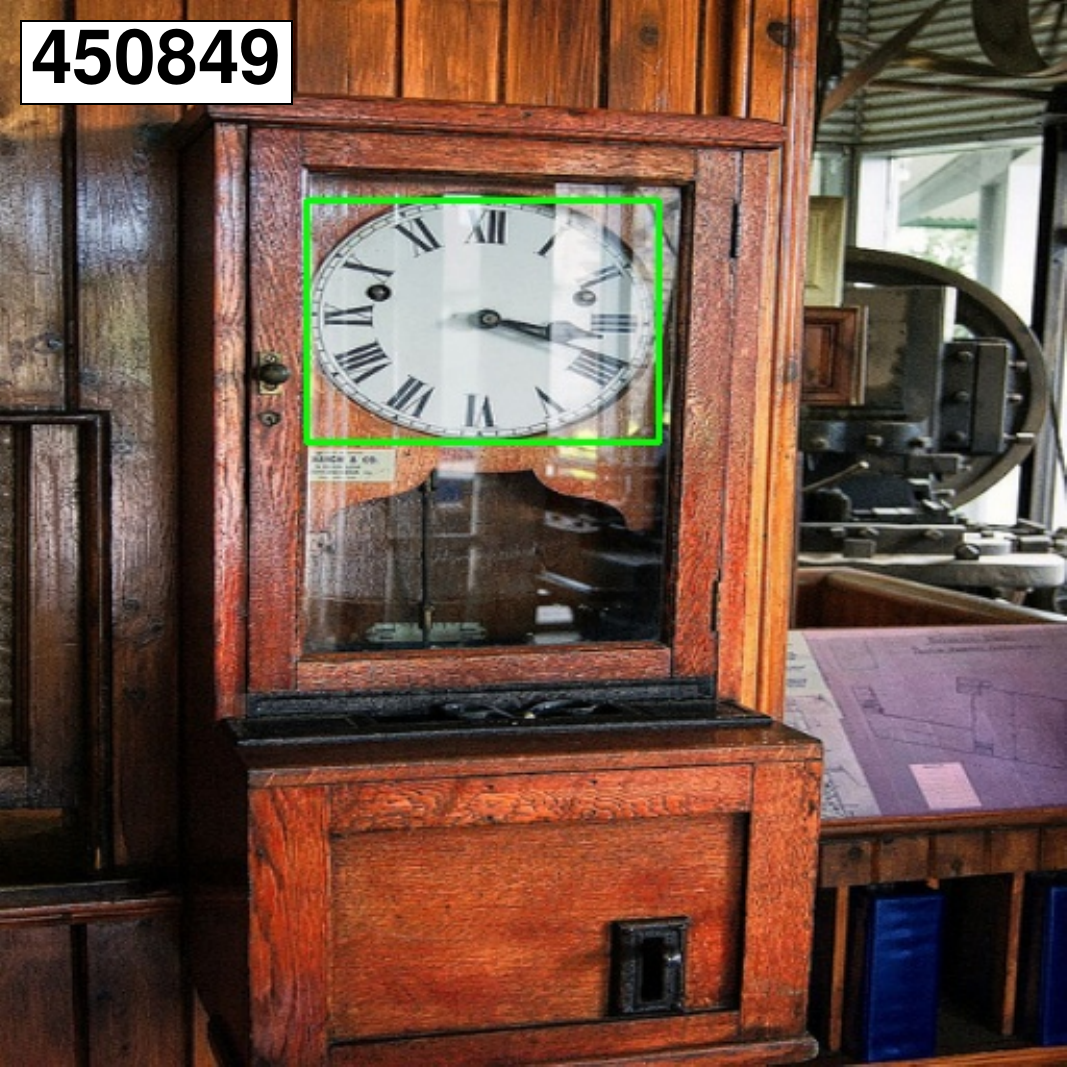}

        \includegraphics[width=0.23\textwidth]{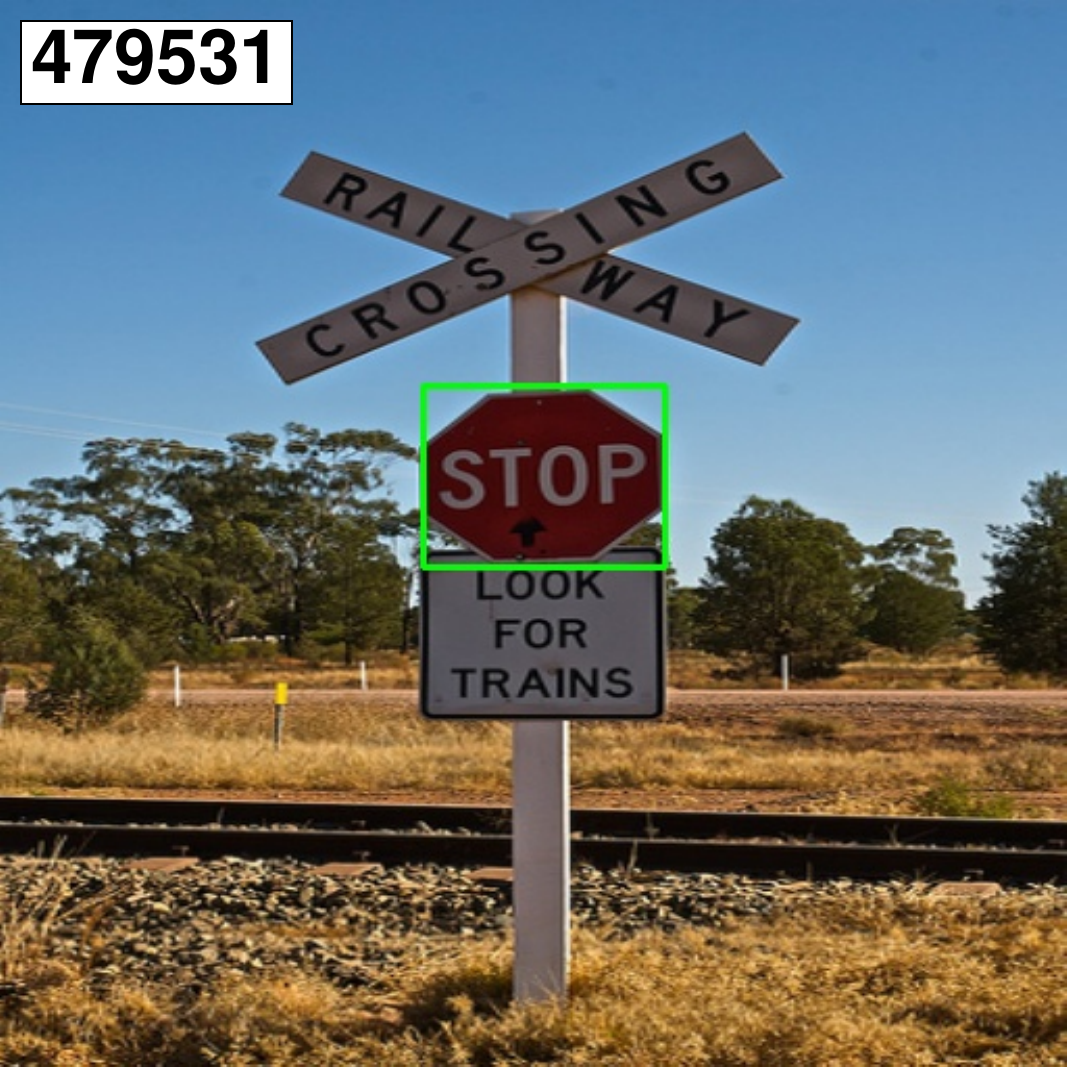}
        \includegraphics[width=0.23\textwidth]{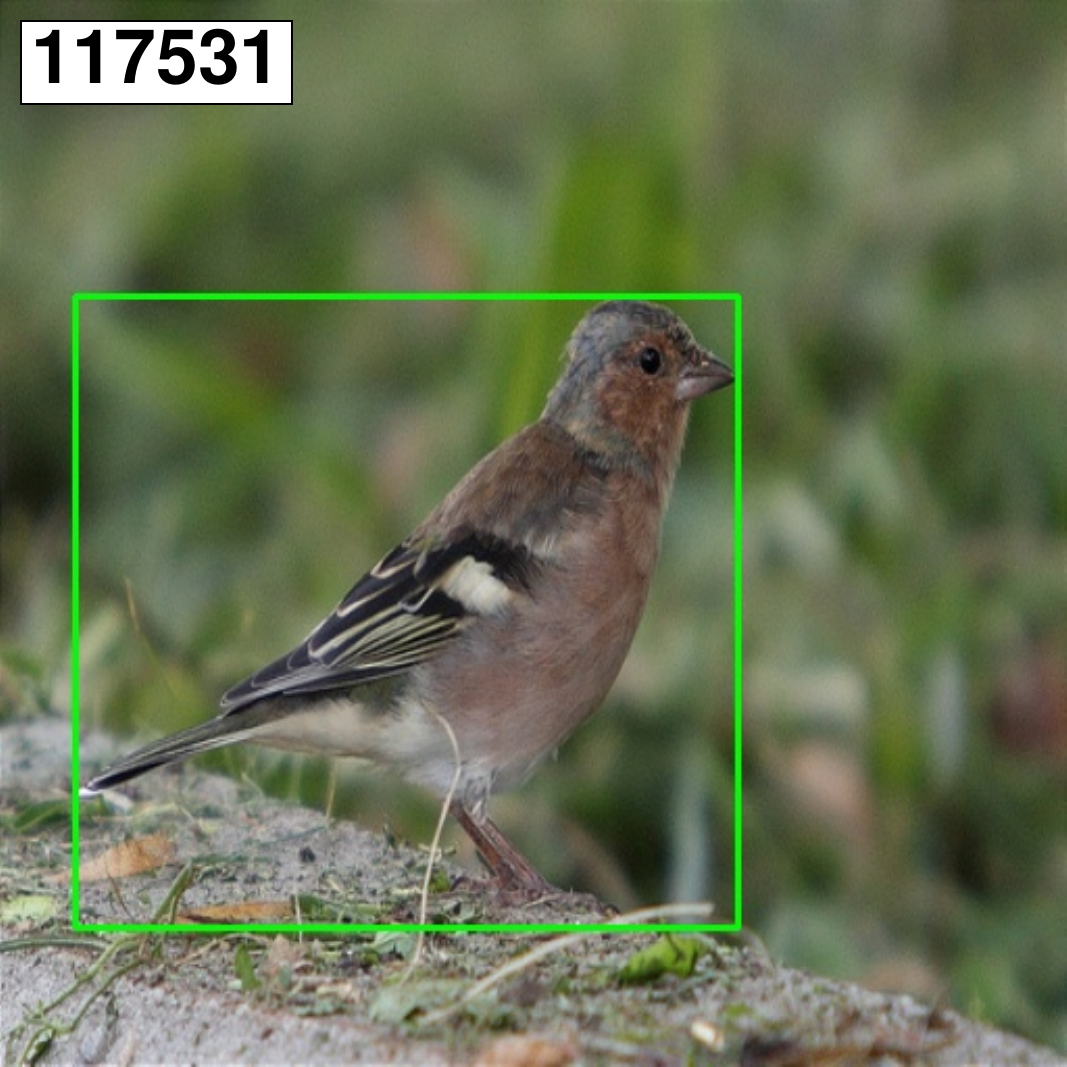}
        \includegraphics[width=0.23\textwidth]{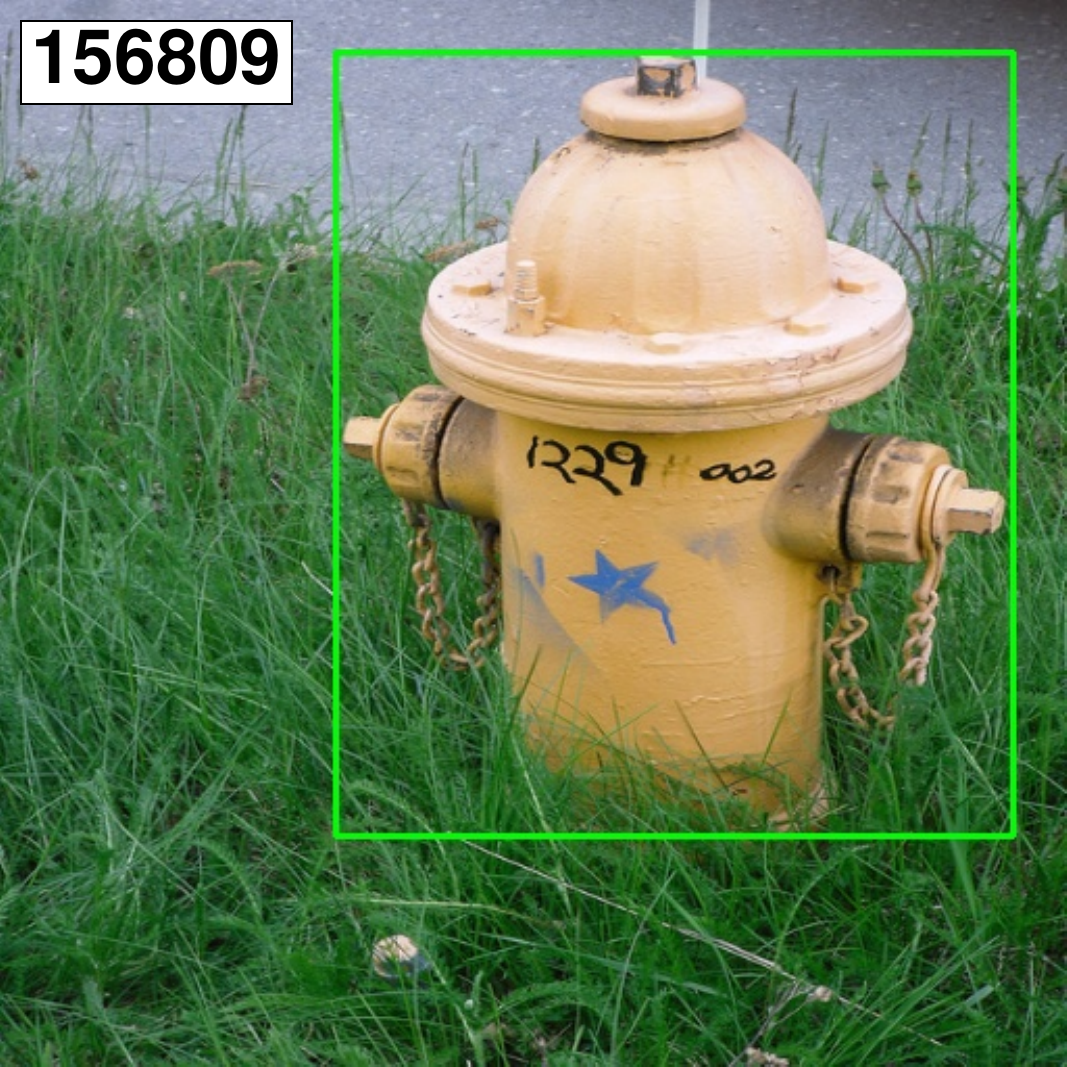}
        \includegraphics[width=0.23\textwidth]{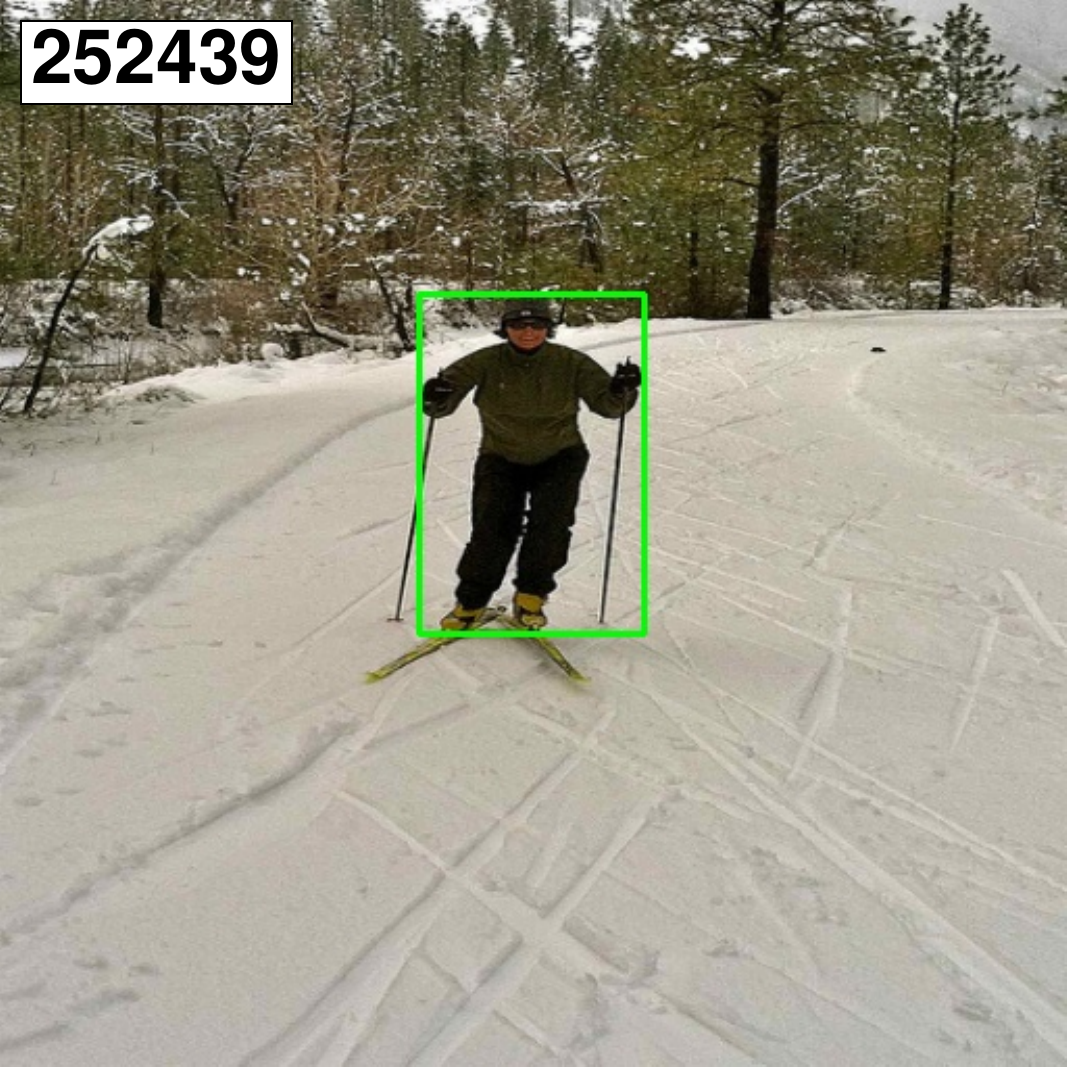}
        \subcaption{Easy pruned samples from MS COCO}
        \label{fig:cocoeasy}
    \end{subfigure}

    \vspace{2mm}

    \begin{subfigure}[t]{\textwidth}
        \centering
        \includegraphics[width=0.23\textwidth]{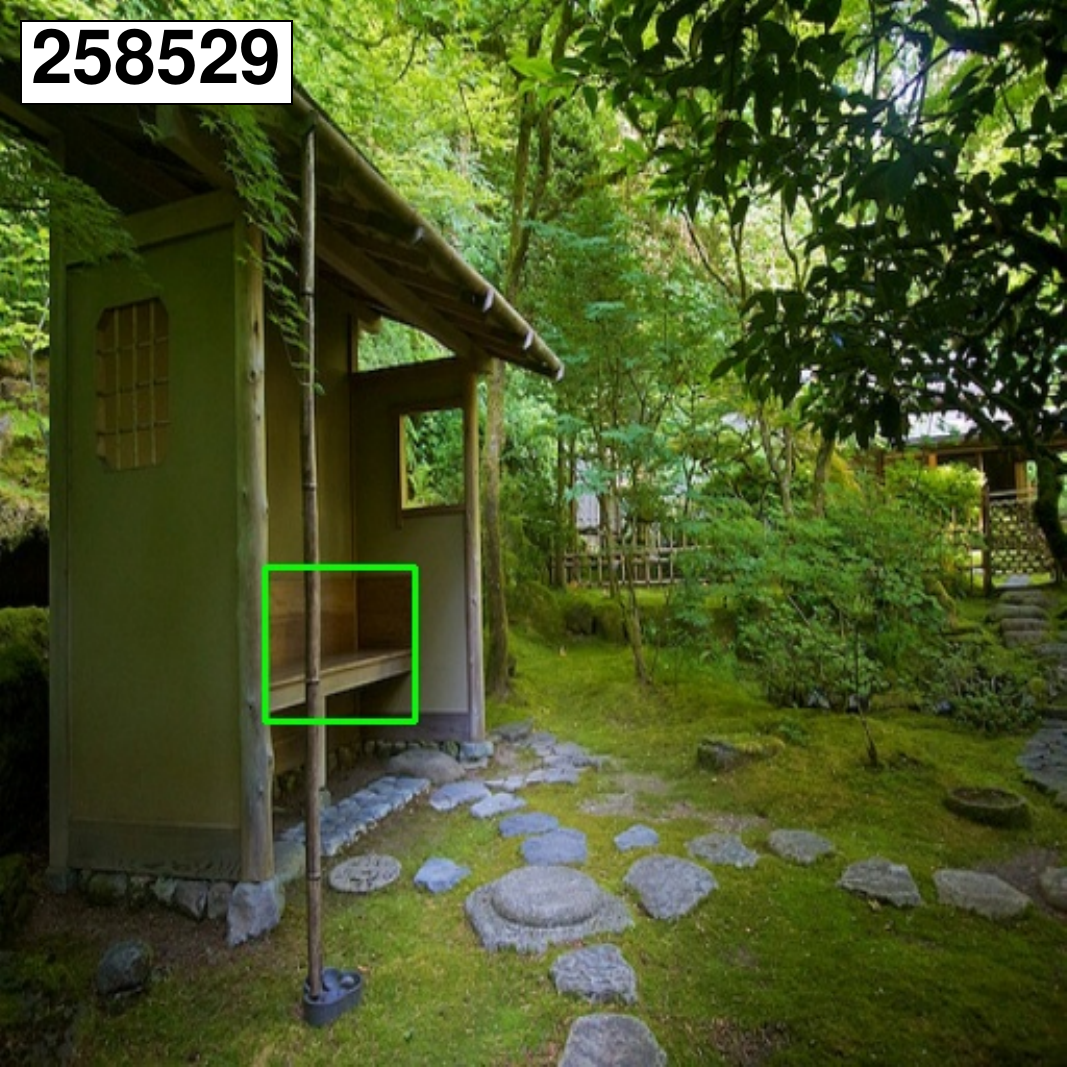}
        \includegraphics[width=0.23\textwidth]{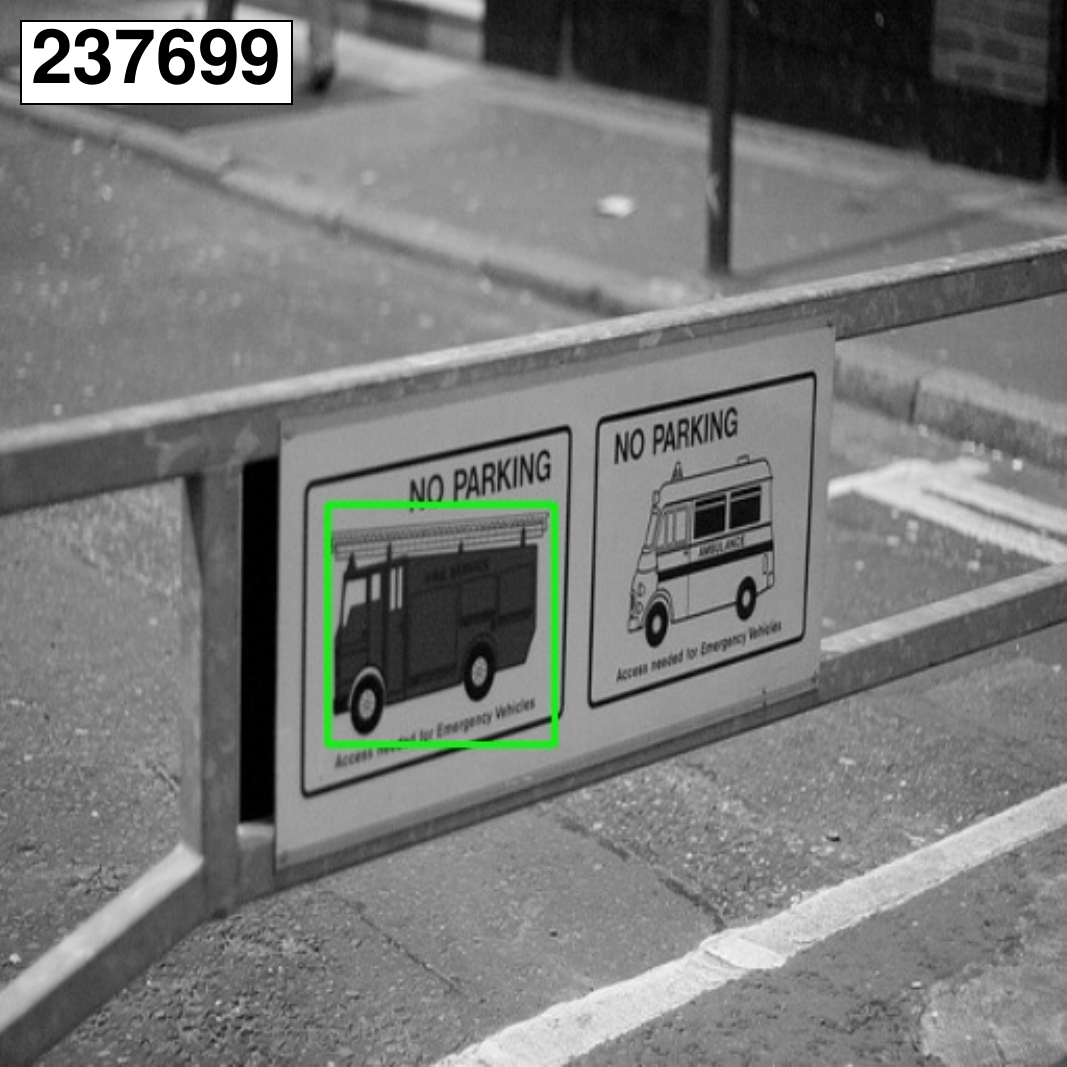}
        \includegraphics[width=0.23\textwidth]{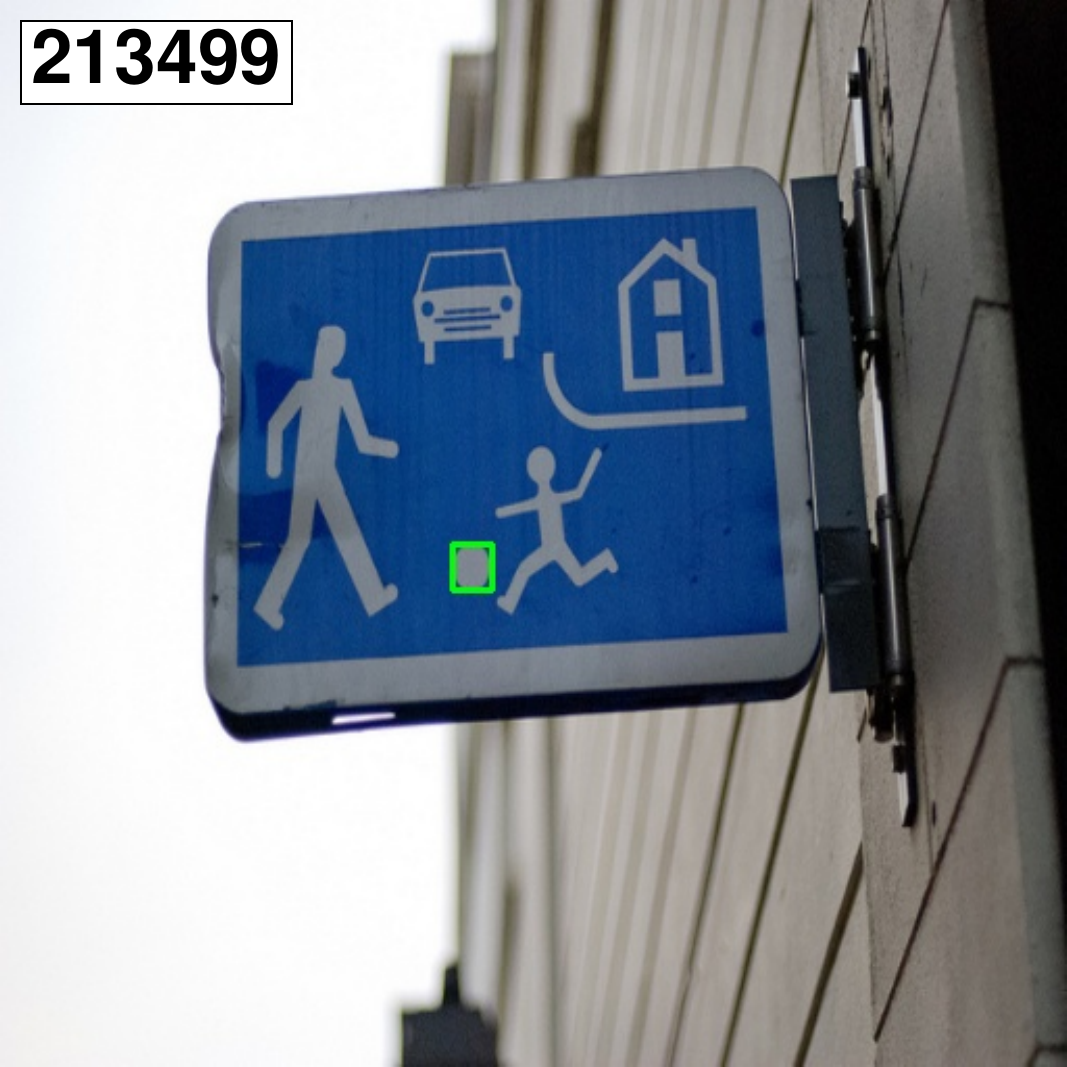}
        \includegraphics[width=0.23\textwidth]{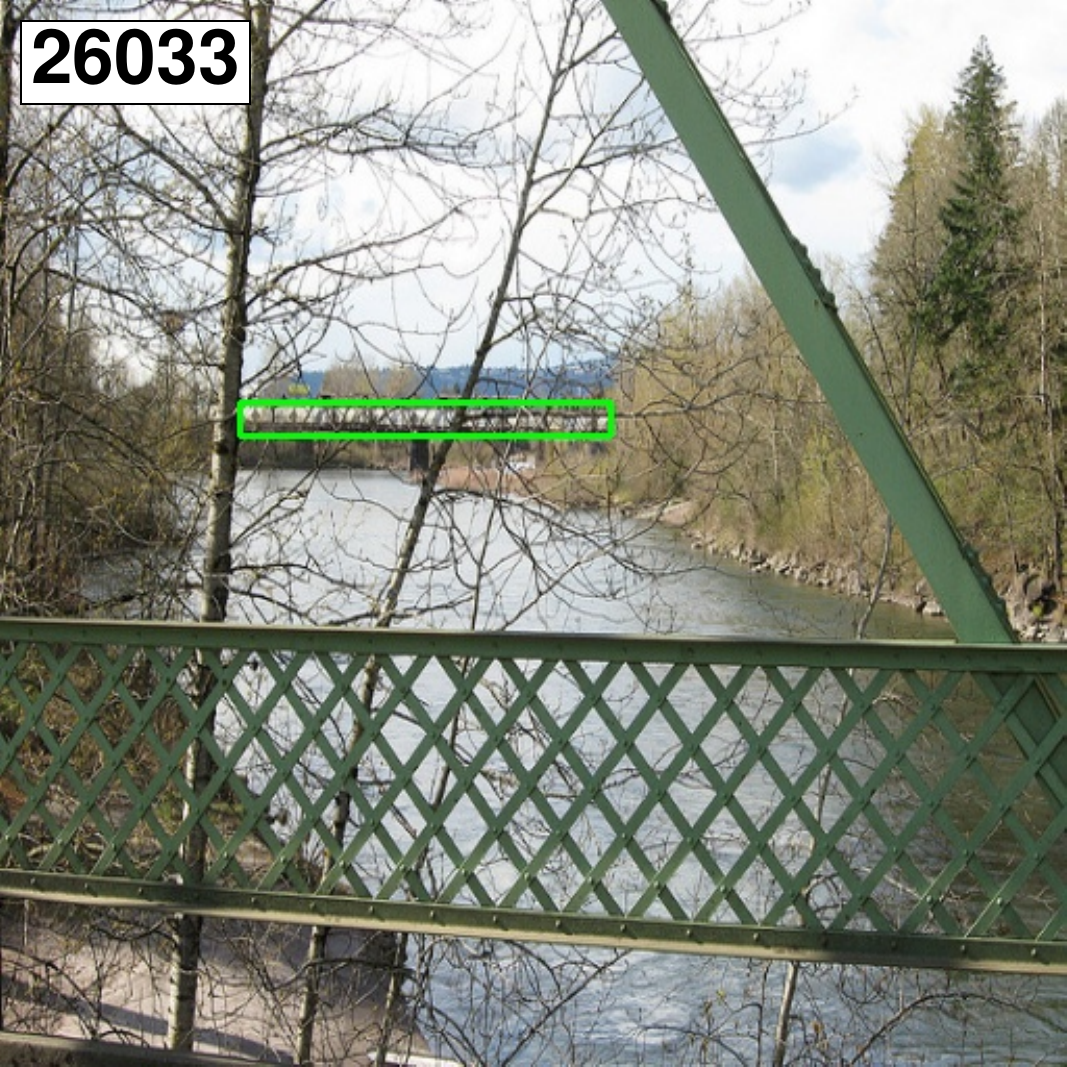}

        \includegraphics[width=0.23\textwidth]{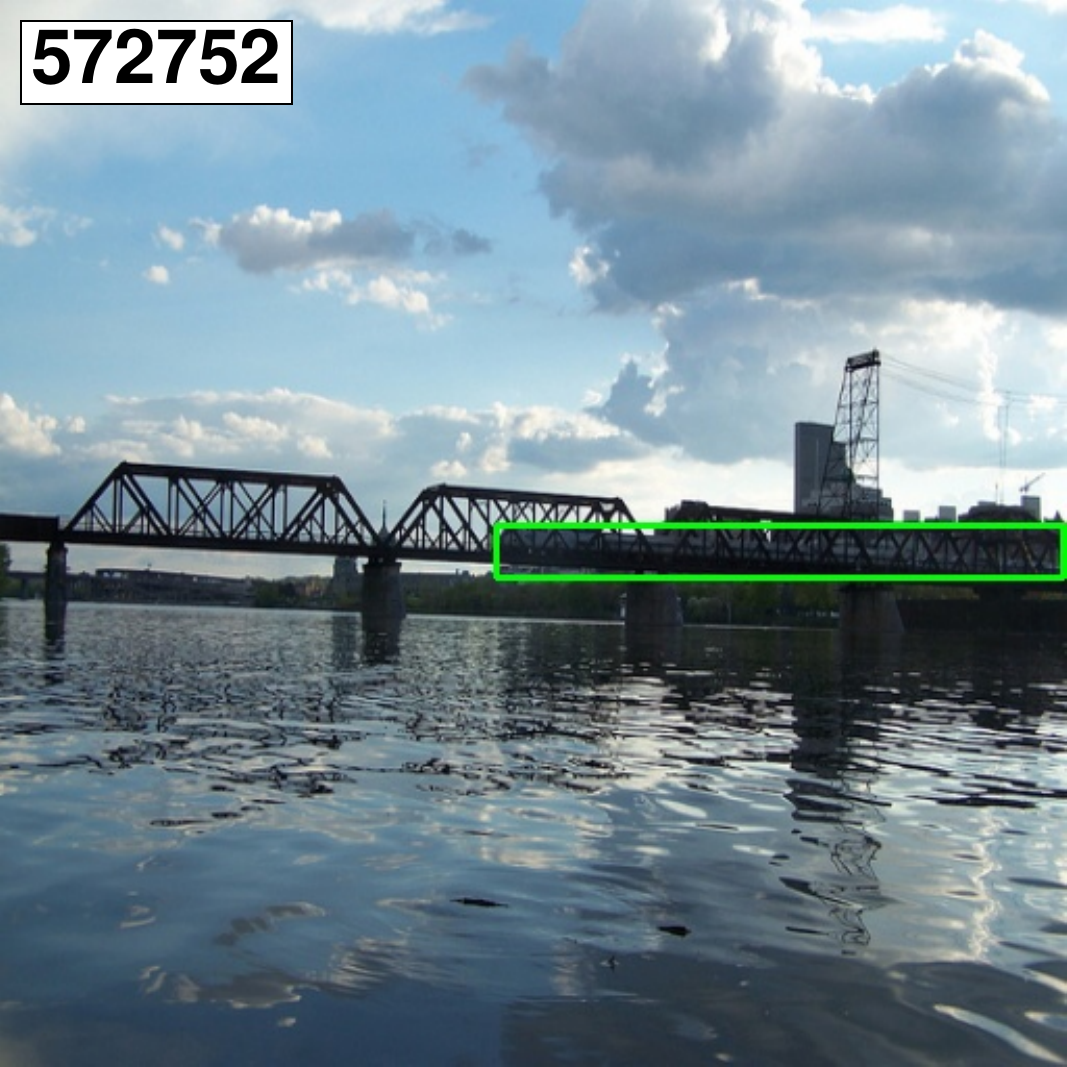}
        \includegraphics[width=0.23\textwidth]{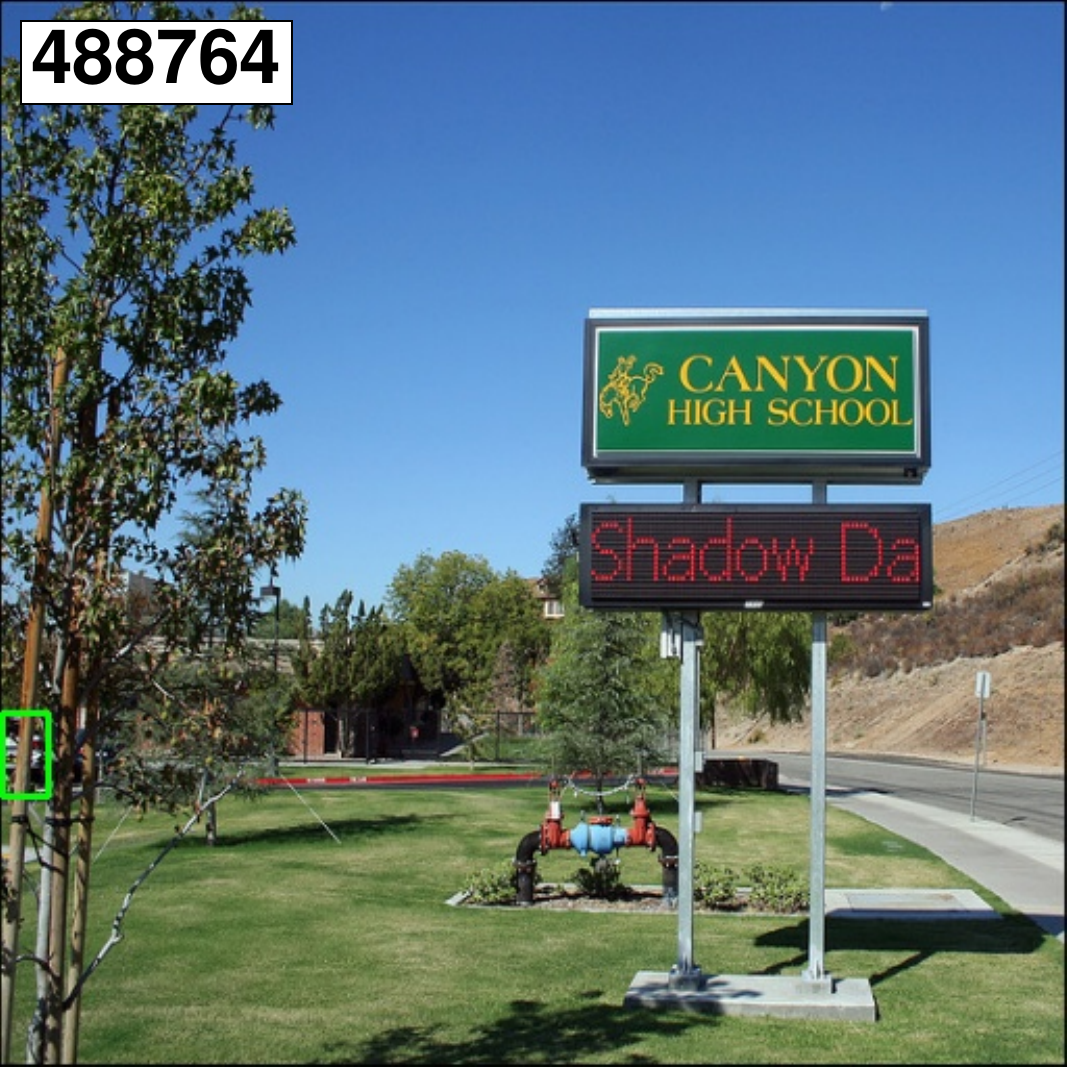}
        \includegraphics[width=0.23\textwidth]{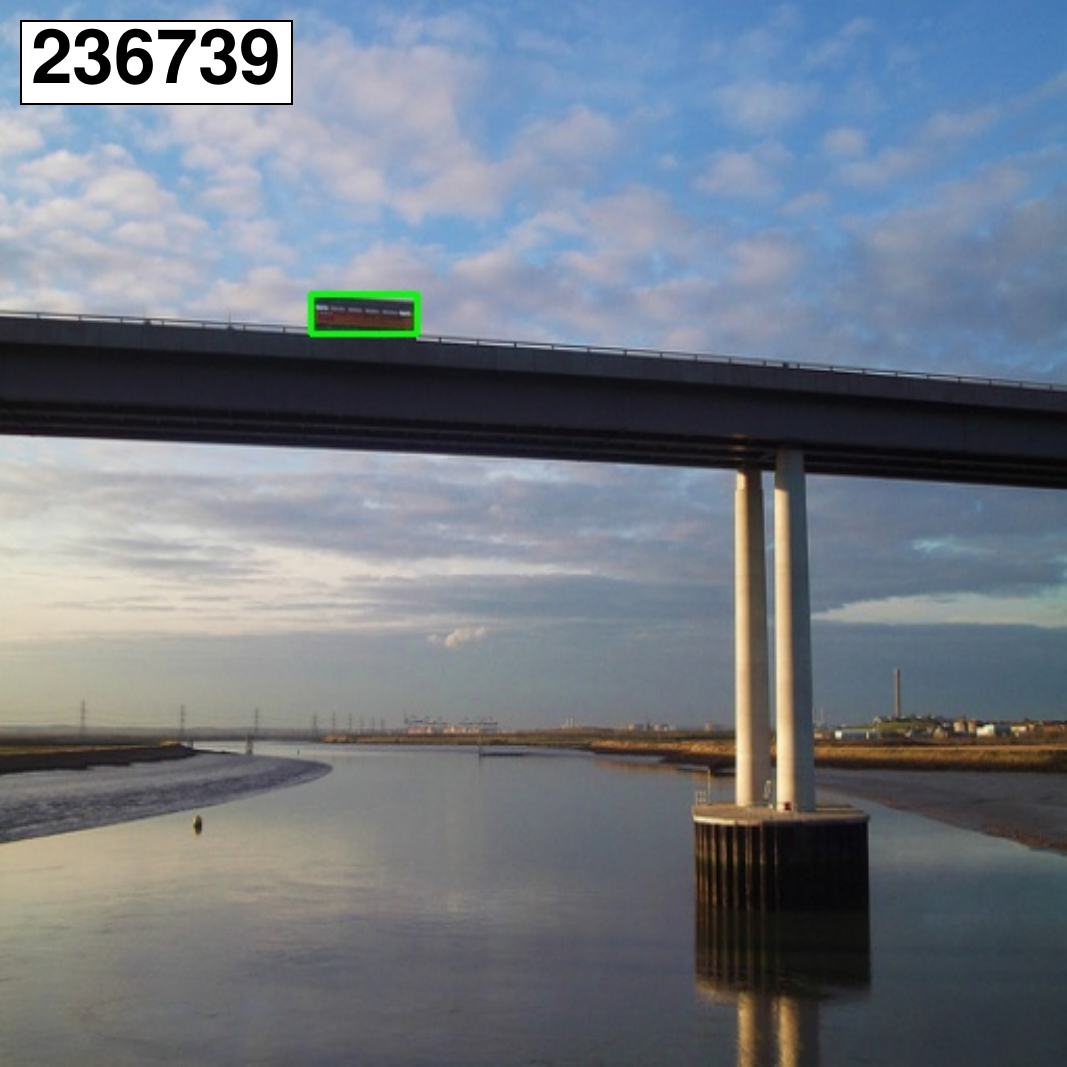}
        \includegraphics[width=0.23\textwidth]{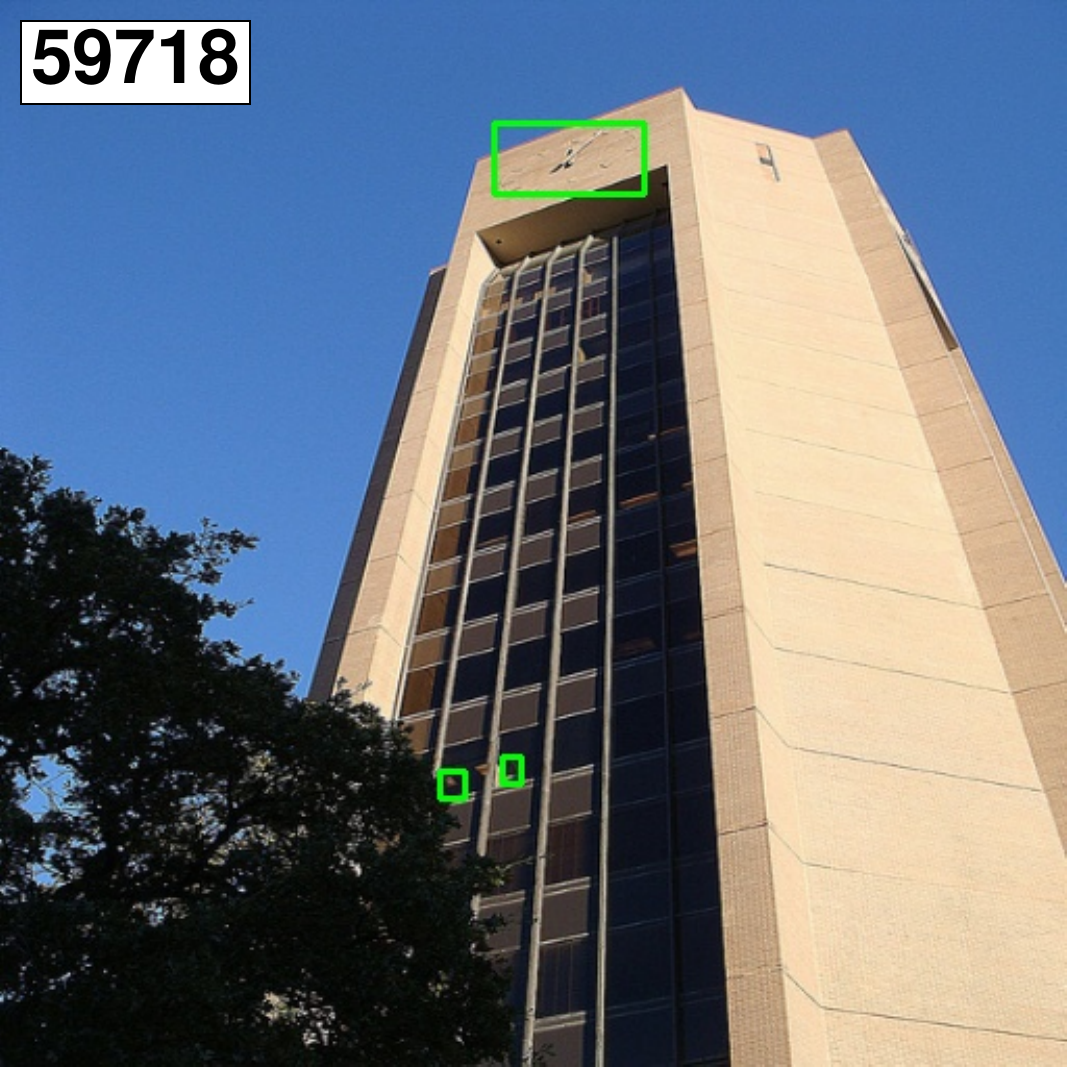}
        \subcaption{Hard pruned samples from MS COCO}
        \label{fig:cocohard}
    \end{subfigure}

    \caption{Comparison of sample selections from MS COCO~\cite{lin2014microsoft}. (a) Selected samples, (b) Easy pruned samples, and (c) Hard pruned samples.}
    \label{fig:coco_all_samples}
\end{figure}

We begin by presenting the top-ranked samples—those with the highest VPS scores according to our ranking methodology. Specifically, Figure~\ref{fig:vocselected} shows samples from the PASCAL VOC dataset selected using the VPS\textsubscript{iou} score with max aggregation, while Figure~\ref{fig:cocoselected} displays samples from the MS COCO dataset selected using the VPS\textsubscript{conf} score with max aggregation.

We then visualize the pruned data points, which were identified as either easy or hard examples based on their average IoU and confidence values. For the PASCAL VOC dataset, Figures~\ref{fig:voceasy} and~\ref{fig:vochard} present easy and hard samples, respectively, selected using VPS\textsubscript{iou} (max aggregation). Similarly, for the MS COCO dataset, Figures~\ref{fig:cocoeasy} and~\ref{fig:cocohard} show easy and hard examples identified using VPS\textsubscript{conf} (max aggregation).
Notably, Figure~\ref{fig:cocohard} includes cases where annotations are assigned to objects depicted on signs, such as trucks or balls, as well as extremely small objects, occluded instances, or examples where the distinction between foreground and background is particularly challenging.

\clearpage

\clearpage

\section{Social Impacts}
\label{sec:social}

Our work explores dataset pruning techniques to reduce the size of training data while preserving model performance. The potential positive societal impacts include improving training efficiency, reducing carbon emissions by lowering computational demand, and increasing accessibility to machine learning for those with limited resources.

We believe there are no direct negative societal impacts associated with our method, as it is a foundational contribution without immediate application to sensitive domains. However, we acknowledge that dataset pruning could unintentionally exacerbate data bias if not applied carefully, especially when removing examples from underrepresented groups. Therefore, we encourage practitioners to evaluate fairness metrics before and after pruning, particularly in sensitive applications such as healthcare or criminal justice.

\end{document}